\documentclass[twoside,11pt]{article}
\usepackage{jair, theapa, rawfonts}

\usepackage{balance} 
\usepackage{graphicx}

\usepackage{amsthm}
\usepackage{amssymb}
\usepackage{xspace}
\usepackage{caption}
\usepackage{subcaption}
\usepackage{calrsfs}
\usepackage[fleqn]{mathtools}
\usepackage{color}
\usepackage{booktabs}
\usepackage{setspace}
\usepackage{float}
\usepackage{siunitx}
\usepackage{multirow}
\usepackage{hyphenat}
\usepackage{url}
\usepackage{amsmath}
\usepackage[displaymath, mathlines]{lineno}

\DeclareMathOperator*{\argmax}{argmax} 
\newcommand{\BibTeX}{\rm B\kern-.05em{\sc i\kern-.025em b}\kern-.08em\TeX}
\newcommand{\stt}[1]{{\small\texttt{#1}}}
\newcommand{\no}[1]{{\; {not} \;}}
\newcommand{\myif}[1]{{\texttt{:-}}}

\DeclareMathAlphabet{\pazocal}{OMS}{zplm}{m}{n}
\theoremstyle{definition}
\newtheorem{defn}{Definition} 

\newenvironment{frcseries}{\fontfamily{frc}\selectfont}{}

\jairheading{79}{2024}{725-776}{12/2023}{02/2024}
\ShortHeadings{Learning Logic Specifications as POMDP Heuristics}
{Meli, Castellini, \& Farinelli}
\firstpageno{725}

\begin{document}

\title{Learning Logic Specifications for Policy Guidance in POMDPs: an Inductive Logic Programming Approach}

\author{\name Daniele Meli \email daniele.meli@univr.it \\
       \name Alberto Castellini \email alberto.castellini@univr.it \\
       \name Alessandro Farinelli \email alessandro.farinelli@univr.it \\
       \addr Department of Computer Science, University of Verona,\\
       strada Le Grazie 15, 37135, Verona, IT}


\maketitle

\begin{abstract}
Partially Observable Markov Decision Processes (POMDPs) are a powerful framework for planning under uncertainty. They allow to model state uncertainty as a belief probability distribution.
Approximate solvers based on Monte Carlo sampling show great success to relax the computational demand and perform online planning.
However, scaling to complex realistic domains with many actions and long planning horizons is still a major challenge, and a key point to achieve good performance is guiding the action-selection process with domain-dependent policy heuristics which are tailored for the specific application domain. 
We propose to learn high-quality heuristics from POMDP traces of executions generated by any solver.
We convert the belief-action pairs to a logical semantics, and exploit data- and time-efficient Inductive Logic Programming (ILP) to generate interpretable belief-based policy specifications, which are then used as online heuristics.
We evaluate thoroughly our methodology on two notoriously challenging POMDP problems, involving large action spaces and long planning horizons, namely, rocksample and pocman.
Considering different state-of-the-art online POMDP solvers, including POMCP, DESPOT and AdaOPS, we show that learned heuristics expressed in Answer Set Programming (ASP) yield performance superior to neural networks and similar to optimal handcrafted task-specific heuristics within lower computational time. Moreover, they well generalize to more challenging scenarios not experienced in the training phase (e.g., increasing rocks and grid size in rocksample, incrementing the size of the map and the aggressivity of ghosts in pocman).
\end{abstract}





\section{Introduction}\label{sec:intro}
Partially Observable Markov Decision Processes (POMDPs) are a popular framework for modeling systems with (partial) state uncertainty \shortcite{cassandra2013incremental}, due, for instance, to inaccurate perception of autonomous agents (such as robots).
The uncertainty over the state is modeled as a probability distribution named \emph{belief}, which can be updated through \emph{observations}.
Unfortunately, the partial observability assumption significantly increases the computational effort for computing an optimal \emph{policy}, i.e., the best belief-action mapping, based on the current observations and inner state representation of the agent \shortcite{papadimitriou1987complexity}. 
In fact, optimality would require exploration of the full belief space, which is impossible, given the continuous belief distribution.
This is clearly incompatible with complex real-world scenarios involving large observation and state spaces, and requiring fast decision making.  
Modern computationally efficient POMDP solvers, as Partially Observable Monte Carlo Planning (POMCP) by \shortciteA{silver2010monte}, Determinized Sparse Partially Observable Tree (DESPOT) by \shortciteA{ye2017despot} and their extensions \shortcite{sunberg2018online,wu2021adaptive}, mitigate this issue, by adopting a particle filter to discretize the belief distribution and Monte Carlo simulations to compute a nearly optimal policy with high cumulative reward. 
However, the quality of the computed policy heavily depends on the size of the action space and the length of the planning horizon which, in turn, require many simulations and affect the computational burden.
For this reason, the performance of POMDP solvers is dramatically influenced by the availability of task-specific \emph{policy heuristics}, representing approximations of the optimal policy, which are used to focus the exploration on most promising actions, thus reducing the number of required online simulations.
Policy heuristics are relations mapping the belief of the agent to actions.
However, since they approximate the policy function, which is typically unavailable in (PO)MDPs, a significant effort to design accurate and efficient heuristics is required, even to domain experts.

In this work, we extend the methodology and empirical evaluation first presented by \shortciteA{mazzi2023learning} and propose to learn \emph{human-interpretable policy heuristics} from datasets of POMDP executions, combining basic commonsense domain knowledge and Inductive Logic Programming (ILP, \shortciteR{muggleton1991inductive}). 
Specifically, starting from example POMDP executions (generated without heuristics), we use ILP to discover logical specifications mapping actions to a higher-level representation of the belief, based on simple domain-specific human-level concepts (\emph{features}). 
Such concepts are typically present in the definition of the transition map or the reward for a specific POMDP problem, hence they can be easily provided by a user with basic knowledge of the domain.
We use the logical formalism of Answer Set Programming (ASP, \shortciteR{lifschitz1999answer}), a state-of-the-art method for representing and reasoning on a planning domain \shortcite{tran2022answer}, to represent learned logical specifications, and use them as policy heuristics in online POMDP solvers. Hence, from now on, we will refer interchangeably to \emph{(policy) specifications} and heuristics.

This work makes the following contributions to the state of the art:
\begin{itemize}
    \item we show how to learn \emph{interpretable logical belief-dependent policy heuristics} from POMDP executions generated with \emph{any off-the-shelf solver} (in this paper, POMCP), requiring \emph{little training data and time} with respect to state-of-the-art methods based, e.g., on neural networks for policy learning;
    \item we propose a method to integrate learned logical heuristics in several POMDP solvers, mainly POMCP, DESPOT and their most recent and efficient versions, AdaOPS by \shortciteA{wu2021adaptive} and POMCPOW by \shortciteA{sunberg2018online}. Our approach can take into account the \emph{confidence level} about learned specifications, depending on their adherence to the training data (or alternatively specified by the user, thanks to the interpretable formalism). This is used for \emph{soft policy guidance}, namely, weighted probabilistic action exploration with completeness guarantees, in POMCP, and for selecting the most probable action in DESPOT and AdaOPS;
    \item we empirically demonstrate our approach in two notably challenging POMDP benchmark domains, namely \emph{rocksample} and \emph{pocman}.
    These two domains present unique challenges for our methodology, compared to other available POMDP benchmarks.
    In fact, rocksample is typically the domain with the largest (discrete) action space \shortcite{wu2021adaptive}, while pocman \shortcite{ye2017despot} requires the longest planning horizon (i.e., the longest action sequence required to complete the task).
    Hence, in such scenarios the quality of policy heuristics is crucial to achieve good performance.
    We assess the \emph{generalization and scalability capabilities of learned specifications}, which are derived from small domains (e.g., rocksample on a small grid and with few rocks) and then applied to online POMDP planning in more challenging contexts (larger maps and belief spaces). We also perform an ablation study to investigate the role of specifications in different parts of Monte Carlo tree search, i.e., rollout and tree exploration. Finally, we discuss the role of the quality and amount of training data on learned policy heuristics and resulting planning performance;
    \item we make publicly available the code for experimental replication at \url{https://gitlab.com/dan11694/ilasp_pomdp}.
\end{itemize}
\noindent
Differently from our preliminary paper \shortcite{mazzi2023learning}, we introduce heuristics also in the most computationally demanding rollout phase of Monte Carlo tree search; we integrate our methodology with the DESPOT solver, while still learning from POMCP-generated traces, showing that training data can be derived from any POMDP solver; we replace the custom battery domain with a more challenging and famous benchmark, pocman, which presents much longer planning horizon (up to $\approx 85$ steps to complete the task); we make a more extensive empirical evaluation of our approach, including a more detailed discussion of the potential limitations of our method, and the comparison with other state-of-the-art POMDP solvers POMCPOW and AdaOPS, and neural architectures.

In Section \ref{sec:sota} we review recent research in heuristics learning and the fusion of (PO)MDP and logics, highlighting the differences with respect to our approach.
Then, in Section \ref{sec:background} we introduce relevant background and notation about POMDP, ASP and ILP, with a description of our benchmark domains as useful leading examples through the rest of the paper. We then present our methodology in Section \ref{sec:met}, and evaluate it thoroughly in Section \ref{sec:exp}. We finally discuss advantages and potential limitations of our approach in Section \ref{sec:discussion}, and summarize our results and possible future extensions in Section \ref{sec:conc}.

\section{Related Works}\label{sec:sota}
Exact computation of optimal POMDP policies is known to be PSPACE-complete \shortcite{papadimitriou1987complexity}, hence hardly scalable to large action, state and observation spaces.
This has often limited the practical application of POMDP solving algorithms to complex real-world scenarios, such as industrial automation, autonomous robotic navigation and medical diagnosis, which would actually benefit from the representational features of this approach \shortcite{cassandra1998survey}.
For this reason, in recent years several \emph{online} POMDP solvers have been proposed, which yield sufficiently reduced computational time for online deployment. Some of them exploit factorization of the state space \shortcite{paquet2005online} or task-based reduction \shortcite{shani2013task}, in order to reduce the search space and mitigate the computational burden. Currently, the most popular approach relies on Monte Carlo Tree Search (MCTS), which computes an estimate of the value of actions and states with multiple efficient simulations. In this way, the state-action search space is explored only locally (following simulations performed from the current state of the agent), thus mitigating the curse of dimensionality.
MCTS represents the foundation of two state-of-the-art online POMDP solvers, POMCP \shortcite{silver2010monte} and DESPOT \shortcite{ye2017despot}, and their more efficient extensions AdaOPS \shortcite{wu2021adaptive} and POMCPOW \shortcite{sunberg2018online}, which have been employed for challenging tasks, such as active visual search \shortcite{pomp2020bmvc,giuliari2021pomp++}, robotic navigation \shortcite{gupta2022intention} and multi-agent planning by \shortciteA{bhattacharya2021multiagent} and \shortciteA{amini2023pomcp}. We will discuss details of POMCP and DESPOT in the following sections, as they represent the algorithmic foundation for more advanced solvers.

Despite the success of online POMDP solvers even in real-world scenarios, their performance degrades significantly in complex environments with very large action and state spaces, or in tasks with long planning horizons, requiring many deep simulations.
Indeed, recent literature shows that integrating available knowledge into (PO)MDPs and Reinforcement Learning (RL), can significantly support the computation of efficient policies \shortcite{Castellini2019,zuccotto2022,zuccotto2022a,vinyals2019grandmaster,cheng2021heuristic}.
Such knowledge is either available in advance or, more conveniently, it can be learned as in \shortciteA{zuccotto2022}.
Existing solutions to incorporate domain-specific knowledge into POMDPs span from the use of neural networks and deep learning methodologies, to more classical graph and logical representations.
For instance, \shortciteA{zuccotto2022} proposed to learn state variable relationships in POMCP using Markov random fields, for application to robotic search.
\shortciteA{cai2021closing} proposed to combine DESPOT with heuristics learned offline as a neural network, and refined online from experience, for autonomous driving in crowds.
In the same scenario, \shortciteA{danesh2023leader} proposed an actor-critic architecture to learn the value function for MCTS, inspired by the success of AlphaZero and its variants \shortcite{silver2018general}.
A similar approach is used by \shortciteA{leemagic} to learn macro-actions for DESPOT, i.e., local trajectories in the action space.
In the more general area of learning generalizable policy heuristics from experience, \shortciteA{de2011scaling} proposed an approach based on binary decision trees, which however offers limited representational capabilities. \shortciteA{toyer2020asnets} relied on a neural network architecture to learn the heuristics, but assuming that the optimal behavior (i.e., the heuristics themselves) was actually encoded in the training dataset. Recently, \shortciteA{subramanian2022approximate} proposed an approach based on recurrent neural networks to learn efficient policies for POMDPs, particularly for rocksample, exploiting approximate information maximization and outperforming state-of-the-art neural architectures in the field, based on proximal policy optimization \shortcite{schulman2017proximal}.

Methods based on neural networks require many examples to learn from, and a significant amount of time and hardware capabilities \shortcite{cai2021closing}. Even when online learning is proposed, many observation points must be acquired before discovering robust and efficient policy heuristics (e.g., $> 10^5$ iterations in \shortciteR{leemagic}).
Moreover, neural networks generate hardly interpretable heuristics, thus providing scarce guarantees about the learned behavior. This may result in exploring bad policies, in case training information is not correct or good enough (as highlighted, e.g., by \shortciteR{cai2021closing}).
A more interpretable knowledge-based approach involves the use of finite state controllers for POMDPs \shortcite{andriushchenko2022inductive}, which are however learned online assuming the availability of oracles setting suitable bounds to exploration.
\shortciteA{maliah2022computing} define contingent planning sub-routines based on a logical formalization of POMDP actions, environmental features and the goal, in order to support the decision making process under uncertainty.
In general, symbolic and logical formalisms have the potential to increase the interpretability of AI systems, enhancing their diffusion and acceptance \shortcite{kambhampati2022symbols}.
Logical formalisms can also be applied to improve policy computation, though existing works are mainly related to the field of fully observable MDPs and semi-MDPs.
For instance, in the REBA framework by \shortciteA{sridharan2019reba} ASP is used to describe spatial relations between objects and rooms in a houseware domain, hence driving a robotic agent to choose a specific room to inspect, while solving simpler MDP problems locally. Similarly, \shortciteA{leonetti2016synthesis} propose DARLING, which uses ASP statements to bound MDP exploration in a simulated grid domain (similar to rocksample) and in a real service robotics scenario. 
Linear temporal logic is used by \shortciteA{de2019foundations,leonetti2012automatic} to drive exploration in MDPs.
Logical statements can help also avoid unwanted behaviors, e.g., in safety-critic scenarios. To this aim, \shortciteA{mazziAAMAS2021,mazzi2021rule,Mazzi2023b} refined rule templates to identify unexpected decisions and shield from unwanted actions, in the context of velocity regulation for a mobile robot. In \shortciteA{wang2021online}, a goal-constrained belief space containing only safely reachable states is defined with propositional logics, and a Satisfiability Modulo Theory (SMT) solver is used to guarantee proper execution of an houseware automation task. 

The aforementioned approaches however assume that logical statements are defined by the user.
This is often infeasible even in relatively simple (PO)MDP domains, since logical heuristics express knowledge related to the policy, which is unknown. Moreover, the wrong definition of logical relations may significantly impact on task performance.
For instance, \shortciteA{de2020imitation} propose to learn temporal logic specifications as finite automata from good example traces. However, rules are used to shape the reward in fully observable MDPs, hence the method strongly relies on the quality of training examples and induced formulas.  

In this paper, we propose to overcome some of existing limitations, learning logical policy heuristics from relatively few examples of POMDP execution (i.e., sequences of belief-action pairs), and using them to drive MCTS exploration, rather then define reward signals or hard constraints having a huge impact on planning performance. We will show that this allows to mitigate the impact of bad policy heuristics, especially in POMCP.
We choose the ASP formalism to represent policy heuristics. 
ASP is the state of the art for logical task planning \shortcite{erdem2018applications,ginesi2020autonomous,meli2021autonomous,tagliabue2022deliberation,tran2022answer,meli2023logic}, providing significantly richer syntax than other approaches, such as finite state controllers adopted by \shortciteA{andriushchenko2022inductive}.
Several solutions exist to learn domain knowledge expressed in ASP or other logical formalisms, e.g., inductive refinement of action theories and operators proposed by \shortciteA{gil:icml94,balduccini:aaaisymp07} and relational reinforcement learning by \shortciteA{mohan:ACS18,mota:aaaisymp20}.
In this paper, we adopt the framework of Inductive Logic Programming (ILP, \shortciteR{muggleton1991inductive}), specifically Inductive Learning of Answer Set Programs (ILASP, \shortciteR{law2018complexity}).
It has been recently used, e.g., to enhance comprehensibility of black-box models \shortcite{d2020towards,rabold2018explaining}, including moral value systems for autonomous agency \shortcite{Veronese2023BEWARE} and learn robotic task knowledge \shortcite{meli2020towards,bonet2020learning,rodriguez2021learning,meli2021inductive}. 
Recently, ILASP was also used to learn reward machines in online reinforcement learning by \shortciteA{furelos2021induction}.
In contrast to it, we here focus on offline heuristics discovery, rather than learning reward representations. Furthermore, our heuristics represent the belief-action mapping in POMDPs, hence they are based directly on the uncertain belief representation, rather than on observable states.

As evidenced in the following of this paper, ILASP offers considerable advantages with respect to other methods for policy heuristics learning. 
In fact, it requires significantly fewer examples ($< 1000$ traces of execution) and less training time (few minutes), with respect to approaches based on neural networks. Furthermore, it can easily incorporate commonsense domain knowledge, e.g., task-specific concepts, which allow to learn interpretable heuristics. As we will show in the next sections, commonsense concepts encode very simple information which can be directly retrieved from the POMDP problem definition itself, thus requiring minimal expertise. 
Another important advantage of our approach is that ILASP can discover high-quality generalizable heuristics even when the training examples are obtained in small-scale scenarios by a solver unscented of the optimal heuristics, differently from, e.g., \shortciteA{de2011scaling,toyer2020asnets}.

\section{Background}\label{sec:background}
In this section, we provide basic description and notation about the foundational parts of our methodology.
Specifically, we first introduce the general POMDP framework and describe the rocksample and pocman domains, which are two benchmark and notably challenging domains for the POMDP community \shortcite{ye2017despot,wu2021adaptive}.
Then, we describe POMCP and DESPOT algorithms for online POMDP solving, on which more advanced solvers as AdaOPS and POMCPOW are based.
Finally, we present ASP and ILASP, with clarifying examples from the rocksample domain.

\subsection{Partially Observable Markov Decision Processes}\label{sec:pomdp}
A Partially Observable Markov Decision Process (POMDP, \shortciteR{Kaelbling98}) is a tuple $(S, A, O, T, Z, R, \gamma)$,
where $S$ is a set of partially observable \emph{states};
$A$ is a set of \emph{actions};
$Z$ is a set of \emph{observations};
$T$:~$S\times A \rightarrow \Pi(S)$ is the \textit{state-transition model}, mapping to a probability distribution $\Pi(\cdot)$ over states;
$O$:~$S\times A \rightarrow \Pi(Z)$ is the \textit{observation model};
$R$ is the \textit{reward function} and $\gamma \in [0,1)$ is a \textit{discount factor}.
An agent must maximize the \emph{discounted return} $E[\sum_{t=0}^{\infty} \gamma^t R(s_t,a_t)]$.
The probability distribution over states $\pazocal{B} = \Pi(S)$, called \emph{belief}, is used to model uncertainty about the true state.
To solve a POMDP it is required to find a \emph{policy}, namely a function $\pi$:~$\pazocal{B} \rightarrow A$ that maps beliefs $\pazocal{B}$ into actions.
However, exact computation of the optimal policy requires representation of the full belief-action space, i.e., exploration of the full continuous belief distribution and evaluation of all possible actions under any circumstances. For this reason, exact solving of POMDPs is PSPACE-complete \shortcite{papadimitriou1987complexity}, and impractical even in relatively small domains of interest, as the benchmark ones presented hereafter.

We now show how our benchmark domains are represented as POMDPs.

\subsubsection{Rocksample Scenario}\label{sec:rs}
\begin{figure}
    \centering
    \begin{subfigure}{0.4\textwidth}
    \centering
    \includegraphics[width=0.9\linewidth]{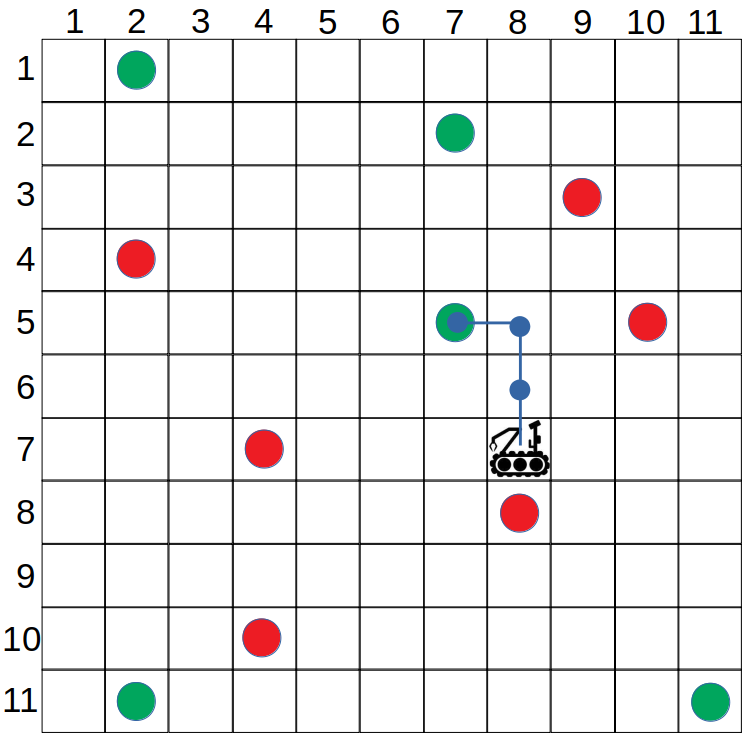}
    \caption{Rocksample\label{fig:setup_rocksmaple}}
    \end{subfigure}
    \begin{subfigure}{0.4\textwidth}
    \centering
    \includegraphics[width=0.9\linewidth]{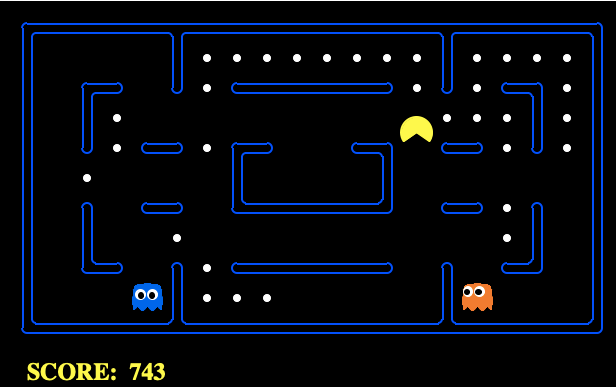}
    \caption{Pocman\label{fig:setup_pocman}}
    \end{subfigure}
    \caption{Example scenarios for our two case studies.}
    \label{fig:setup_rocksample}
\end{figure}
In the rocksample domain, an agent can move in cardinal directions on a $N \times N$ grid, one cell at a time, with the goal to reach and sample a set of $M$ rocks with known position on the grid. Rocks may be valuable or not. Sampling a valuable rock yields a positive reward (+10), while sampling a worthless rock yields negative reward (-10). 

The observable state of the system is described by the current position of the agent $p_c = (x_c, y_c)$ and each $i$-th rock $p_i = (x_i, y_i)$, while the value of rocks is uncertain before sampling. However, the agent can check one rock per time step and obtain an observation about its value, with an accuracy depending on the distance between the rock and the agent. Finally, the agent obtains a positive reward (+10) exiting the grid from the right-hand side. The belief then contains $\left\{Pr(r_i = 1)\right\}_{i=1\ldots M}$, i.e., the probability that each rock $r_i$ is valuable ($r_i = 1$).

An explanatory scenario with $N=11$ and $M=11$ is depicted in Figure \ref{fig:setup_rocksample}, where valuable and invaluable rocks are marked as green and red dots, respectively, and the agent is at location $(8,7)$ initially.
The rocksample task is particularly challenging because of the large number of actions. In fact, $\pazocal{A} = \{\stt{sample}, \stt{exit}, \stt{north}, \stt{south}, \stt{west}, \stt{east}, \stt{check}(i) \forall i=1\ldots M\}$, hence, $|\pazocal{A}| = 6 + M$. Moreover, the belief size increases with $M$ as well, while the state space depends also on the size $N$ of the map.

\subsubsection{Pocman Scenario}\label{sec:poc}
In the pocman domain, an agent (pocman) can move in cardinal direction (hence, 4 actions) in a grid world with $N$ cells, possibly presenting walls, with the goal to eat pellets of food, yielding a positive reward (+1). Each cell contains food pellets with probability $\rho_f$. $G$ ghost agents are also present in the grid, which normally move randomly, but may start chasing the pocman (with probability $\rho_g$) if they are within 2-cell distance from it. If the pocman clears the level, i.e., it eats all food pellets, a positive reward is yielded (+1000). If the pocman is eaten or hits a wall, it receives a negative reward (-100). Moreover, at each step the pocman gets a small negative reward (-1). 

The fully observable state contains coordinates of each $i$-th wall cell $p_{wi} = (x_{wi}, y_{wi})$ and the current position of pocman $p_c = (x_c, y_c)$. 
At each step, the pocman receives \emph{certain information} (observation) about surrounding environment within 2-cell distance. Specifically, within this range the agent can see or hear ghosts and smell food. Then, the agent builds two belief distributions, $\pazocal{B}_g = \left\{Pr(g_i = 1)\right\}_{i=1\ldots N}, \pazocal{B}_f = \left\{Pr(f_i = 1)\right\}_{i=1\ldots N}$, representing the probability that $i$-th cell contains a ghost ($g_i = 1$) or a food pellet ($f_i = 1$), respectively.

An explanatory scenario with $G=2$ is depicted in Figure \ref{fig:setup_pocman}.
Pocman is a particularly challenging task because of the very large dimension of the state and belief space ($\approx 10^{56}$ states, as reported by \shortciteR{ye2017despot}) and the long planning horizon (to clear the level, the agent must potentially explore the full environment to find all pellets, and spurious movements are often required to escape ghosts). Given the step penalization for pocman, it is then fundamental to optimize action selection to achieve good planning performance.

\subsection{Online Solvers for POMDPs}
We now describe the two main online POMDP solvers on which our methodology relies: POMCP and DESPOT.

\subsubsection{Partially Observable Monte Carlo Planning (POMCP)}\label{sec:pomcp}
POMCP \shortcite{silver2010monte} computes a sub-optimal policy based on Monte Carlo techniques.
The belief distribution is approximated using a \emph{particle filter}. More specifically, each time the agent receives an observation, it perceives a specific state realization $s_i$. A particle is then associated to each $s_i$, and the probability of each realization is $\frac{p_i}{\sum_{i=1}^{|S|} p_i}$, being $p_i$ the number of particles corresponding to $s_i$. In other words, the continuous belief distribution is approximated as a set of particles partitioned between multiple state realizations.
The particle filter can be initialized randomly or considering prior knowledge about the environment, as shown by \shortciteA{Castellini2019}.

POMCP then computes an online policy, i.e., the best action to be performed at each time step, combining the particle filter with MCTS.
Specifically, during execution, the history $h$ of explored belief-action pairs is stored.
At each step, a tree is built from the current state (root), branching across all possible actions.
Then, multiple simulations are performed, i.e., multiple actions are chosen, resulting in exploration of new future states, approximated via the particle filter.
A particle (representing a specific state) is selected from the filter and used as an initial point to perform a simulation in the Monte Carlo tree.
In order to reduce the number of simulations and exploit the particle filter efficiently, Upper Confidence Bound for Trees (UCT, \shortciteR{Kocsis2006}) is used as a search strategy to select most convenient subtrees to explore. In particular, UCT suggests to explore action $a$ which maximizes the action value:
\begin{equation} \label{eq:UCT}
V_{UCT}(ha) = V(ha) + c \cdot \sqrt{\frac{\log N(h)}{N(ha)}}
\end{equation}
where $V(ha)$ is the expected return achieved by selecting action $a$, $N(h)$ is the number of simulations performed from history $h$, $N(ha)$ is the number of simulations performed while selecting action $a$, and $c$ is known as the exploration constant.
UCT is used to balance the exploration of new actions (i.e., actions with low $N(ha)$) and the exploitation of effective actions (i.e., action with high $V(ha)$).
In case a leaf node is reached in the tree (i.e., $N(ha) = 0$), an action is chosen according to a default \emph{rollout} criterion (typically, randomly).
At the end of all simulations for a given step, only the action which leads to the highest discounted return is executed.

POMCP has the advantage to use simulations to explore only a specific part of the belief-action space, relevant from the current state of execution.
However, it fails to scale to very large belief spaces, or when many actions are available, since it would require a large amount of simulations in order to properly estimate the value of actions and closely approximate the optimal policy.
For this reason, in \shortciteA{silver2010monte} heuristics are still needed to achieve good performance in challenging scenarios as pocman and rocksample. Other solutions have also been proposed to mitigate the computational burden, e.g., by leveraging on model-specific structural properties \shortcite{katt2017learning} or existing state variable relationships to shrink the search space and require fewer simulations \shortcite{Castellini2019,zuccotto2022}, and parallel solving \shortcite{basu2021parallelizing}.

\subsubsection{Deterministic Sparse Partially Observable Trees (DESPOT)}\label{sec:despot}
DESPOT \shortcite{ye2017despot} tries to solve some POMCP limitations, by an intensive use of heuristics to drive the exploration in MCTS.
Assuming very good heuristics are available, relatively few but carefully selected \emph{scenarios}, i.e., Monte Carlo sub-trees, are explored by the algorithm and allow to converge to a nearly optimal policy.

In more detail, starting from a root belief node $b_0$ (represented as a set of particles), DESPOT approximates the belief-action tree in simulation by $K$ random scenarios, consisting of particle-observation-action sequences (each particle represents a specific state) in the form $\langle s_0, a_0, s_1, o_1, a_1, \ldots, s_n, o_n, a_n \rangle$, being $s_i$ a specific particle at a given belief node. 
Similarly to POMCP, the goal of DESPOT is to compute an optimal policy by building and exploring an approximate belief-action tree, executing a \emph{default policy} (equivalent to the rollout policy) if observations out of the $K$ scenarios are received.
Given a policy $\pi$, the algorithm estimates the value function $V_{\pi}(b_0)$ resulting at $b_0$ under $\pi$ as:
\begin{equation*}
    \hat{V}_{\pi}(b_0) = \frac{1}{K}\sum_{i=1}^K V_{\pi, \Phi_i}
\end{equation*}
being $\Phi_i$ the $i$-th scenario and $V_{\pi, \Phi_i}$ the discounted return achieved from $\pi$ in $\Phi_i$.
Clearly, $\hat{V}_{\pi}(b_0) \rightarrow V_{\pi}(b_0)$ as $K \rightarrow \infty$, which however is in conflict with online application.
On the other hand, choosing a finite (relatively low) value for $K$ induces the risk of \emph{overfitting} the policy on a limited set of scenarios.
In order to mitigate this issue, DESPOT performs \emph{regularization} on the size of the optimal policy, i.e., it maximizes $\hat{V}_{\pi}(b_0) - \lambda |\pi|$, with $\lambda \in \mathbb{R}^+$ and $|\cdot|$ denoting the length of $\pi$, i.e., the number of actions to be executed.

The key extension of DESPOT with respect to POMCP is the use of pre-defined lower and upper bounds on $\hat{V}_{\pi}(b_0)$, respectively $l(b_0)$ and $u(b_0)$, in order to build an useful and informative belief-action sub-tree and perform efficient \emph{anytime heuristics search}. 
The lower bound is computed as:
\begin{equation}\label{eq:lb_despot}
    l(b_0) = \hat{V}_{\pi_0}(b_0)
\end{equation}
\noindent
being $\pi_0$ the pre-defined \emph{default policy} used at leaf nodes.
In general, given $\epsilon(b) = u(b) - l(b)$, the goal of DESPOT is to incrementally reduce $\epsilon(b_0)$ to $\xi \epsilon(b_0), 0 < \xi < 1$, until $\epsilon(b_0) < \epsilon_0 \in \mathbb{R}^+$. In fact, this means that the lower and upper bound become more and more similar, thus the algorithm is \emph{more aware} about the current state and more confidently performs the best action.
More specifically, at each root node, the algorithm uses the upper bound to select an action $a^{\star}$ to be explored, as:
\begin{equation}\label{eq:ub_despot}
    a^{\star} = \argmax_{a\in A} u(b_0,a) = \argmax_{a\in A} \left\{\rho(b_0,a) + \sum_{b'} u(b') \right\}
\end{equation}
\noindent
being $\rho(b_0,a)$ the average (over all scenarios) discounted return achieved from $b_0$ executing $a$, and $b'$ any descendant node generated from $b_0$ by executing $a$.
Then, the observation branch to be explored is chosen in order to maximize:
\begin{equation}\label{eq:gap_despot}
    E(b') = \epsilon(b') - \frac{|\Phi(b')|}{K} \xi \epsilon(b_0)
\end{equation}
\noindent
where $|\Phi(b')|$ is the number of scenarios passing through $b'$. $E(b')$ represents the discrepancy between the current gap at $b'$ and the expected gap resulting from the reduction $\epsilon(b_0) \rightarrow \xi \epsilon(b_0)$. As proved by \shortciteA{ye2017despot}, this allows to actually converge faster.

As stated in the beginning of this section, one crucial requirement of DESPOT is the availability of good task-specific heuristics, including the default policy for the definition of the lower bound. This is a major limitation in complex real-world scenarios, where often domain knowledge is scarcely available.

\subsection{Answer Set Programming}\label{sec:asp}
An ASP program represents a domain of interest with a \emph{signature} and \emph{axioms} \shortcite{calimeri2020asp}. The signature is the alphabet of the domain, defining its relevant attributes as variables (with discrete ranges) and predicates of variables (\emph{atoms}).
For example, in the rocksample domain variables of interest are: rock identifiers \stt{R}$\in \{1, \ldots, M\}$, possible values of distances between agent and rocks \stt{D}$\in \mathbb{Z}$, and (discretized) probabilities computed from the belief \stt{V}$\in \{0, 1, \ldots, 100\}$\footnote{Since probabilities are normalized in $[0, 1]$, discrete probability values represent percentages.}.
Atoms typically represent environmental features and actions.
In rocksample, possible environmental features are \stt{guess(R,V)}, representing the discrete (percentage) probability \stt{V} that rock \stt{R} is valuable, and \stt{dist(R,D)}, representing the Manhattan distance \stt{D} between the agent and rock \stt{R};
possible action atoms are \stt{sample(R)} and \stt{east}, representing the action of sampling rock \stt{R} and moving east, respectively.

In ASP, a variable whose value is assigned is said to be \emph{ground} (e.g., \stt{R}=1). An atom is ground if its variables are ground (e.g., \stt{guess(1, 90)}, meaning that rock 1 is valuable with 90\% probability).
In particular, we define the \emph{Herbrand base} of an ASP program $P$ (or a set of ASP atoms), $\pazocal{H}_b(P)$, as the set of ground atoms which can be derived from $P$ (in other words, all possible variable assignments for atoms in $P$).
Axioms are logical relations between atoms. 
In this paper, we consider \emph{normal rules}, \emph{aggregate rules} and \emph{weak constraints}.
A \emph{normal rule} \stt{h :- b$_{1}$, \ldots, b$_n$} defines preconditions for grounding \emph{head} \stt{h} as the logical conjunction $\bigwedge_{i=1}^n$\stt{b$_i$} (\emph{body} of the rule).
For instance,
\begin{equation}
    \label{eq:asp_ex}
    \stt{sample(R) :- guess(R,V), V>60.}
\end{equation}
\noindent
means that rock \stt{R} can be sampled if the agent believes it is valuable with probability \stt{V}$>60\%$. Symbol \stt{:-} represents the logical implication ($\leftarrow$) in ASP.

An \emph{aggregate rule} expresses a choice over normal rules.
It has the form \stt{l \{h$_j$ : b$_{ij}$, \ldots, b$_{nj}$ \} u}, where symbol \stt{:} is equivalent to \stt{:-} and \stt{l,u}$\in \mathbb{N},\ \stt{l} \leq \stt{u}$.
Given the set $H$ of ground heads \stt{h}$_j$, according to respective body atoms \stt{b}$_{ij}$, the aggregate specifies that only a subset $H_l^u \subseteq H$ can be grounded, such that \stt{l} $\leq |H_l^u| \leq $ \stt{u}. 
For instance, if the following holds: 
\begin{equation}
    \label{eq:asp_agg}\stt{0 \{check(R) : guess(R,V), V>60\} 1.}
\end{equation}
and \stt{guess(0, 100)} and \stt{guess(1,90)} are ground atoms, then $H = \{ \stt{check(0), check(1)}\}$, but only three subsets of it can be actually grounded, namely $H_0 = \{\}$, $H_1 = \{ \stt{check(0)}\}$, $H_2 = \{ \stt{check(1)}\}$.

A specific choice among the possible ground sets resulting from an aggregate rule can be made following a \emph{weak constraint} in the form:
\begin{equation*}
    :\sim \stt{b}_1\stt{(V}_1, \ldots, \stt{V}_n\stt{)}, \ldots, \stt{b}_m\stt{(V}_1, \ldots, \stt{V}_n\stt{)}.\stt{[w@l, V}_1, \ldots , \stt{V}_n\stt{]}
\end{equation*}
where \stt{w} is the weight, \stt{l} an integer priority level (used if multiple weak constraints are specified), \stt{b}$_i$ are atoms and \stt{V}$_j$ variables. 
The weight can either be one variable among \stt{V}$_j$'s, or an integer. If the weight is a positive (respectively, negative) integer, the weak constraint means that grounding of atoms \stt{b}$_i$'s should be penalized (respectively, encouraged). If the weight is a variable, say \stt{V$_1$}, then the constraint means that lower values for \stt{V$_1$} should be preferred.
As an example, if axiom \eqref{eq:asp_agg} holds and the following weak constraint is specified:
\begin{equation}
    \label{eq:asp_th_wc}:\sim \stt{check(R), guess(R,V). [-V@1, R, V]}
\end{equation}
\noindent
$H_0 \leq H_2 \leq H_1$ will be the preference ordering among above allowed sets.

Given an ASP task description, an ASP solver computes all its \emph{answer sets}. An answer set is the minimal set of ground atoms satisfying axioms. 
Starting from an initial variable assignment, all ground atoms in the body of axioms are computed, hence ground head atoms are derived.
For instance, assuming an ASP program contains only axiom \eqref{eq:asp_ex} and the set of ground atoms \{\stt{guess(1,50), guess(2,70)}\}, then action \stt{sample(2)} will be grounded by the solver, and the unique answer set will contain all three ground atoms, i.e., \{\stt{guess(1,50), guess(2,70), sample(2)}\}.

\subsection{Inductive Logic Programming under the Answer Set Semantics}\label{sec:ilp}
An ILP problem $\pazocal{T}$ under the answer set semantics is defined as the tuple $\pazocal{T} = \langle B, S_M, E \rangle$, where $B$ is the \emph{background knowledge}, i.e. a set of atoms and axioms in ASP syntax (e.g., ranges of variables); $S_M$ is the \emph{search space}, i.e. the set containing all possible ASP axioms that can be learned; and $E$ is a set of \emph{examples}, i.e., a set of ground atoms (constructed, e.g., from traces of execution). 
Informally, the goal is to find a \emph{hypothesis}, i.e., a set $H \in S_M$ that explains as many as possible of the examples in $E$. To this end, we use state-of-the-art ILASP learner \shortcite{law2018complexity}, where examples are \emph{Context-Dependent Partial Interpretations} (CDPIs).
\begin{defn}[Partial interpretation]
\label{def:partial_int}
Let $P$ be an ASP program. A \emph{partial interpretation} of $P$ is defined as $e = \langle e^{inc}, e^{exc} \rangle$, where $e^{inc}$ is named \emph{included set}, i.e., a subset of ground atoms which can be part of an answer set of $P$; $e^{exc}$ is named \emph{excluded set}, i.e., a subset of ground atoms which are not part of an answer set of $P$.
\end{defn}
\begin{defn}[Context-dependent partial interpretation (CDPI)]
\label{def:CDPI}
A CDPI of an ASP program $P$ is a tuple $\langle e, C \rangle$, where $e$ is a partial interpretation of $P$ and $C$ is a set of atoms called \emph{context}.
\end{defn}
\noindent
In this paper, partial interpretations contain atoms for actions, while the context involves environmental atoms. 
In this way, policy specifications (representative of the belief-action map) can be learned.

We can now formalize the ILASP problem considered in this paper, hence the properties to be satisfied by $H$:
\begin{defn}[ILASP task with CDPIs]
\label{def:ILASP_CDPI}
An ILASP learning task with CDPIs is a tuple $\pazocal{T} = \langle B, S_M, E \rangle$, where $E$ is a set of CDPIs such that:
\begin{equation*}
     \forall e = \langle \langle e^{inc}, e^{exc} \rangle, C \rangle \in E : \ B \cup H \cup C \models e^{inc} \land \ B \cup H \cup C \not\models e^{exc}
\end{equation*}
\end{defn}
\noindent
In other words, ILASP finds axioms which guarantee that actions in $e^{inc}$, observed in the examples, can be executed (i.e., can be grounded in an answer set), while unobserved actions in $e^{exc}$ cannot, given the context set of environmental features\footnote{This section, and specifically Definition \ref{def:ILASP_CDPI}, ignores the concept of \emph{negative examples} in ILP \shortcite{law2018complexity}, i.e., examples where $e^{inc}$ contains a set of \emph{forbidden} ground atoms. In fact, in this paper we extract examples from POMDP executions, hence only from \emph{observed behavior}. A negative example, on the contrary, allows to learn \emph{hard constraints}, i.e., conditions that shall never be verified. Typically, negative examples are provided by domain experts, hence they are not considered for the scope of this work.}.
In addition, ILASP finds the \emph{minimal} hypothesis $H$, i.e., axioms with the least number of atoms explaining examples. 
This may contrast with the accuracy of specifications, but increases their interpretability. Moreover, ILASP algorithm \shortcite{law2023conflict} solves the learning problem as an instance of conflict-driven constraint learning \shortcite{marques2021conflict}, which tries to balance between these two factors.
ILASP can also learn weak constraints from \emph{ordered CDPIs} \shortcite{LRB16}, i.e., partial interpretations with pre-defined preference values $pr$, expressed in the form $\langle \langle e^{inc}, e^{exc} \rangle, C \rangle @ pr$.

Finally, ILASP finds the hypothesis which explains \emph{most} of the examples, i.e., there may be CDPIs which are not entailed by $H \cup B$. When a hypothesis is found, ILASP also returns the number of not covered CDPIs, which can be used to quantify the level of confidence of $H$ with respect to a given example set.

\section{Methodology}\label{sec:met}
\begin{figure}
    \centering
    \begin{subfigure}{0.3\textwidth}
    \includegraphics[width=0.9\linewidth]{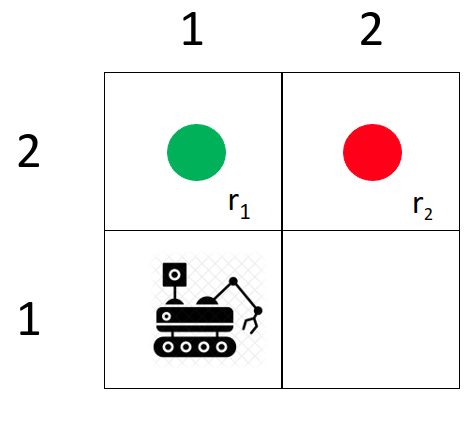}
    \caption{$t=1$: \stt{check(1)}.\label{fig:leading_ex}}
    \end{subfigure}
    \begin{subfigure}{0.3\textwidth}
    \includegraphics[width=0.9\linewidth]{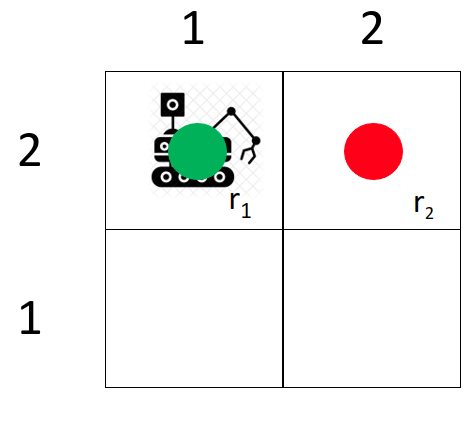}
    \caption{$t=2$: \stt{north}.\label{fig:leading_ex_2}}
    \end{subfigure}
    \begin{subfigure}{0.3\textwidth}
    \includegraphics[width=0.9\linewidth]{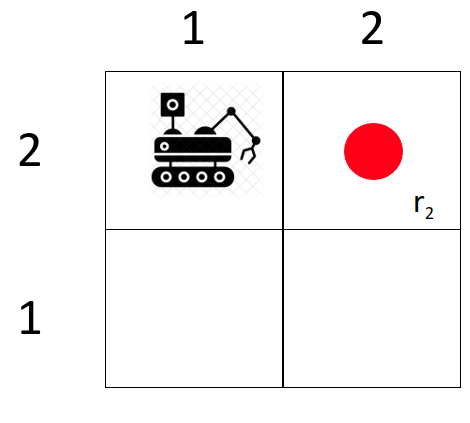}
    \caption{$t=3$: \stt{sample(1)}.\label{fig:leading_ex_3}}
    \end{subfigure}
    \caption{Simplified rocksample scenario with $N=M=2$.}
    \label{fig:leading_exs}
\end{figure}
We now describe our methodology for learning policy-related logical (ASP) specifications from POMDP traces of execution.
Our goal is to find interpretable policy heuristics encoded in the traces of execution, which consist of belief-action pairs.
As an example, in the rocksample domain, the handcrafted policy heuristics by \shortciteA{silver2010monte} is:
\begin{quote}
    \emph{Either check a rock whenever bad observations were received and it has been measured few times ($<5$); or sample a rock if the agent is at its location and collected observations are more positive (good value) than bad; or move towards a rock with more good than bad observations.}
\end{quote}
It relates the execution of specific actions to the \emph{observations} received by the agent. On the contrary, we want to discover policy heuristics \emph{depending on the belief distribution}.
To this aim, three steps are required:
\begin{itemize}
    \item ASP formalization of the POMDP problem;
    \item generation of ILASP examples from execution traces;
    \item definition of the ILASP task.
\end{itemize}
\noindent
In the following, we explain the phases of our pipeline, with reference to the clarifying example of a simple rocksample domain in a $2\times 2$ ($N=2$) grid with $M=2$ rocks (Figure \ref{fig:leading_exs}). Finally, we show how to use learned policy specifications for online POMDP solving in POMCP and DESPOT.

\subsection{ASP Formalization of the POMDP Problem}\label{sec:met_asp}
The POMDP problem is originally formalized as in Section \ref{sec:pomdp}. 
In order to represent it in ASP formalism, we need to define a set of high-level environmental features, which represent basic concepts about the belief. We denote the set of environmental features as \stt{$\pazocal{F} = \{$F$_i\}$}.
In the rocksample domain, such concepts depend on the variables introduced in Section \ref{sec:asp} and include\footnote{With respect to the work presented by \shortciteA{mazzi2023learning}, we omit \stt{min\_dist(R), best\_guess(R)}, representing higher-level concepts about the task which cannot be directly retrieved from the basic POMDP definition. In this way, we learn slightly different policy specifications, which are efficient for online planning and make our approach more generalizable to other domains.}:
\begin{itemize}
    \item \stt{guess(R,V)}, representing the probability \stt{V} that rock \stt{R} is valuable;
    \item \stt{dist(R,D)}, representing the Manhattan distance \stt{D} between the agent and rock \stt{R};
    \item \stt{delta\_x(R,D)} and \stt{delta\_y(R,D)}, representing the relative position \stt{D} along $x$ and $y$ axis of rock \stt{R} with respect to the agent;
    \item \stt{sampled(R)}, meaning that rock \stt{R} has been sampled;
    \item \stt{num\_sampled(V)}, meaning that a percentage \stt{V} of rocks has been sampled.
\end{itemize}
\noindent
As stated at the beginning of this paper, environmental atoms represent simple domain concepts which appear in the definition of the transition and reward maps for the POMDP problem, hence they can be easily defined by the user. For instance, \stt{sampled(R)} is the consequence of the sampling action on rock \stt{R}, while \stt{guess(R,V)} is updated after the agent senses (checks) rock \stt{R}.
Similarly, we must define the set of ASP atoms $\pazocal{A} = \{$act$_i\}$ representing the action space.
In the rocksample domain, they are \stt{sample(R), check(R), east, north, south, west}, with straightforward interpretation.

\subsection{Generation of ILASP Examples}\label{sec:met_ilasp}
Given the ASP formalization of the problem, we can represent the traces of execution accordingly, in order to extract examples (i.e., CDPIs) for ILASP.
Traces of execution are sequences of belief-action pairs.
Hence, we need to lift POMDP belief and actions to the equivalent ASP representation.
Specifically, the belief is lifted using the \emph{feature map} $F_\pazocal{F} : \pazocal{B} \rightarrow \pazocal{H}_b(\pazocal{F})$.
For instance, in the simplified rocksample scenario of Figure \ref{fig:leading_ex}, assume that 100 particles are available to approximate the belief. At the initial time $t=1$ of the task, particles are distributed uniformly among rocks, thus in 50 particles $r_1$ is valuable and in 50 particles $r_2$ is valuable. Hence, the belief $\Bar{b}$ can be represented as $\left\{Pr(r_1=1) = Pr(r_2=1) = 50\%, p_c=(1,1), p_1=(1,2), p_2=(2,2)\right\}$.
This can be lifted, using the feature map, as \stt{$F_{\pazocal{F}}(\Bar{b}) = \{$guess(1,50), guess(2,50), dist(1,1), dist(2,2), delta\_x(1,0), delta\_x(2,1), delta\_y(1,1), delta\_y(2,1), num\_sampled(0)$\}$}.

Similarly, we can define an \emph{action map} $F_\pazocal{A} : A \rightarrow \pazocal{H}_b(\pazocal{A})$.
For instance, if the action $\Bar{a}$ of the agent in Figure \ref{fig:leading_ex} is to check $r_1$, \stt{$F_{\pazocal{A}}(\Bar{a}) = \{$check(1)$\}$}.
In this way, we can build the following CDPI corresponding to the specific situation depicted in Figure \ref{fig:leading_ex}:
\begin{align}
    \label{eq:ex}&\langle \langle F_\pazocal{A}(\Bar{a}), \pazocal{H}_b(\Bar{a}) \setminus F_\pazocal{A}(\Bar{a}) \rangle, F_\pazocal{F}(\Bar{b}) \rangle\\\nonumber
    =& \langle \langle \{\stt{check(1)}\}, \{\stt{check(2)}\} \rangle, \{\stt{guess(1,50), guess(2,50), dist(1,1), dist(2,2),}\\\nonumber
    &\ \ \ \ \stt{delta\_x(1,0), delta\_x(2,1), delta\_y(1,1), delta\_y(2,1), num\_sampled(0)}\} \rangle
\end{align}
\noindent
The included set  $e^{inc}$ ($F_\pazocal{A}(\Bar{a})$) contains the observed action per each time step, while the excluded set $e^{exc}$ ($\pazocal{H}_b(\Bar{a}) \setminus F_\pazocal{A}(\Bar{a})$) contains all other possible not observed realizations for that action. In this way, ILASP can learn non only \emph{what to do} (given the current context), but also \emph{which actions to avoid}.
For the same reason, we also define CDPIs in the form:
\begin{align}
    \label{eq:cex}&\langle \langle \emptyset, \pazocal{H}_b(a) \rangle, F_\pazocal{F}(\Bar{b}) \rangle, \forall a \in A \setminus \{\Bar{a}\}\\\nonumber
    =&\langle \langle \emptyset, \{\stt{sample(\_)}\} \rangle, \{\stt{guess(1,50), guess(2,50), dist(1,1), dist(2,2), }\ldots\} \rangle\\\nonumber
    \cup &\langle \langle \emptyset, \{\stt{east}\} \rangle, \{\stt{guess(1,50), guess(2,50), dist(1,1), dist(2,2), }\ldots\} \rangle\\\nonumber
    \ldots&
\end{align}
\noindent
for all not executed (i.e., not observed) actions.

\subsection{Definition of the ILASP Task}\label{sec:met_ilasp_learn}
We can now set up the ILASP problem of Definition \ref{def:ILASP_CDPI}, defining the search space $S_M$ and the background knowledge $B$.
$B$ contains definitions of variables and constants, as reported in Section \ref{sec:asp}.
The search space definition is crucial for the computational performance of ILASP \cite{law2018complexity}, since more candidate hypotheses need to be explored. In this paper, we are interested in policy specifications following the map $\Gamma : \pazocal{F} \rightarrow \pazocal{A}$, which lifts the classical policy map $\pi : \pazocal{B} \rightarrow A$ to the ASP representation formalism. Hence, in $S_M$ we only consider normal axioms in the form: 
\begin{equation}
    \label{eq:ss}
    \bigwedge_{i=1}^n f_i \rightarrow a, \forall a \in \pazocal{A}
\end{equation}
being $f_i, \ldots, f_n \in \pazocal{F} \cup \{\stt{X} \lesseqgtr x\}$, where \stt{X} is any variable with possible integer constant value $x$ (e.g., \stt{D, V} in rocksample, for distances and probabilities).
In this way, ILASP can find lower and upper bounds to quantify the semantic information encoded in atoms as \stt{dist(R,D), guess(R,V), delta\_x(R,D), delta\_y(R,D), num\_sampled(V)}.
To define the search space, we use the mode bias syntax provided by ILASP, which allows to specify the possible head and body atoms in rules of $S_M$. We then prune axioms where an integer variable \stt{X} (\stt{D, V}) appears, but $\stt{X} \lesseqgtr x$ is missing. In fact, a rule as \stt{east :- dist(R,D)} is not meaningful until the possible values of \stt{D} are specified by $\stt{D} \lesseqgtr d$. In this way, we also reduce the number of axioms in $S_M$, which improves the performance of conflict-driven constraint learning in ILASP. 
More details about the definition of the search space can be found in the experimental section and the public repository.

In Equation \eqref{eq:ss}, we assume that the axioms for each action are \emph{independent} of other actions (i.e., other actions cannot appear in the body of axioms). This is a reasonable and not restricting assumption, since all POMDP actions have an influence on the state representation (hence, the environmental features) through the transition map. 
For instance, in the scenario depicted in Figure \ref{fig:leading_ex}, where $Pr(r_1=1) = Pr(r_2=1) = 50\%$, the agent decides to check $r_1$ at time $t=1$. Then, at $t=2$, the particle distribution (hence the probability of rocks to be valuable) may be updated such that $Pr(r_1=1) = 80\%$, $Pr(r_2=1) = 50\%$.
As a consequence, the agent may choose to move north to $r_1$ (Figure \ref{fig:leading_ex_2}), in order to sample it at $t=3$ (Figure \ref{fig:leading_ex_3}).
This results in the following CDPIs:
\begin{align}
    &\label{eq:ex2} \langle \langle \{\stt{north}\}, \{\stt{}\} \rangle, \{\stt{guess(1,80), guess(2,50), dist(1,1), dist(2,2),}\\\nonumber
    &\ \ \ \ \stt{delta\_x(1,0), delta\_x(2,1), delta\_y(1,1), delta\_y(2,1), num\_sampled(0)}\} \rangle\\\nonumber
    &\langle \langle \emptyset, \{\stt{check(\_)}\} \rangle, \{\stt{guess(1,80), guess(2,50), dist(1,1), dist(2,2), }\ldots\} \rangle\\\nonumber
    &\ldots
\end{align}
namely, \stt{north} action is executed in the context holding at $t=2$;
and:
\begin{align}
    &\label{eq:ex3} \langle \langle \{\stt{sample(1)}\}, \{\stt{sample(2)}\} \rangle, \{\stt{guess(1,80), guess(2,50), dist(1,0), dist(2,1),}\\\nonumber
    &\ \ \ \ \stt{delta\_x(1,0), delta\_x(2,1), delta\_y(1,0), delta\_y(2,0), num\_sampled(0)}\} \rangle\\\nonumber
    &\langle \langle \emptyset, \{\stt{check(\_)}\} \rangle, \{\stt{guess(1,80), guess(2,50), dist(1,1), dist(2,2), }\ldots\} \rangle\\\nonumber
    &\ldots
\end{align}
namely, \stt{sample(1)} action is executed in the context holding at $t=3$;
These examples may be explained by the following ASP theory:
\begin{align*}
    &\stt{sample(R, t) :- north(t-1), delta\_y(R, D, t-1), D}\geq \stt{1.}\\
    &\stt{north(t) :- check(R, t-1).}
\end{align*}
namely, rock \stt{R} should be sampled after moving north, if rock \stt{R} was at $y$-distance \stt{D}$\geq 1$ from the agent at the previous time step \stt{t-1}, and moving north should happen after checking \stt{R}.
The axioms above relate each action to the previous one, introducing an additional variable \stt{t} representing the time step of execution.
However, since \stt{check(1)} at $t=1$ modifies the probability of $r_1$ to be valuable from \stt{guess(1,50)} to \stt{guess(1,80)}, and \stt{north} at $t=2$ determines \stt{dist(1,0)} (according to the transition map of the POMDP problem), the following equivalent axioms are derived:
\begin{align*}
    &\stt{sample(R) :- dist(R, D), D}\leq \stt{0.}\\
    &\stt{north :- guess(R, V), V} \geq \stt{80, delta\_y(R, D), D > 0.}
\end{align*}
\noindent
This allows to consider single CDPIs separately, looking for policy specifications which match the context and the action \emph{at the same time step}, rather than searching for temporal relations between actions.
As a consequence, we are able to set up separate ILASP problems, one for each task action $a$, with $S_M$ containing axioms in the form of Equation \eqref{eq:ss}, but referred only to $a$. This significantly reduces the size of the search space and the computational performance of ILASP.

Finally, since POMDP traces contain only information about executed actions, there may occur time steps where other actions were possible, but less convenient and hence discarded.
Then, we also want to learn weak constraints encoding preferences between multiple feasible actions.
To this aim, assume that a normal axiom $\pazocal{R}$ is learned for action $a\in \pazocal{A}$, and $a^* \in \pazocal{H}_b(a)$ is the executed (ground) instance of $a$ at a given time step under belief $b$ (thus, it appears in $e^{inc}$ of a CDPI in the form of Equation \eqref{eq:ex}).
We generate CDPIs in the form:
\begin{equation*}
    \langle \langle \{\Bar{a}\}, \emptyset \rangle, F_{\pazocal{F}}(\Bar{b}) \rangle\ @ pr_{\Bar{a}} \ \ \forall \Bar{a} \in \pazocal{H}_b(\pazocal{R} \cup F_{\pazocal{F}}(\Bar{b}))
\end{equation*}
namely, the included set contains ground actions $\Bar{a}$ which are entailed by $\pazocal{R}$, given the initial grounding $F_{\pazocal{F}}(\Bar{b})$.
Each CDPI comes with its preference value $pr_{\Bar{a}}$ (see Section \ref{sec:ilp}), which is set higher for $a^*$ (which was actually executed). 
In other words, for each action, we generate CDPIs corresponding to alternative task realizations where another feasible action could have been executed, according to policy specifications discovered by ILASP. We then assign a preference weight to CDPIs, in order to penalize non observed realizations with respect to the actually executed action in the recorded trace.

For instance, in our simplified rocksample scenario, the rule $\pazocal{R}$ for $a=$\stt{check(R)} is:
\begin{equation}
    \label{eq:normal_check}\stt{check(R) :- guess(R, V), V}\leq \stt{50.}
\end{equation}
\noindent
which would ground both \stt{check(1), check(2)} at $t=1$ (Figure \ref{fig:leading_ex}). However, only the former is executed, hence $a^*=\stt{check(1)}$.
We then build the following CDPIs 
\begin{subequations}
\begin{align*}
    &\langle \langle\{\stt{check(1)}\}, \{\} \rangle, \{\stt{guess(1,50), guess(2,50), dist(1,1), }\ldots \rangle @ pr_1\\
    \cup&\langle \langle\{\stt{check(2)}\}, \{\} \rangle, \{\stt{guess(1,50), guess(2,50), dist(1,1), }\ldots \rangle @ pr_2
\end{align*}
\end{subequations}
\noindent
with $pr_1 > pr_2$. 

To learn preferences, we set up another ILASP task with ordered examples as in \shortciteA{LRB16}, with a search space (defined via mode bias) in the form:
\begin{equation}\label{eq:ss_weak}
    :\sim\ \stt{f}_1\stt{(V}_1, \ldots, \stt{V}_m\stt{)}, \ldots, \stt{f}_n\stt{(V}_1, \ldots, \stt{V}_m\stt{)}, \stt{a(V}_1, \ldots, \stt{V}_m\stt{)}.\stt{[w@pr, V}_1, \ldots , \stt{V}_m\stt{]}\ \forall a \in \pazocal{A}
\end{equation}
where $f_i, \ldots, f_n \in \pazocal{F}$. $\stt{V}_1, \ldots \stt{V}_m$ are ASP variables, \stt{pr} is the priority level and \stt{w} is any integer ASP variable (e.g., \stt{D,V}).
In our example scenario, this leads to learning the following weak constraint:
\begin{equation*}
    :\sim \stt{check(R), dist(R, D). [D@1, R, D]}
\end{equation*}
meaning that the agent prefers sensing the closest rock (with the lowest weight \stt{D}), if two rocks have the same probability to be valuable ($\leq 50$ from Equation \eqref{eq:normal_check}).

\subsection{Logical Policy Specifications for Online POMDP Solving}
ILASP generates policy specifications which can be used to drive the decision-making process of an online POMDP agent.
In the following, we show how the set of ILASP axioms can be integrated as heuristics in POMCP and DESPOT. The generalization to their extensions (e.g., AdaOPS by \shortciteR{wu2021adaptive} and POMCPOW by \shortciteR{sunberg2018online}) is straightforward, since the underlying solving principle is analogous.
With reference to our leading example in Figure \ref{fig:leading_exs}, we obtain the following set of axioms for the observed actions:
\begin{align}
\label{eq:learned_simple}
    &\stt{0 \{check(R) : guess(R, V), V} \leq \stt{50\} 2.}\\\nonumber
    &:\sim \stt{check(R), dist(R, D). [D@1, R, D]}\\
    &\stt{sample(R) :- dist(R, D), D}\leq \stt{0.}\\\nonumber
    &\stt{north :- guess(R, V), V} \geq \stt{80, delta\_y(R, D), D}\geq \stt{1.}
\end{align}
The first aggregate axiom states that multiple rocks may be sensed (at most 2 in our scenario), if they are valuable with probability $\leq 50\%$. However, the weak constraint suggests to check only the closest one (weight \stt{D} to be minimized).
Other axioms specify that a rock should be sampled when the distance from it is null, and the agent should move north towards a rock \stt{R} which is at positive $y$-distance ($\stt{D}\geq 1$) and is valuable with probability $\stt{V}\geq 80\%$.
Percentages of covered examples $cov$, i.e., CDPIs entailed by learned axioms, are equal to 66\%, 100\% and 100\% for \stt{check(R), sample(R), north}, respectively.

\subsubsection{Policy Specifications in POMCP}\label{sec:met_pomcp}
As explained in Section \ref{sec:pomcp}, POMCP exploration of the belief-action space is performed with MCTS, specifically maximizing the value of the chosen action via UCT in Equation \eqref{eq:UCT}, or selecting the next action randomly in the rollout phase when a leaf node is encountered in the tree.
We use learned policy heuristics in both phases of the exploration.

Specifically, at each root node in MCTS, we ground environmental features from the current belief according to map $F_\pazocal{F}$.
For instance, in Figure \ref{fig:leading_ex} corresponding to belief $\Bar{b} = \left\{Pr(r_1=1) = Pr(r_2=1) = 50\%, p_c=(1,1), p_1=(1,2), p_2=(2,2)\right\}$, the following atoms are grounded: \stt{$F_{\pazocal{F}}(\Bar{b}) = \{$guess(1,50), guess(2,50), dist(1,1), dist(2,2), delta\_x(1,0), delta\_x(2,1), delta\_y(1,1), delta\_y(2,1), num\_sampled(0)$\}$}.
According to learned specifications in Equation \eqref{eq:learned_simple}, \stt{check(1)} can be grounded.
Only for ground actions, we introduce a prior in UCT, increasing the value $V(ha)$ to the same value of $c$, and $N(ha)$ to a fixed high value (this is domain-dependent, we empirically set it to $10$ in our domains). This is the same as implying that we have already performed $N(ha)$ simulations in the current belief, and they achieved a good discounted return.
When we reach a non-root node, the belief is assumed invariant with respect to the root node, hence environmental features corresponding to the unobservable state (\stt{guess(R,V)} in rocksample) are kept the same, while the ones corresponding to the observable state (\stt{dist(R,D), delta\_x(R,D), delta\_y(R,D), num\_sampled(V)}) are updated according to the transition map during the simulation.
This implies that the belief distribution is assumed invariant during the simulation process, which may affect the optimality of decision making in highly dynamic scenarios, where the environment may significantly change while the agent is still making its decision. However, as we will empirically show in Section \ref{sec:exp}, this does not significantly impact POMCP performance. In fact, after the optimal action is selected, the simulation process starts again at the subsequent time step, thus the belief is updated at each time step and consequently, so does the grounding of policy specifications.

When a leaf (unexplored) node is reached, the rollout step is invoked.
Differently from \shortciteA{mazzi2023learning}, where we only adopted policy heuristics in UCT, we now exploit learned heuristics also in the rollout phase.
Similarly to UCT, we first ground environmental features and generate ground (suggested) actions.
We then define a discrete distribution over all actions, i.e., in the scenario of Figure \ref{fig:leading_exs}, $\pazocal{A}=$\{\stt{east, north, south, west, check(1), check(2), sample}\}.
For each action, the weight in the probability distribution is set according to the outcome of learned policy specifications and the coverage percentages.
Specifically, for each action $a_i \in \pazocal{A}$ we define the probabilistic weight as:
\begin{equation}\label{eq:prob_rollout}
    \rho_i = 
    \begin{cases}
        cov_i\ &\text{if } a_i \in \pazocal{H}_b(\pazocal{R} \cup F_{\pazocal{F}}(\Bar{b}))\\
        \min \{cov_j\}_{ij1}^{|\pazocal{A}|}\ &\text{otherwise}
    \end{cases}
\end{equation}
\noindent
$\pazocal{R}$ being the set of learned specifications (Equation \eqref{eq:learned_simple}). 
In other words, if an action $a_i$ is entailed by ASP specifications $\pazocal{R}$ under the current belief interpretation $F_{\pazocal{F}}(\Bar{b})$, its selection probability depends on the percentage of covered examples $cov_i$.
Otherwise, the minimum $cov_i$ is assigned to not entailed actions.
Then, the probabilistic weight for each action is normalized as $\frac{\rho_i}{\sum_i \rho_i}$ and the rollout procedure chooses a random action according to these probabilities. 
In our example, in Figure \ref{fig:leading_ex}, \stt{check(1)} is the only suggested action and has the minimal coverage percentage, thus $\rho_{i} \approx 12\% \ \forall a_i \in \pazocal{A}$. Hence, all actions have equal probability to be selected, similarly to the classical uninformed rollout.
However, in Figure \ref{fig:leading_ex_2}, actions \stt{north} and \stt{check(1)} are grounded, hence selection probabilities are $\rho_{i \neq \stt{north}} \approx 12\%, \ \rho_{\stt{north}} \approx 16\%$, with a bias towards the action with the highest confidence level (i.e., \stt{north}). In Figure \ref{fig:leading_ex_3}, \stt{sample(1)} and \stt{check(2)} are entailed at $t=3$, thus $\rho_{i \neq \stt{sample}} \approx 12\%, \ \rho_{\stt{sample}} \approx 16\%$, with a bias toward the sampling action.

We remark that we use learned ASP axioms for \emph{soft policy guidance}.
In fact, at non-leaf nodes, branches corresponding to actions not entailed from ASP heuristics can still be explored by MCTS, since their probability is $\rho_i > 0$. This preserves the asymptotic optimality of POMCP (i.e., for infinite simulations).
In Section \ref{sec:discussion}, we will show empirically that this mitigates the impact of bad policy heuristics learned by ILASP.

\subsubsection{Policy Specifications in DESPOT}\label{sec:met_despot}
As mentioned in Section \ref{sec:despot}, DESPOT performs anytime heuristics search in the sub-tree generated by $K$ random scenarios approximating the belief-action tree. The fundamental ingredient of DESPOT is the definition of suitable lower ($l(b_0)$) and upper ($u(b_0)$) bounds on the value of the current root node $b_0$ in the tree. 
These bounds are used to drive the exploration process, hence they must be carefully handcrafted for each domain, encoding task knowledge in order to achieve good performance \shortcite{ye2017despot}. 
In fact, the best action branches are chosen according to Equation \eqref{eq:ub_despot}, marginalizing over $u(b_0, a)$. Similarly, the exploration of the algorithm is driven by the minimization of $E(b')$, as expressed in \eqref{eq:gap_despot}. Since $\epsilon(b_0) = u(b_0) - l(b_0)$, choosing a good lower bound (i.e., close enough to the upper bound) can significantly drive DESPOT towards most promising branches in the belief tree, speeding up the exploration process until the termination condition $\epsilon(b_0) < \epsilon_0$ is met.

As explained by \shortciteA{ye2017despot}, $u(b_0)$ can be computed with approximate \emph{hindsight optimization} \shortcite{yoon2008probabilistic}, considering the \emph{deterministic problem} with the initial node corresponding to the initial state $s_{\Phi_i, 0}$ of each $\Phi_i$ of the $K$ scenarios, then evaluating:
\begin{equation}
    \label{eq:hind}
    u(b_0) = \frac{1}{K} \sum_{i=1}^{K} u(s_{\Phi_i, 0})
\end{equation}
\noindent
where $u(s)$ is a \emph{state-dependent} upper bound, often easier to define.
For instance, in the pocman scenario:
\begin{equation}
    \label{eq:pocman_hind}
    u(s) = \sum_{f_i}\gamma ^{1+d(f_i)} R(f_i) + \gamma ^ {1 + \max_{f_i} d(f_i)} R_{end}
\end{equation}
\noindent
where $R(f_i)$ is the reward for eating food pellet $f_i$ (+1), $R_{end}$ is the reward for completing the level (+100) and $d(f_i)$ is the distance between the pocman agent and a food pellet.
It is evident that the upper bound does not represent a belief-action map as the one we want to learn in this paper, but rather an \emph{optimistic evaluation of the maximum achievable return from the current state}. Hence, the domain knowledge required for defining an upper bound is very basic, and it is related to the definition of the specific POMDP problem, e.g., the reward function. On the contrary, the lower bound is typically defined as the value achieved by a \emph{default policy} $\pi_0$, thus we can exploit the learned heuristics for its definition. However, there is a fundamental difference between our default policy and the ones proposed for DESPOT \shortcite{ye2017despot}. In fact, typically $l(b_0)$ is computed similarly to Equation \eqref{eq:hind} as:
\begin{equation*}
    l(b_0) = \frac{1}{K} \sum_{i=1}^{K} V_{\pi_0}(s_{\Phi_i, 0})
\end{equation*}
\noindent
In this way, it is possible to define a policy based only on \emph{observable states} (specific belief realizations). 

In contrast, ILASP learns policy specifications based on the \emph{semantic interpretation} of the current belief (through the feature map $F_{\pazocal{F}}$). 
Specifically, at each belief node $b_0$, we compute the ground environmental atoms through map $F_{\pazocal{F}}$, and compute grounded actions according to policy specifications. Then, similarly to the rollout phase in POMCP explained in Section \ref{sec:met_pomcp}, we select a default action according to a discrete probability distribution weighted by the confidence levels $\rho_i$'s of each action (derived from counterexamples found by ILASP).
The only difference with respect to POMCP is that we set $\rho_i = 0$ in Equation \eqref{eq:prob_rollout} for actions \emph{not entailed by policy heuristics}. In fact, the lower bound in DESPOT must suggest the minimum expected value of a given node. If this value is high, because it is generated from a good default action (suggested by the policy specifications), the gap $\epsilon(b_0) = u(b_0) - l(b_0)$ reduces, resulting in faster convergence to the termination condition $\epsilon(b_0) < \epsilon_0$ (see Section \ref{sec:despot}). Thus, our methodology aims at selecting \emph{the best possible action} as the default one (hence computing the corresponding lower bound) at each DESPOT step.
For instance, with reference to our leading example, in Figure \ref{fig:leading_ex} the policy specifications entails only \stt{check(1)}, which is selected as default action with $\rho_i = 100\%$. In Figure \ref{fig:leading_ex_2} \stt{north} and \stt{check(2)} are entailed, with selection probabilities $\rho_{\stt{north}} \approx 60\%, \ \rho_{\stt{check}} \approx 40\%$. Finally, in Figure \ref{fig:leading_ex_3} \stt{sample} and \stt{check(2)} are entailed, with selection probabilities $\rho_{\stt{sample}} \approx 60\%, \ \rho_{\stt{check}} \approx 40\%$.

\section{Experiments}\label{sec:exp}
\begin{table}
    \centering
    \caption{Details of experiments. For each domain (RS = rocksample, Poc = pocman), we report main parameters which are either augmented ($\uparrow$) or reduced ($\downarrow$) with respect to the training setting, following the nomenclature in Sections \ref{sec:rs}-\ref{sec:poc}. We also report main POMDP challenges analyzed in this paper: $|\pi|$ denoting the average planning horizon, and $|A|$ denoting the size of the action space.}    
    \resizebox{\textwidth}{!}{%
    \begin{tabular}{c|c|c|c|c|c|c|c}
        \textbf{EXP} & \textbf{Domain} & \multicolumn{6}{c}{\textbf{Solver}} \\
        \toprule
        & & POMCP & \multicolumn{2}{c|}{DESPOT} & AdaOPS & POMCPOW & Deep RL\\
        \toprule
        & & C++ & C++ & Julia & Julia & Julia & Python\\
        \toprule
        \multirow{2}{*}{\shortstack[c]{\textbf{EXP-1}\\Training}} & \shortstack[2]{RS($N=12, M=4$)\\\textcolor{red}{$|A|=10, |\pi|=17$}} & \checkmark & &  & & & \\ 
        \cmidrule{2-8}
        & \shortstack[2]{Poc($10\times 10, G=2, \rho_f=0.5, \rho_g=0.75$)\\\textcolor{red}{$|A|=4, |\pi|=73$}} & \checkmark & & & & & \\ 
        \toprule
        \multirow{2}{*}{\shortstack[c]{\textbf{EXP-2}\\Generalization}} & \shortstack[2]{RS($N\uparrow, M\uparrow$)\\\textcolor{red}{$|A|=10\ldots 26, |\pi|=17\ldots 67$}} & \checkmark & \checkmark & \checkmark & \checkmark & \checkmark & \\ 
        \cmidrule{2-8}
        & \shortstack[2]{Poc(size$\uparrow, G\uparrow, \rho_g\uparrow, \rho_f\downarrow$)\\\textcolor{red}{$|A|=4, |\pi|=73\ldots 85$}} & \checkmark & \checkmark & & & & \\ 
        \toprule
        \shortstack[2]{\textbf{EXP-3}\\Bad examples} & \shortstack[2]{RS($N\uparrow, M\uparrow$)\\\textcolor{red}{$|A|=10\ldots 14, |\pi|=17\ldots 47$}} & \checkmark & \checkmark &  & & & \\
        \toprule
        \shortstack[2]{\textbf{EXP-4}\\Few examples} & \shortstack[2]{RS($N\uparrow, M\uparrow$)\\\textcolor{red}{$|A|=10\ldots 14, |\pi|=17\ldots 47$}} & \checkmark & \checkmark &  & & & \\
        \toprule
        \shortstack[2]{\textbf{EXP-5}\\RL comparison} & \shortstack[2]{RS($N\uparrow, M\uparrow$)\\\textcolor{red}{$|A|=10\ldots 14, |\pi|=17\ldots 47$}} & \checkmark & \checkmark &  & & & \checkmark \\
        \bottomrule
    \end{tabular}}
    \label{tab:exps}
\end{table}
For both pocman and rocksample scenarios, we first use POMCP to generate 1000 POMDP traces to learn policy specifications from. Each trace is a sequence of belief-action pairs collected during the execution of one specific instance of the problem. 
All traces share a similar environmental setting, e.g., the same grid size and number of rocks for rocksample, and the same grid size and number of ghosts for pocman.
POMCP is run \emph{without any heuristics} in this phase, so no bias is present in the generated traces.
We empirically set $2^{15}$ online simulations and particles in POMCP\footnote{Following standard practice in POMDP, the numbers of particles and simulations coincide. Hence, from now on, we will refer to particles and simulations indistinguishably.}, in order to generate good-quality traces with sufficiently high discounted return. Moreover, we only select traces achieving discounted return above the average over the total runs. This is a very simple and \emph{domain-agnostic} criterion to select only the best traces and most relevant examples to learn good policy heuristics.
The discount factor $\gamma$ is set to $0.95$ for all experiments.  

The empirical analysis presented in this section investigates the following aspects: 
\begin{itemize}
\item \textbf{EXP-1} the impact of learned policy heuristics on the performance of online POMDP solving, in scenarios \emph{similar to the training setting} (i.e., with the same grid size and number of rocks for rocksample, and the same grid size and number of ghosts for pocman). This is a preliminary assessment of the quality of learned heuristics;
\item \textbf{EXP-2} the generalization capabilities of learned policy specifications, e.g., in larger grids or with more rocks / ghosts; 
\item \textbf{EXP-3} the robustness of our methodology to low-quality heuristics, e.g., from bad examples or when the set of user-defined semantic features is incomplete;
\item \textbf{EXP-4} the influence of the number of example traces on the convergence of ILASP learner and the performance of online POMDP solvers;
\item \textbf{EXP-5} the comparison with black-box methods for POMDP planning and policy heuristics learning, e.g., neural network architectures.
\end{itemize}
\noindent
For \textbf{EXP-1}, we consider POMCP solver. Similarly to \shortciteA{mazzi2023learning}, we conduct experiments with different numbers of particles \emph{below} $2^{15}$, but with the same number of rocks / ghosts and grid size. However, we introduce a novel ablation study, to highlight the role of policy heuristics in different phases of MCTS (namely, UCT and rollout). We also add the comparison with handcrafted policy specifications, as provided in former published research. 
For \textbf{EXP-2}, we consider POMCP, DESPOT and the more recent and powerful AdaOPS, which is meant to deal with large complex domains. We consider POMCPOW only as a baseline, because it has worse performance than the others in our experiments, as confirmed by \shortciteA{wu2021adaptive}. We cannot perform a cross-planner validation, since the software implementations are different\footnote{POMCP and DESPOT are implemented in C++, adapting the original code released by \shortciteA{silver2010monte} and \shortciteA{ye2017despot}, respectively. AdaOPS and POMCPOW are implemented in Julia, following the original codes at \url{https://github.com/LAMDA-POMDP/AdaOPS.jl.git} and \url{https://github.com/JuliaPOMDP/POMCPOW.jl}, respectively.} and we decided to rely on publicly available packages to foster reproducibility.
This does not limit the validity of our analysis, since the goal of \textbf{EXP-1} and \textbf{EXP-2} is to assess the performance of a given online POMDP solver \emph{with and without the use of learned policy heuristics}, in order to validate the outcome of our proposed methodology. 
Moreover, for DESPOT and POMCP we analyze both rocksample and pocman. For AdaOPS, we only consider the rocksample scenario, since the pocman is not implemented in the available package. 
\textbf{EXP-3} and \textbf{EXP-4}, we consider both the rocksample and pocman scenarios, limiting our analysis to POMCP and DESPOT. In fact, POMCPOW and AdaOPS are efficient extensions of POMCP and DESPOT, hence, results on the robustness of our methodology do not differ from the original algorithms.
Finally, in \textbf{EXP-5} we evaluate the performance of Approximate Information State (AIS) learning in POMDPs, by \shortciteA{subramanian2022approximate}. We use the related code\footnote{\url{https://github.com/info-structures/ais}}, restricting our analysis only to rocksample since pocman is not implemented.
We summarize the settings and solvers for the 5 experiments in Table \ref{tab:exps}, highlighting in red the challenges of different scenarios related to the planning horizon $|\pi|$ and the size of the action space $|A|$.

In the following, in Section \ref{sec:learning_res} we report the learned heuristics for the two domains, detailing the learning performance and highlighting the high interpretability, which allows to compare our specifications to the handcrafted ones provided by domain experts in \shortciteA{silver2010monte,ye2017despot}. Then, we show quantitative experiments for \textbf{EXP-1} to \textbf{EXP-5} in the subsequent subsections.
All tests in this paper are run on standard hardware equipped with \SI{3.6}{GHz} e.g. Ryzen~7 and \SI{32}{GB} RAM\footnote{Since the hardware differs from the one used in our original work \shortcite{mazzi2023learning}, we have run also the experiments already reported there on the new machine, in order to perform a more fair empirical evaluation.}.
Moreover, we do not exploit any ASP reasoner during online planning with either DESPOT or POMCP. Instead, we implement learned heuristics in C++ syntax, to achieve the best online computational performance.

\subsection{Learned Policy Specifications}\label{sec:learning_res}
We now report the learning performance and the policy specifications for rocksample and pocman.

\subsubsection{Rocksample}
In rocksample, the training scenario consists of a $12 \times 12$ ($N=12$) grid with $M = 4$ rocks.
Scenarios differ because of the random positioning and real values of rocks.
The ASP encoding of the problem (i.e., the set of environmental features $\pazocal{F}$ and action atoms $\pazocal{A}$) is reported in Section \ref{sec:met_asp}.
Additionally, we introduce the atom \stt{target(R)}, meaning that rock \stt{R} is the next to be sampled, which is considered both as an environmental feature and an action atom when defining ILASP search space as in Section \ref{sec:met_ilasp_learn}. It allows to learn more compact rules (since ILASP does not support predicate invention, see \shortciteR{law2023conflict}), and can be easily derived from traces of execution, identifying sub-traces which do not contain \stt{sample(R)} actions and assuming that all actions executed before sampling were directed to the final sampling. Moreover, we consider the \stt{exit} action, which is equivalent to \stt{east}, but has possibly different specifications.

Starting from 1000 traces of execution (generated within $\approx$ \SI{12}{min}), we generate 8487 ILASP examples (corresponding to single steps of execution) for each possible head atom in $\pazocal{A}$ (after selecting only 492 traces with discounted return higher than the average). 
The following ASP heuristics are found:
\begin{align}
    \label{eq:rs_h}
    &\stt{east :- target(R), delta\_x(R,D), D} \geq \stt{1.}\\
    \nonumber&\stt{west :- target(R), delta\_x(R,D), D} \leq \stt{-1.}\\
    \nonumber&\stt{north :- target(R), delta\_y(R,D), D} \geq \stt{1.}\\
    \nonumber&\stt{south :- target(R), delta\_y(R,D), D} \leq \stt{-1.}\\
    \nonumber&\stt{0 \{target(R) : dist(R,D), not sampled(R), D}\leq \stt{1.}\\
    \nonumber&\ \ \ \ \stt{target(R) : guess(R,V), not sampled(R), 70} \leq \stt{V} \leq \stt{80\} }M.\\
    \nonumber&\stt{:}\sim \stt{target(R), dist(R,D).[D@1, R, D]}\\
    \nonumber&\stt{:}\sim \stt{target(R), guess(R,V).[-V@2,R,V]}\\
    \nonumber&\stt{check(R) :- target(R), not sampled(R), guess(R,V), V} \leq \stt{50.}\\
    \nonumber&\stt{check(R) :- guess(R,V), not sampled(R), dist(R,D), D} \leq \stt{0, V} \leq \stt{80.}\\
    \nonumber&\stt{sample(R) :- target(R), dist(R,D), D} \leq \stt{0, not sampled(R), guess(R,V), V} \geq \stt{90.}\\
    \nonumber&\stt{exit :- guess(R,V), V} \leq \stt{40, not sampled(R), num\_sampled(N), N} \geq \stt{25.}\\
    \nonumber&\stt{exit :- dist(R,D), 5} \leq \stt{D} \leq \stt{8, not sampled(R), num\_sampled(N), N} \geq \stt{25.}
\end{align}
\noindent
The first 4 axioms define basic conditions for moving in the cardinal directions towards a chosen target rock.
The aggregate rules for \stt{target(R)} explain that an unsampled rock is chosen as a target when either it is close to the agent (\stt{D $\leq$ 1}) or probably valuable (\stt{70 $\leq$ V $\leq$ 80}).
Moreover, the weak constraints state that, among multiple possible targets, first the closest one should be chosen (priority 1 with cost \stt{D}), then the most probably valuable (priority 2 with cost \stt{-V}).
Axioms for \stt{check(R)} specify that it is worth sampling either when the chosen target rock has low probability of being valuable (\stt{V $\leq$ 50}), or the agent is at the location of an unsampled rock with value probability $\leq 80$. In the first case, the heuristics pushes the agent to refine its belief for the most uncertain rocks, in order to compute the best strategy for the task. Similarly, in the second case, the effect is a \emph{cautious} behavior of the agent, i.e., it wants to be almost sure about the value of the rock before picking it.
When the agent is on the location of the target rock and the probability of the rock being valuable is $\geq 90$, sampling should be executed. 
Finally, the agent should terminate the task, exiting the grid whenever the percentage of unsampled rocks is $\geq 25\%$ and either an unsampled rock has low value probability (\stt{V $\leq$ 40}) or it is far (\stt{5 $\leq$ D $\leq$ 8}).

ILASP is able to find all axioms within $\approx$ \SI{6}{min}, in a search space of $\approx 29500$ axioms, setting a maximum length of 8 atoms for normal rules (Equation \eqref{eq:ss}) with mode bias. Specifically, we allow in the axioms 4 environmental features plus the comparison operators $\stt{V}\lesseqgtr\Bar{v}, \stt{D}\lesseqgtr\Bar{d}$. This is empirically set to balance between computational requirements (more features imply longer axioms in $S_M$, which influences the conflict-driven constraint learning approach of ILASP) and expressiveness of policy specifications. When learning weak constraints for \stt{target(R)} atom, we set a maximum rule length (Equation \eqref{eq:ss_weak}) of 6 since we do not need to include comparison operators.
The percentage of not covered examples $cov_i$ (used for action selection in the rollout phase of POMCP and for the definition of the default policy in DESPOT) is 65\% for \stt{north} and \stt{south} actions, 57\% for \stt{east}, 73\% for \stt{west}, 84\% for \stt{exit}, 85\% for \stt{check(R)} and 65\% \stt{sample}. 

The handcrafted heuristics proposed by \shortciteA{silver2010monte} is defined as follows in the original paper:
\begin{quote}
    \textit{Either check a rock whenever the value probability is uncertain ($< 100\%$) and it has been measured few times ($<5$); or sample a rock if the agent is at its location and collected observations are more positive (good value) than bad; or move towards a rock with more good than bad observations.}
\end{quote}
\noindent
Thanks to the \emph{interpretability of logical statements}, we can make a qualitative comparison between our learned policy specifications \eqref{eq:rs_h} and the handcrafted ones. 
Our rules state that a rock should be checked when the value probability is low ($\leq 50\%$), which is a possible consequence of having few measurements about that rock, as stated in the handcrafted heuristics. Moreover, our rules suggest to check a rock when the agent is at its location, even with high probability of the rock being valuable, in order to ensure a good pick.
Our rules also indicate that a rock with a high probability of being valuable ($V \geq 90\%$) should be sampled when the agent is at its location, similarly to the handcrafted heuristics.
Finally, ASP heuristics suggest to move towards candidate target rocks, i.e., the ones with high goodness probability ($V \geq 70\%$) as in the handcrafted heuristics, or even the closest ones to the agent. Hence, ILASP is able to find policy specifications which are similar to the handcrafted ones, but evaluate the belief distribution rather than single observations. Our methodology cannot capture the policy heuristics about the number of times a rock has been checked, since it would require more complex temporal expressiveness, e.g., the \emph{until} operator from linear temporal logic. This will be discussed in Section \ref{sec:discussion}.

\subsubsection{Pocman}
In pocman, the training scenario consists of a $10 \times 10$ grid with $G=2$ ghosts.
The 1000 training instances generated differ on the basis of the motion of ghosts, which start always at the same location but wander randomly in the grid. Moreover, when close enough to the pocman agent ($< 4$ Manhattan distance), the ghosts can chase it with probability $\rho_g = 75\%$ probability. Each non-wall cell may contain a food pellet with probability $\rho_f = 50\%$ probability.
We generate ILASP examples according to the following ASP environmental features $\pazocal{F}$:
\begin{itemize}
    \item \stt{food(C,D,V)}, representing the discrete probability (represented as a percentage) \stt{V}$\in \{0, 10, \ldots, 100\}$ that a food pellet is within \stt{D}$\in \mathbb{Z}$ Manhattan distance from the pocman in \stt{C} cardinal direction, \stt{C}$\in$\stt{\{north, south, east, west\}};
    \item \stt{ghost(C,D,V)}, representing the discrete (percentage) probability \stt{V}$\in \{0, 10, \ldots, 100\}$ that a ghost is within \stt{D} Manhattan distance from the pocman in \stt{C} cardinal direction;
    \item \stt{wall(C)}, representing the presence of a wall in \stt{C} cardinal direction.
\end{itemize} 
\noindent
The set $\pazocal{A}$ of ASP actions consists of the single atom \stt{move(C)}, being \stt{C$\in$\{north, south, east, west\}}.

Over 1000 traces of execution (generated within $\approx$ \SI{6}{min}), only the 460 with discounted return greater than the average are selected. Following the methodology presented in Section \ref{sec:met_ilasp}, we then have 43055 ILASP examples. 
The following policy heuristics is found:
\begin{equation}
    \label{eq:poc_h}
    \stt{move(C) :- not wall(C), ghost(C, D, V), V}\leq \stt{50, D}\leq \stt{6.} 
\end{equation}
\noindent
meaning that the agent should move in direction \stt{C} whenever there is no wall and the probability to find a ghost within Manhattan distance of 6 is $\leq 50\%$. ILASP is able to find axioms within $\approx $\SI{48}{s}, in a search space of $\approx 21000$ axioms, generated considering a maximum normal rule length of 6 atoms (3 environmental features + comparison atoms) with mode bias. We do not learn weak constraints, hence we do not report details about the search space in Equation \eqref{eq:ss_weak}.
The confidence level for \stt{move(C)} action is $cov_i=98\%$.

The handcrafted heuristics by \shortciteA{silver2010monte} is: 
\begin{quote}
    \emph{Move in a direction where no ghost nor wall is seen, and avoid changing direction to minimize the number of steps.}
\end{quote}
\noindent
Similarly to the rocksample domain, we leverage on interpretability of ASP heuristics to make a qualitative comparison. The learned rule in Equation \eqref{eq:poc_h} captures the first part of the handcrafted heuristics, stating that the pocman should move in a direction where there is no wall and the probability of finding a ghost within 6 steps is low ($\leq 50\%$). The second handcrafted suggestion (avoid direction changes) is not represented in the environmental features for the task, since it requires memory of the past actions, hence introducing a temporal dimension in the learning problem. However, in the following we will see that this does not impact on the performance of the planner. Moreover, one fundamental difference is that the handcrafted heuristics depends on the observations of the agent (e.g., seeing a ghost), while the learned heuristics is based on the belief distribution about ghost locations, hence it more directly involves reasoning about the uncertain model of the world built through experience during task execution.

\subsection{Results in the Training Setting (\textbf{EXP-1})}
\begin{figure}
    \centering
    \begin{subfigure}{0.45\textwidth}
    \includegraphics[width=\linewidth]{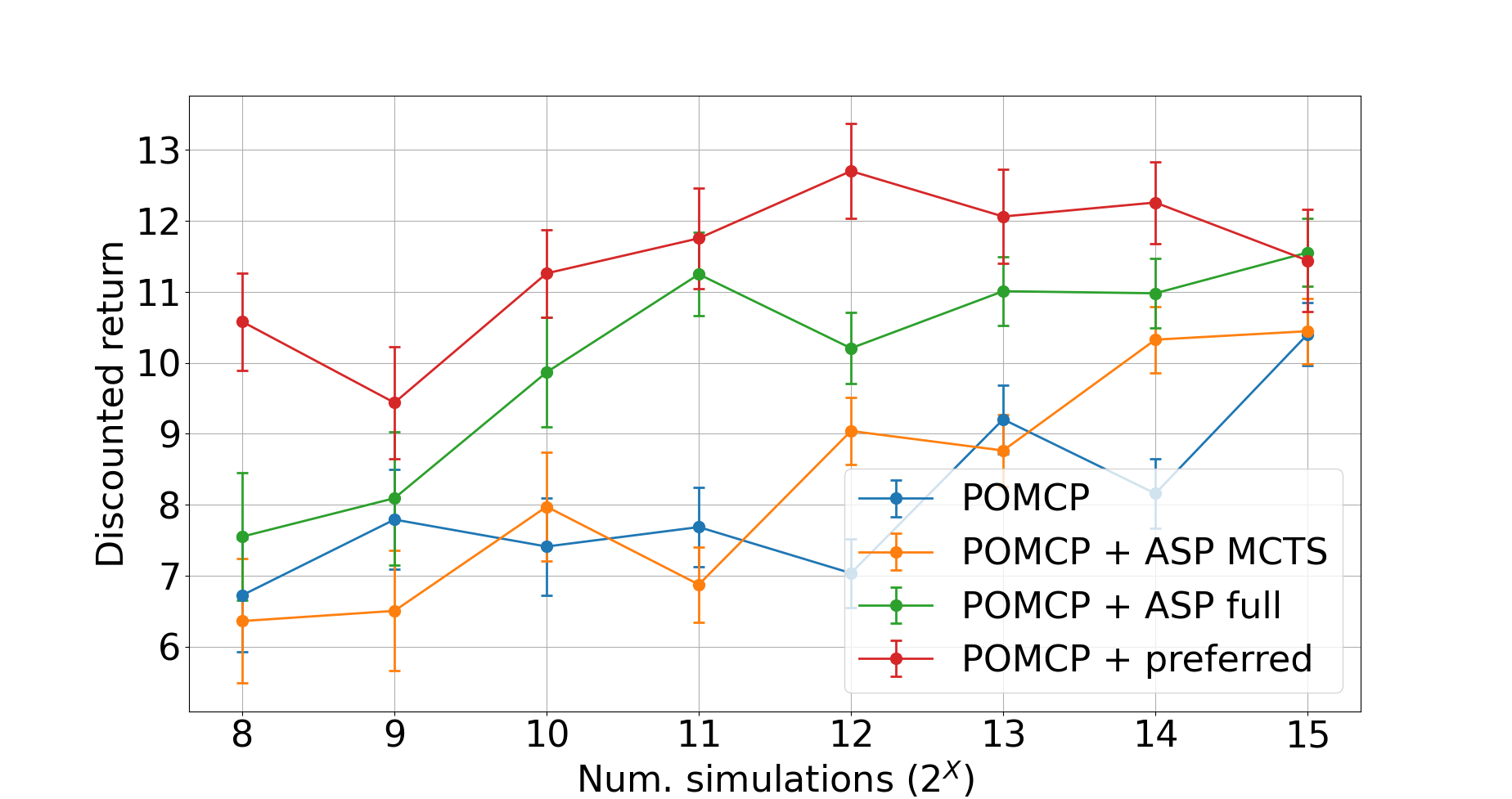}
    \caption{Rocksample ($N=12$, $M=4$)\label{fig:part_4rocks}}
    \end{subfigure}
    \begin{subfigure}{0.45\textwidth}
    \includegraphics[width=\linewidth]{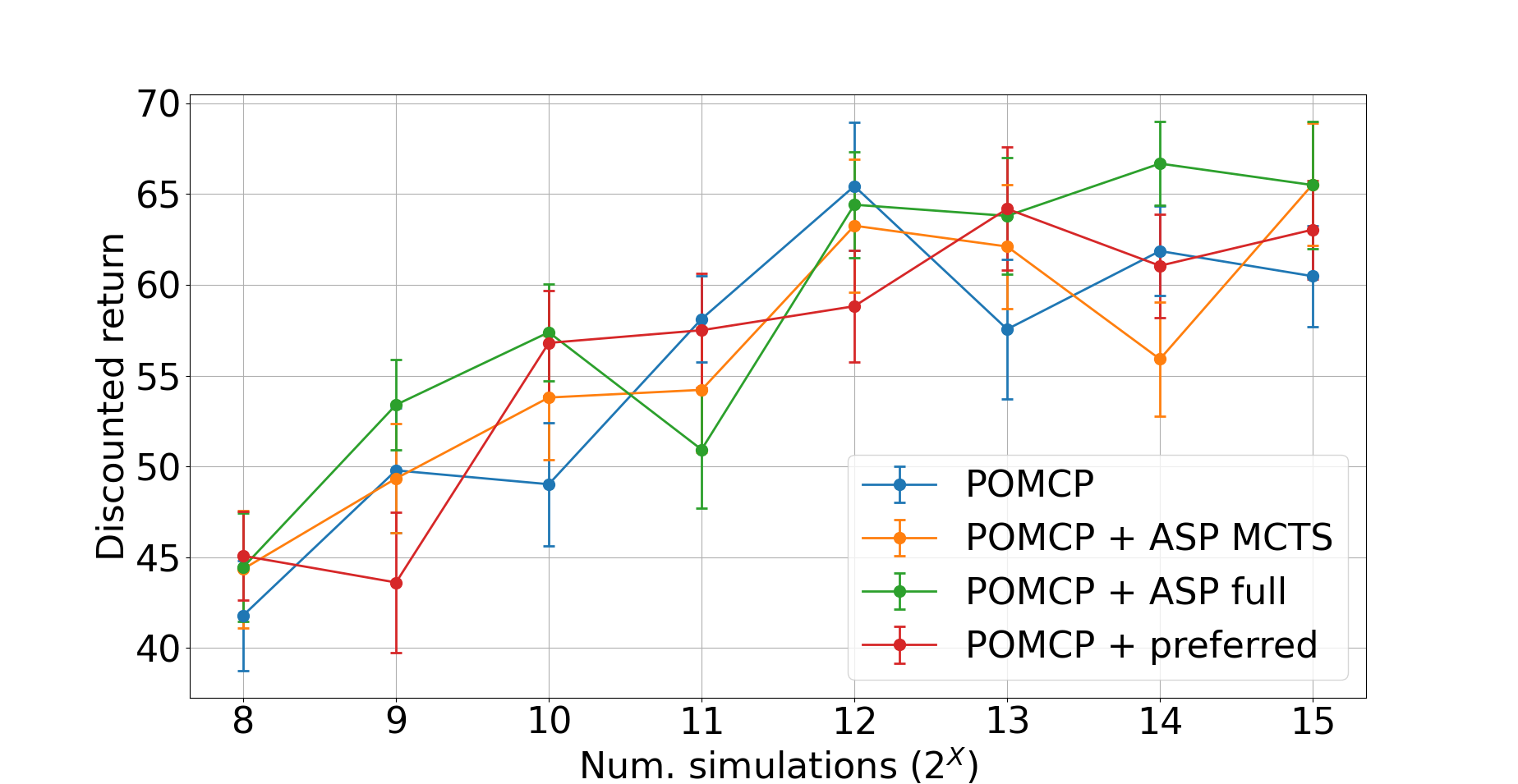}
    \caption{Pocman ($10\times 10$ grid, $G=2$)\label{fig:part_minipocman}}
    \end{subfigure}\\
    \begin{subfigure}{0.45\textwidth}
    \includegraphics[width=\linewidth]{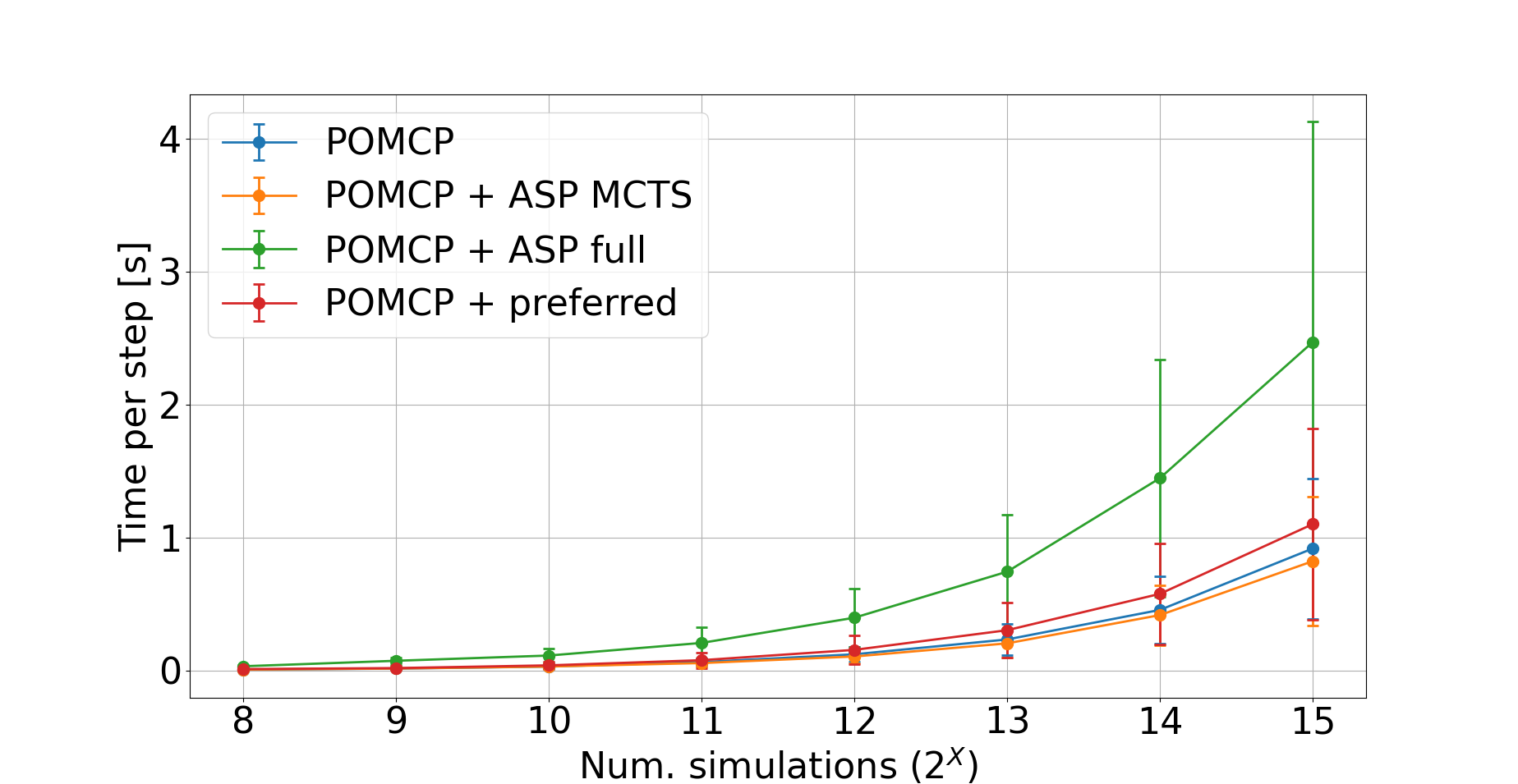}
    \caption{Rocksample ($N=12$, $M=4$)\label{fig:time_part_4rocks}}
    \end{subfigure}
    \begin{subfigure}{0.45\textwidth}
    \includegraphics[width=\linewidth]{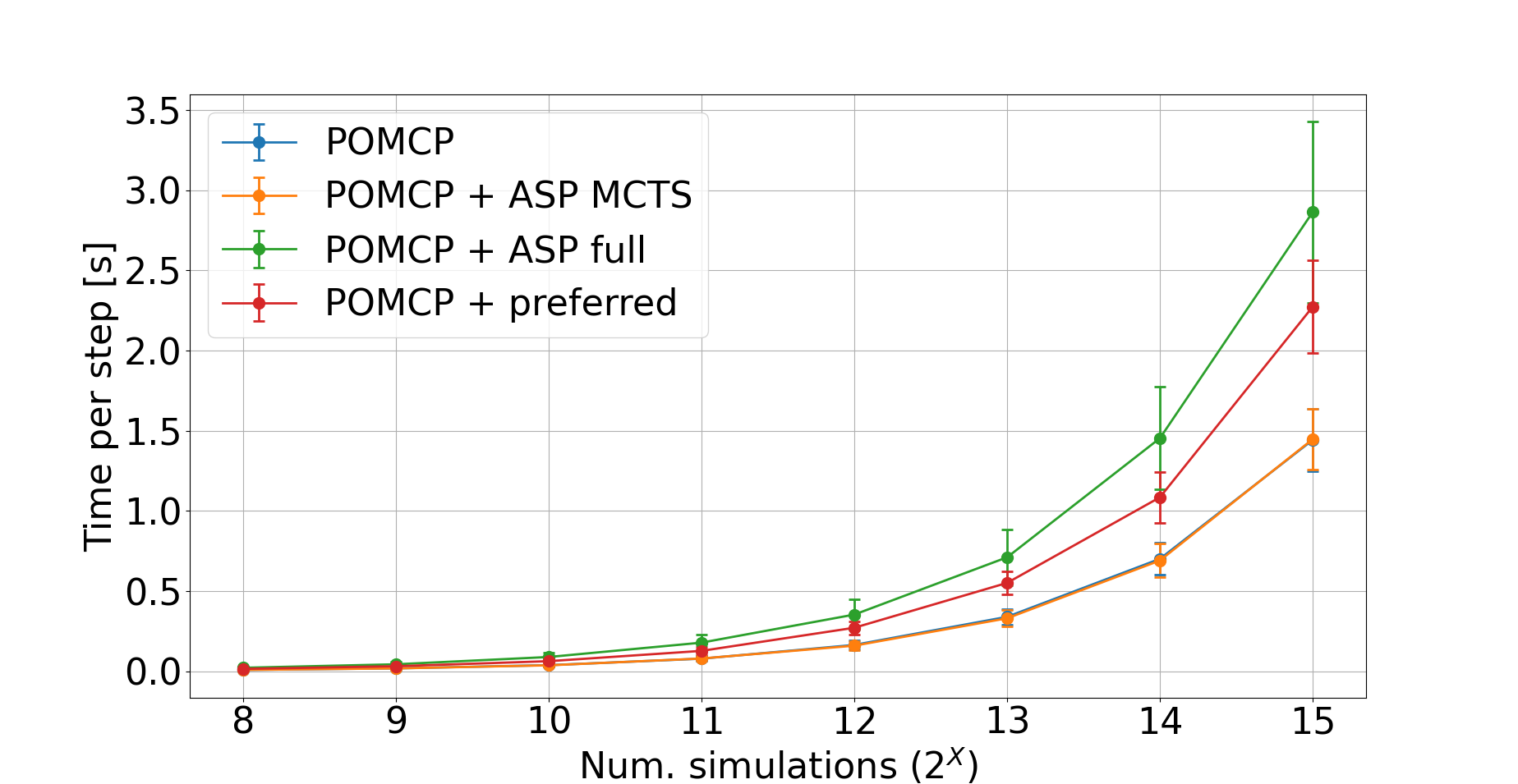}
    \caption{Pocman ($10\times 10$ grid, $G=2$)\label{fig:time_part_minipocman}}
    \end{subfigure}
    \caption{Discounted return (\textbf{top}) and computational time per step (\textbf{bottom}) with POMCP (mean $\pm$ standard deviation) for rocksample and pocman with different number of particles in the training setting (\textbf{EXP-1}).}
    \label{fig:exp1}
\end{figure}
We now evaluate the performance of POMCP with and without learned policy heuristics in scenarios similar to the training setting (i.e., same grid size and same number of rocks / ghosts), but with fewer online simulations and particles ($2^{x}\ \forall x \in \{8,9,10,11,12,13,14,15\} $) to approximate the belief. This affects the accuracy of the policy found by POMCP, hence having a good heuristics is crucial to achieve high discounted return.
For both rocksample and pocman, we perform an ablation study of our methodology, comparing 4 different algorithms:
\begin{itemize}
    \item \emph{POMCP}, i.e., standard POMCP without any policy heuristics(the one used also to generate learning traces);
    \item \emph{POMCP + ASP MCTS}, i.e., standard POMCP with policy specifications added in the UCT phase, as explained in Section \ref{sec:met_pomcp}. This corresponds to the methodology proposed by \shortciteA{mazzi2023learning};
    \item \emph{POMCP + ASP full}, i.e., standard POMCP with policy specifications added in the UCT and rollout phases. This corresponds to the methodology proposed in this paper;
    \item \emph{POMCP + preferred}, i.e., standard POMCP with handcrafted policy specifications (presented in Section \ref{sec:learning_res}) added in the UCT and rollout phases.
\end{itemize}

\subsubsection{Rocksample}
The agent moves in a $12\times 12$ ($N=12$) grid with $M=4$ rocks.
For each number of particles, we run POMCP 50 times with random location and value of rocks.
This experiment was partly already presented in our former paper \shortcite{mazzi2023learning}.
However, for a more fair evaluation, here we set rocks sufficiently far one from the other, sub-dividing the original grid into $M$ micro-grids, each one hosting at most one rock. In this way we avoid problem instances where two valuable rocks are almost adjacent, resulting in higher potential return for the agent.

Figure \ref{fig:part_4rocks} shows the discounted return achieved by the agent.
\emph{POMCP + ASP full} (our methodology) is significantly better than \emph{POMCP + ASP MCTS}\footnote{Notice that these results differ from \shortciteA{mazzi2023learning} because that paper focued only on the impact of policy heuristics \emph{in UCT phase}, hence the rollout (both for \emph{POMCP} and \emph{POMCP + ASP MCTS}) was driven by the handcrafted heuristics proposed by \shortciteA{silver2010monte}. In this paper, we introduce learned heuristics in the rollout phase, hence the standard \emph{POMCP} is completely uninformed (i.e., it does not use any heuristics, hence implementing standard UCT with no prior initialization of $N(ha)$ and uniformly random rollout). Similarly, \emph{POMCP + ASP MCTS} performs random rollout. Moreover, the experiments in our former paper assumed a different location of rocks, which were not equally distributed. This affected the average discounted return, since two valuable rocks might be adjacent on the map, thus potentially helping the agent.} and \emph{POMCP}, because it considers policy specifications \emph{also in the rollout phase}. The performance are slightly worse than \emph{POMCP + preferred}. This is probably due to the different knowledge encoded in the handcrafted heuristics. For instance, from Section \ref{sec:learning_res}, \emph{POMCP + preferred} is driven to perform \stt{check(R)} \emph{if \stt{R} was checked less than 5 times}. This temporal information requires to track the history of the task and cannot be captured by simply looking at traces of execution. On the other hand, while checking a rock is costless for the agent, multiple checks affect the plan length, hence the discounted return.

Figure \ref{fig:time_part_4rocks} shows the computational time per step taken by the agent to choose the next action.
Using policy heuristics also in the rollout, \emph{POMCP + ASP full} takes more time as the number of simulations increases. 
However, comparing the performance of \emph{POMCP + ASP full} and \emph{POMCP} in Figure \ref{fig:part_4rocks}, our algorithm requires only $2^{10}$ simulations to achieve the same performance as \emph{POMCP} with $2^{15}$ simulations. Thus, from Figure \ref{fig:time_part_4rocks}, the computational time per step is significantly reduced even with respect to pure POMCP (\emph{POMCP + ASP full} with $2^{10}$ simulations requires $\approx$\SI{0.1}{s} per step, while \emph{POMCP} with $2^{15}$ simulations requires $\approx$\SI{1}{s} per step).
\emph{POMCP + preferred} does not take much time because the policy heuristics\emph{only considers observations}, while our heuristics takes into account the full belief distribution.

\subsubsection{Pocman}\label{sec:exp1_pocman}
The agent moves in a $10\times 10$ grid with $G=2$ rocks.
Each non-wall cell may have a food pellet with $50\%$ probability, while ghosts can chase the agent with $75\%$ probability.
For each number of particles, we run POMCP 50 times.

Figures \ref{fig:part_minipocman}-\ref{fig:time_part_minipocman} show that there is no significant difference between the different algorithms, especially in terms of discounted return.
In fact, given the lower number of actions (only 4) with respect to rocksample (9 with $M=4$), the influence of the heuristics in POMCP is minor. Finally, as stated by \shortciteA{silver2010monte,ye2017despot}, the main challenge of pocman lies in the long planning horizon (for comparison, the rocksample agent on a $12 \times 12$ grid with 4 rocks executes $\approx 17$ steps per execution, while the pocman agent in the small domain takes $\approx 73$ actions), thus varying the number of simulations and particles to approximate the belief is not crucial for POMCP performance (over a small action space).

\subsection{Generalization to Unseen Scenarios (\textbf{EXP-2}) - Rocksample}
We now assess the performance of learned policy specifications in scenarios \emph{different from the training setting}.
Specifically, in the rocksample domain we increase the grid size $N$ (affecting the planning horizon) and the number of rocks $M$ (affecting the size of the action space).
This aims at showing the generality of our heuristics, especially in more complex environments involving larger actions spaces and planning horizons.
We extend our analysis to different online POMDP solvers, including POMCP, DESPOT and their more recent extensions POMCPOW \shortcite{sunberg2018online} and AdaOPS \shortcite{wu2021adaptive}, dealing with larger action spaces and state/observation spaces, respectively.

\subsubsection{POMCP}
\begin{figure}
    \centering
    \begin{subfigure}{0.32\textwidth}
    \includegraphics[width=\linewidth]{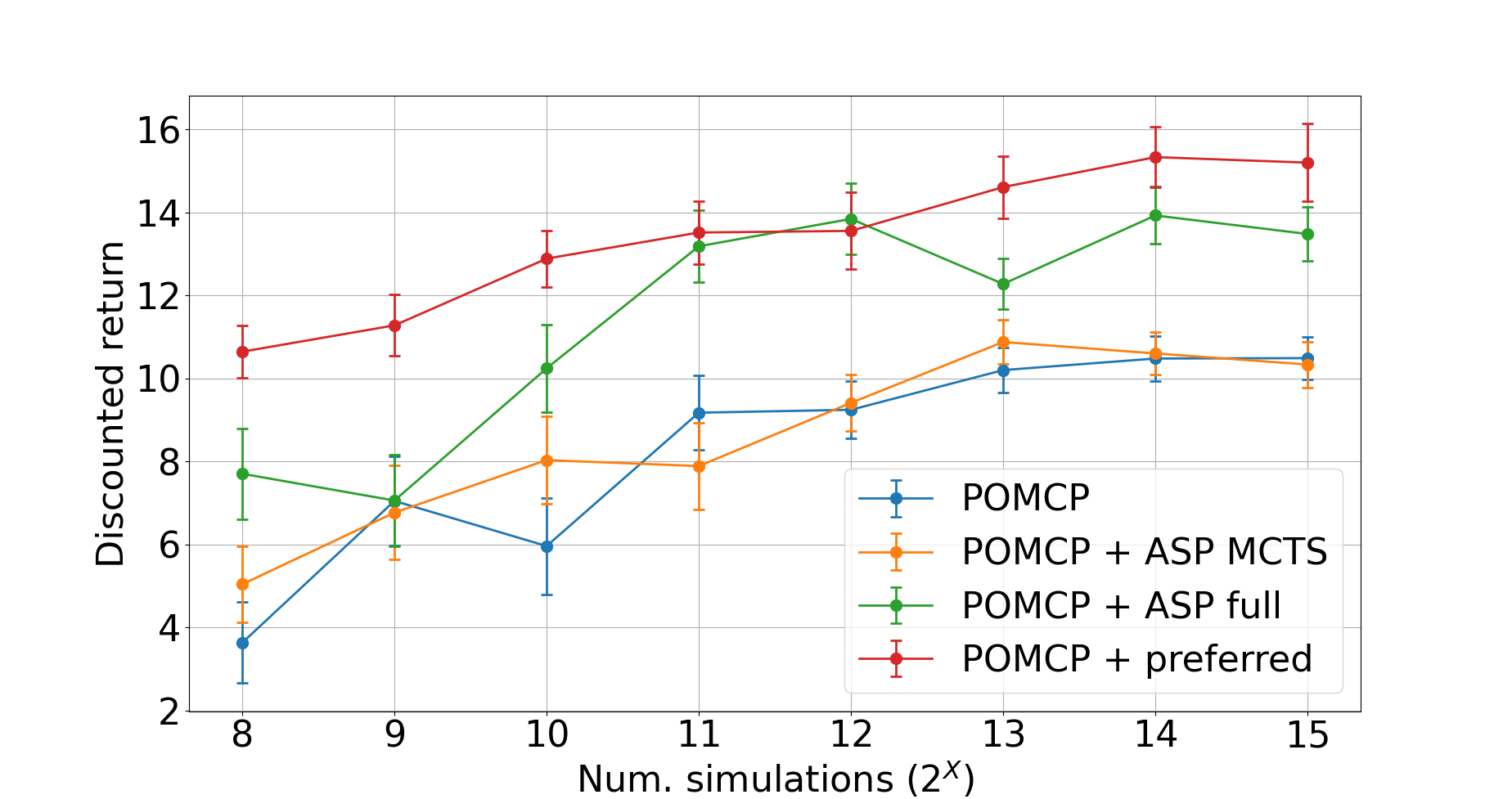}
    \caption{$N=12, M=8$\label{fig:part_8rocks}}
    \end{subfigure}
    \begin{subfigure}{0.32\textwidth}
    \includegraphics[width=\linewidth]{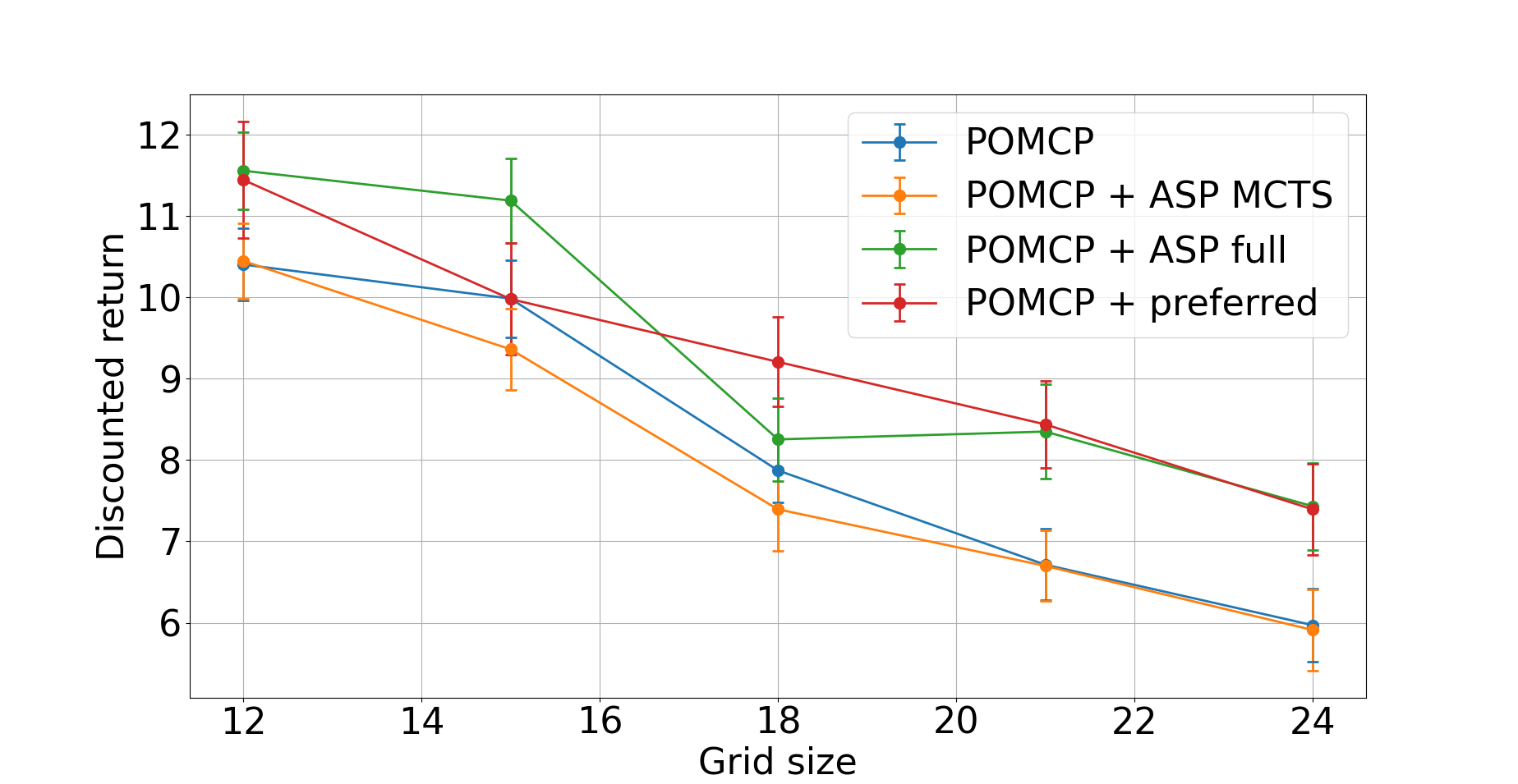}
    \caption{$M=4, 2^{15}$ simulations\label{fig:size_4rocks}}
    \end{subfigure}
    \begin{subfigure}{0.32\textwidth}
    \includegraphics[width=\linewidth]{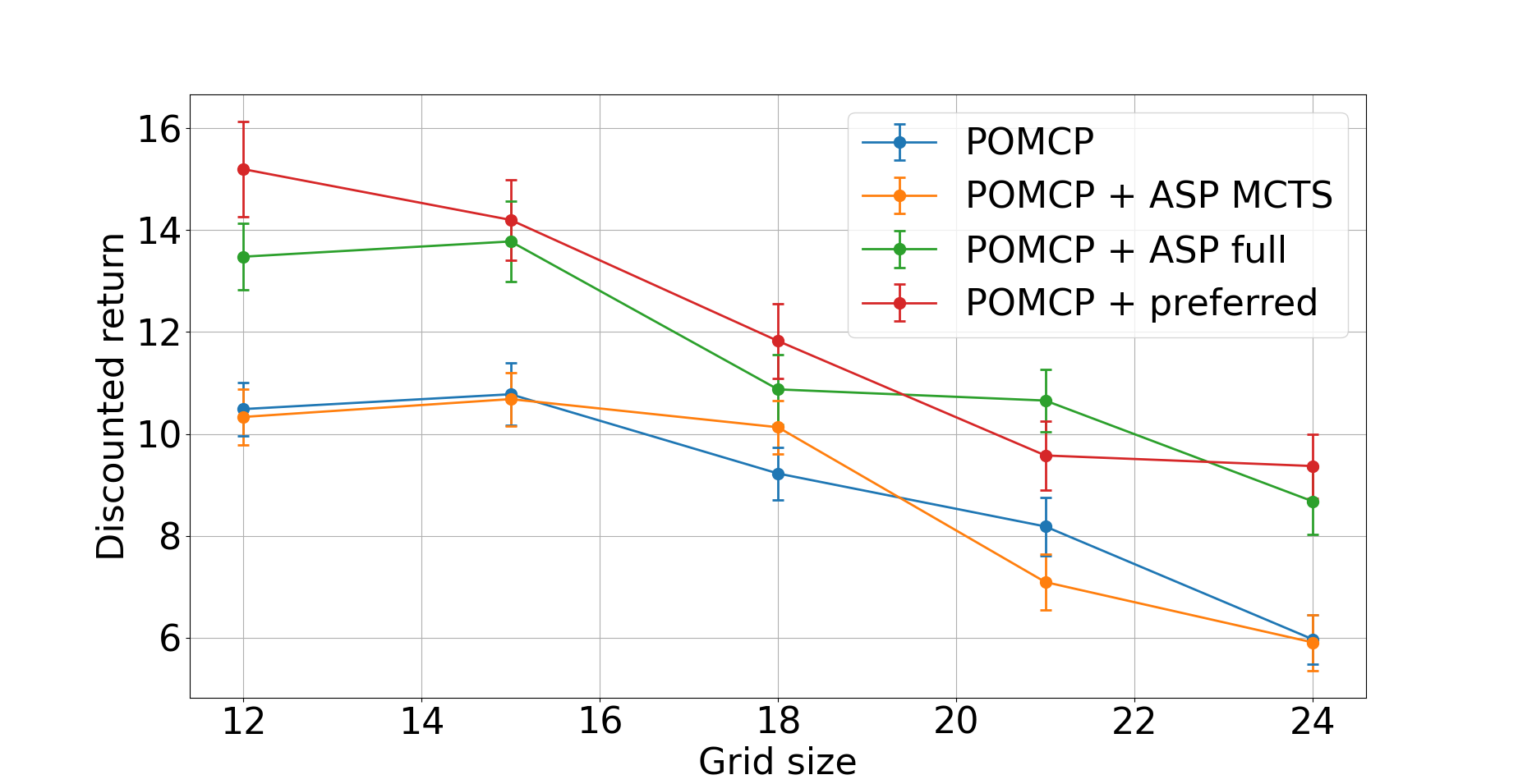}
    \caption{$M=8, 2^{15}$ simulations\label{fig:size_8rocks}}
    \end{subfigure}\\
    \begin{subfigure}{0.32\textwidth}
    \includegraphics[width=\linewidth]{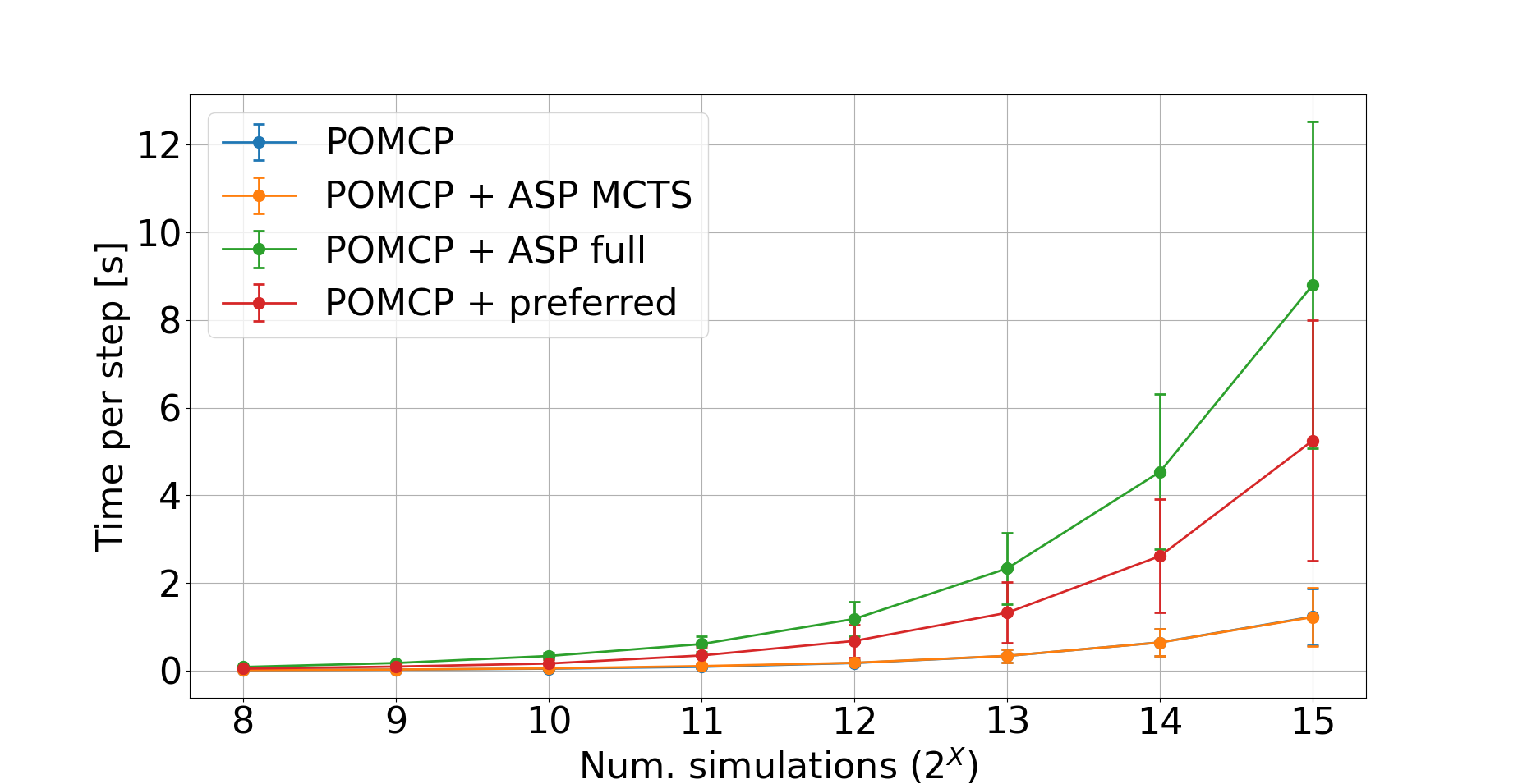}
    \caption{$N=12, M=8$\label{fig:time_part_8rocks}}
    \end{subfigure}
    \begin{subfigure}{0.32\textwidth}
    \includegraphics[width=\linewidth]{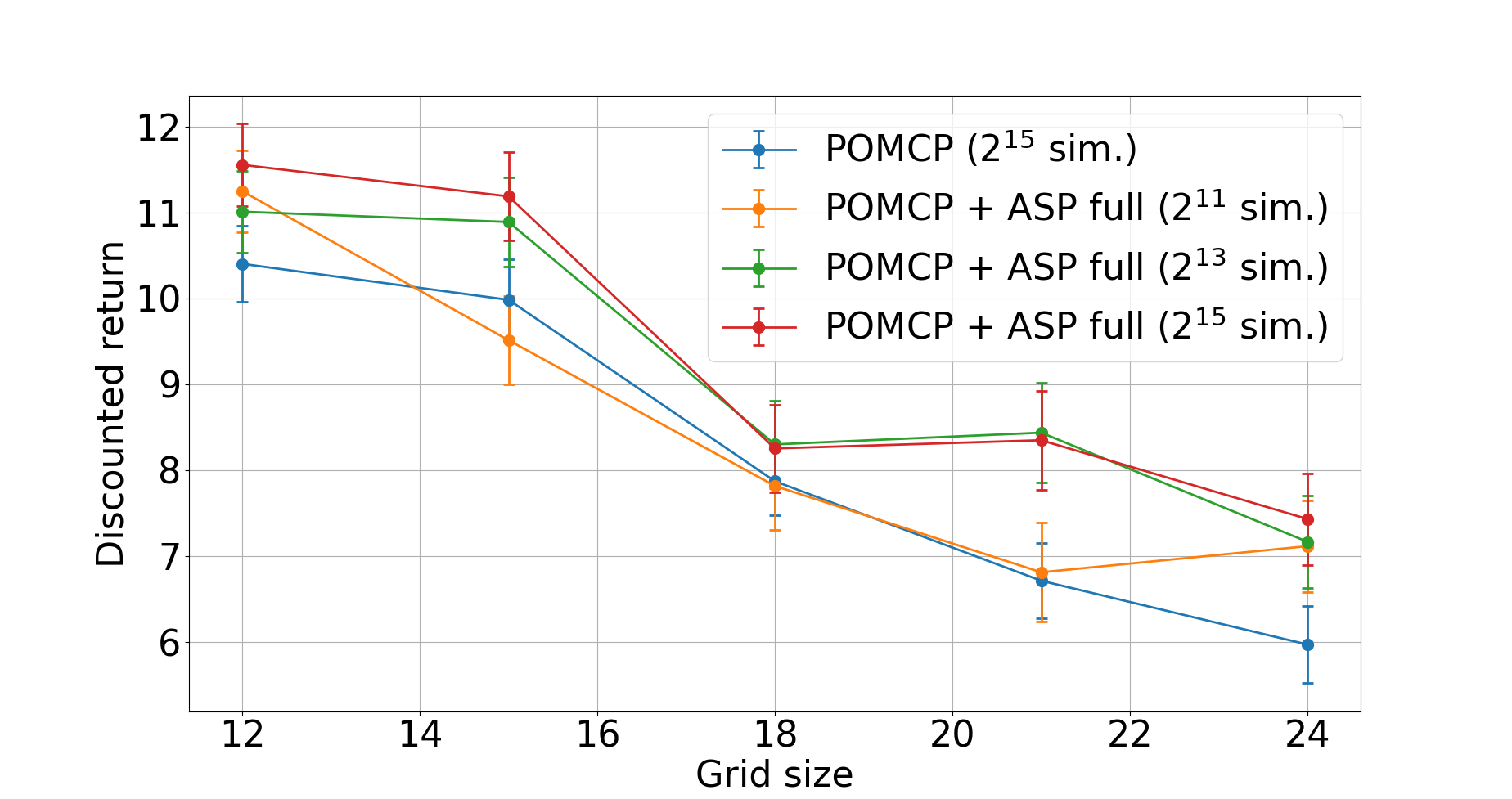}
    \caption{$M=4$\label{fig:size_manypart_4rocks}}
    \end{subfigure}
    \begin{subfigure}{0.32\textwidth}
    \includegraphics[width=\linewidth]{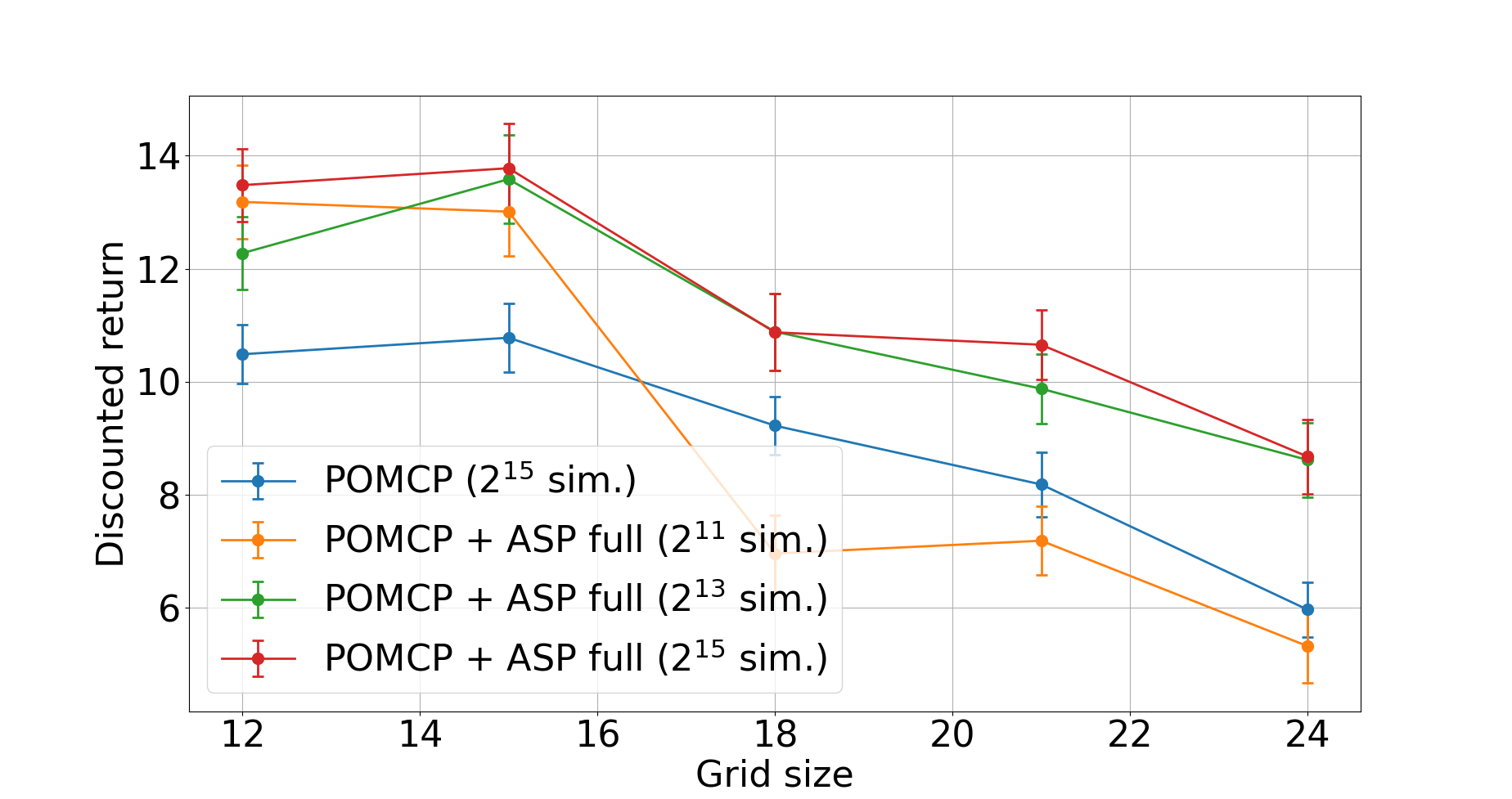}
    \caption{$M=8$\label{fig:size_manypart_8rocks}}
    \end{subfigure}
    \caption{POMCP performance (mean $\pm$ standard deviation) for rocksample with different number of rocks and grid size (\textbf{EXP-2}).}
    \label{fig:exp2_rs_pomcp}
\end{figure}
As a first experiment in POMCP, we increase the number of rocks ($M=8$) on a $12\times 12$ grid, and evaluate performance as the number of online simulations decreases.
Figures \ref{fig:part_8rocks}, \ref{fig:time_part_8rocks} show similar results as Figures \ref{fig:part_4rocks}, \ref{fig:time_part_4rocks} (\textbf{EXP-1}), i.e., \emph{POMCP + ASP full} achieves better discounted return than \emph{POMCP} and \emph{POMCP + ASP MCTS}, though slightly worse than \emph{POMCP + preferred}. Similarly our methodology requires the highest computational time per step (given the same number of simulations), but the overall computational effort is lower to achieve the same performance as \emph{POMCP} with $2^{15}$ particles, requiring only $2^{10}$ particles (hence, \emph{POMCP+ASP full} requires $\approx$\SI{0.2}{s}, while \emph{POMCP} $\approx$\SI{1.1}{s}).

Then, in Figures \ref{fig:size_4rocks}, \ref{fig:size_8rocks} we evaluate the contribution of learned policy specifications in larger grid sizes $N$ ($x$-axis) and with more rocks ($M=4$ and $M=8$, respectively), assuming a fixed number of particles and online simulations equal to $2^{15}$.
In both cases, \emph{POMCP + ASP full} achieves comparable performance to \emph{POMCP + preferred}, significantly outperforming \emph{POMCP} and \emph{POMCP + ASP MCTS}, especially at the largest grid size ($N=24$).

In Figures \ref{fig:size_manypart_4rocks}, \ref{fig:size_manypart_8rocks} ($M=4$ and $M=8$, respectively), we show that this trend is confirmed even when the number of online simulations is lowered. In particular, \emph{POMCP + ASP full} achieves similar discounted return even with $2^{13}$ particles, while its performance becomes comparable (Figure \ref{fig:size_manypart_4rocks}) or slightly lower (Figure \ref{fig:size_manypart_8rocks}) than \emph{POMCP} when only $2^{11}$ simulations are used. Then, these experiments show that the use of learned heuristics in the rollout phase reduces the number of particles and online simulations required by POMCP, improving the planning performance even in unseen scenarios with longer planning horizon (on average 47 steps per execution) and larger action space (14 actions with 8 rocks).

We finally run a test in a very challenging rocksample domain, with $N=M=20$.
Here, \emph{POMCP} achieves discounted return $9.83 \pm 0.75$, while \emph{POMCP + ASP full} reaches $13.99 \pm 0.81$. \emph{POMCP + preferred} achieves $16.32 \pm 0.71$, but at a higher computational time per step, because the observation space becomes significantly large. In fact, it takes $34.29 \pm$\SI{7.50}{s}, while \emph{POMCP + ASP full} requires only $15.84 \pm$\SI{3.55}{s}. This shows that learned policy specifications are slightly less accurate and informative than handcrafted ones, as explained in former sections. However, in very large maps with very long planning horizons and many actions (26 actions available and 67 average steps executed with $M=20$ rocks), they reach a suitable tradeoff between computational effort and performance in POMCP.

\subsubsection{DESPOT}
\begin{figure}
    \centering
    \begin{subfigure}{0.45\textwidth}
    \includegraphics[width=\linewidth]{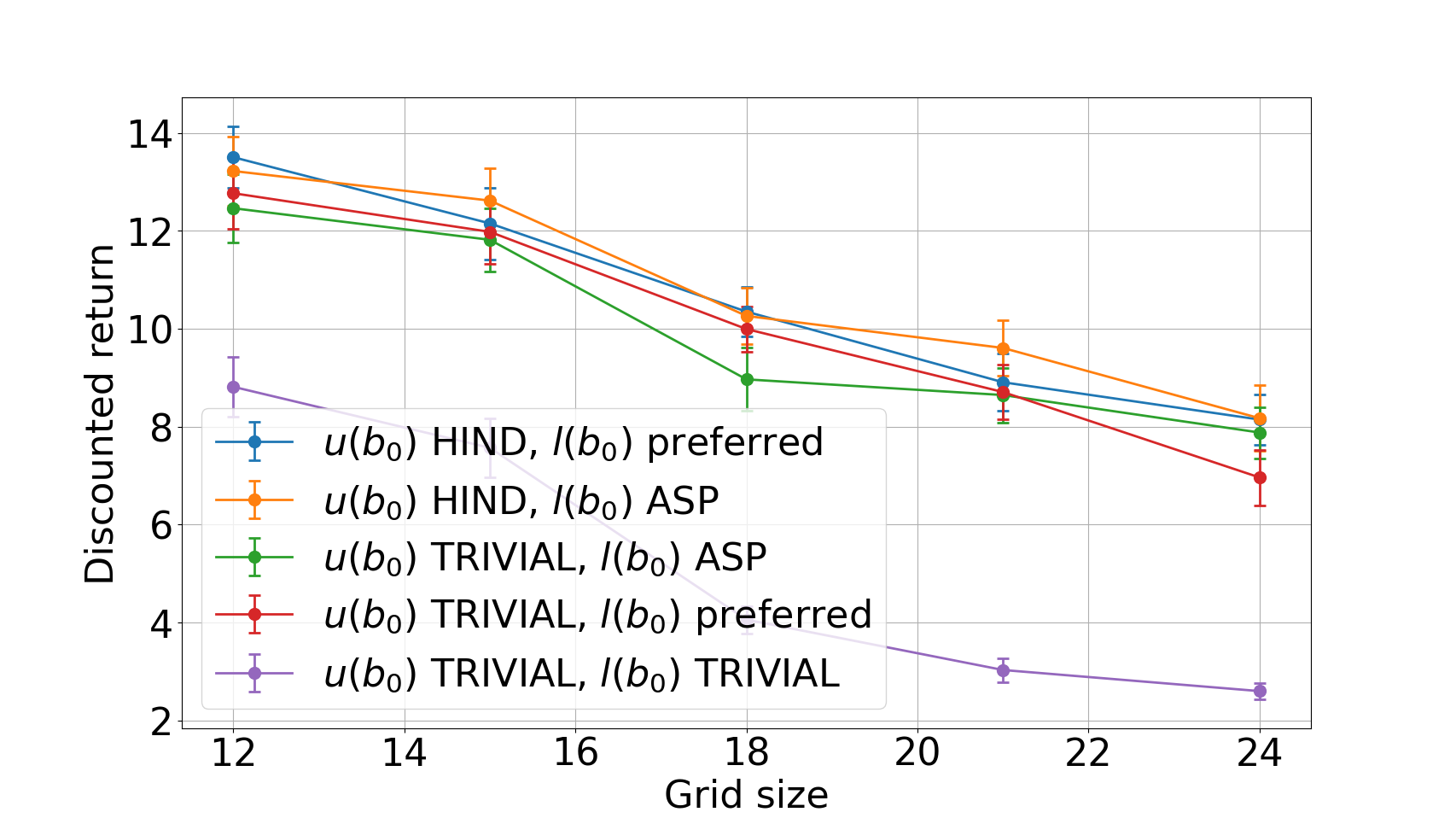}
    \caption{$M=4$ rocks.\label{fig:ret_4rocks}}
    \end{subfigure}
    \begin{subfigure}{0.45\textwidth}
    \includegraphics[width=\linewidth]{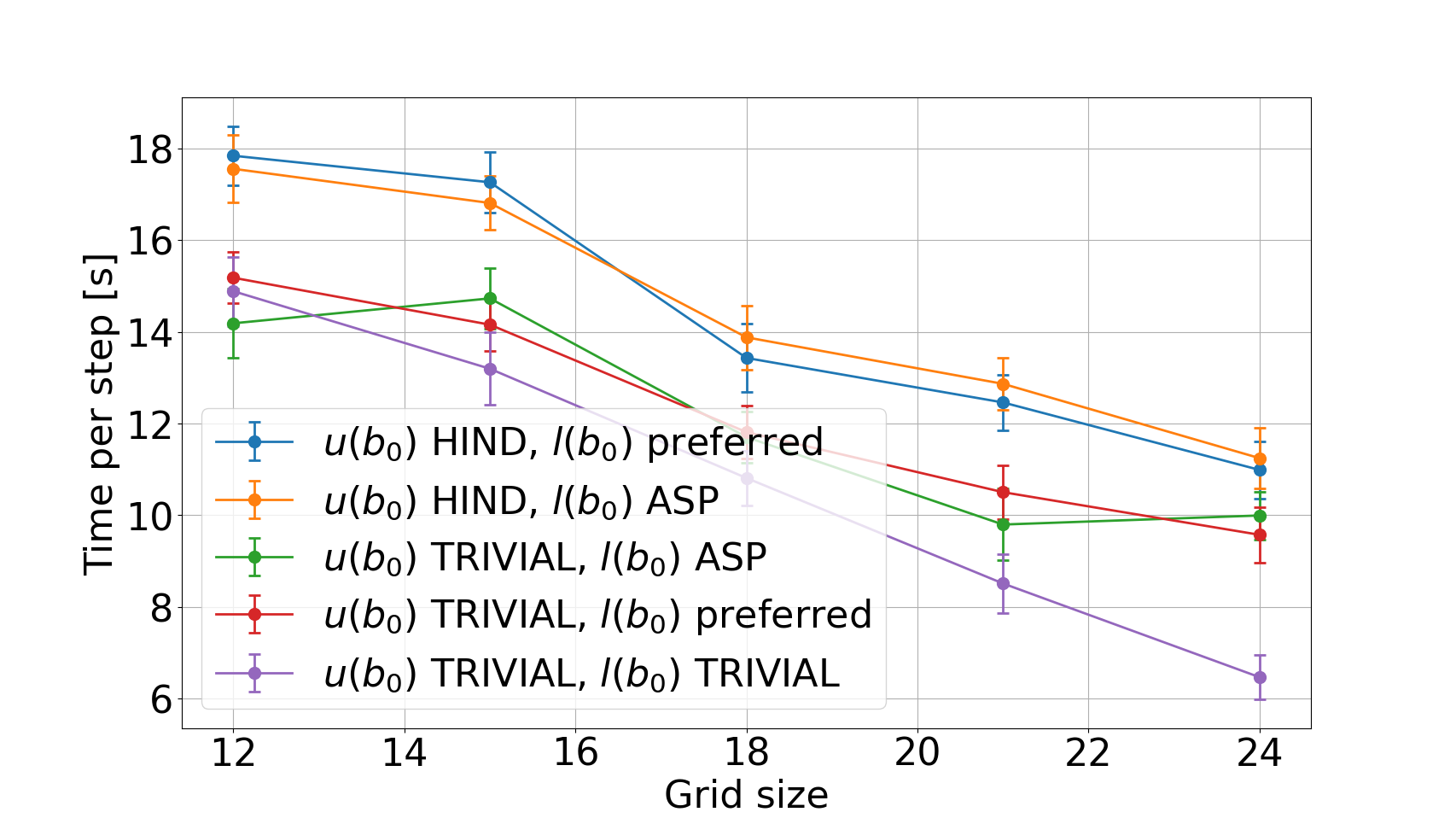}
    \caption{$M=8$ rocks.\label{fig:ret_8rocks}}
    \end{subfigure}\\
    \begin{subfigure}{0.45\textwidth}
    \includegraphics[width=\linewidth]{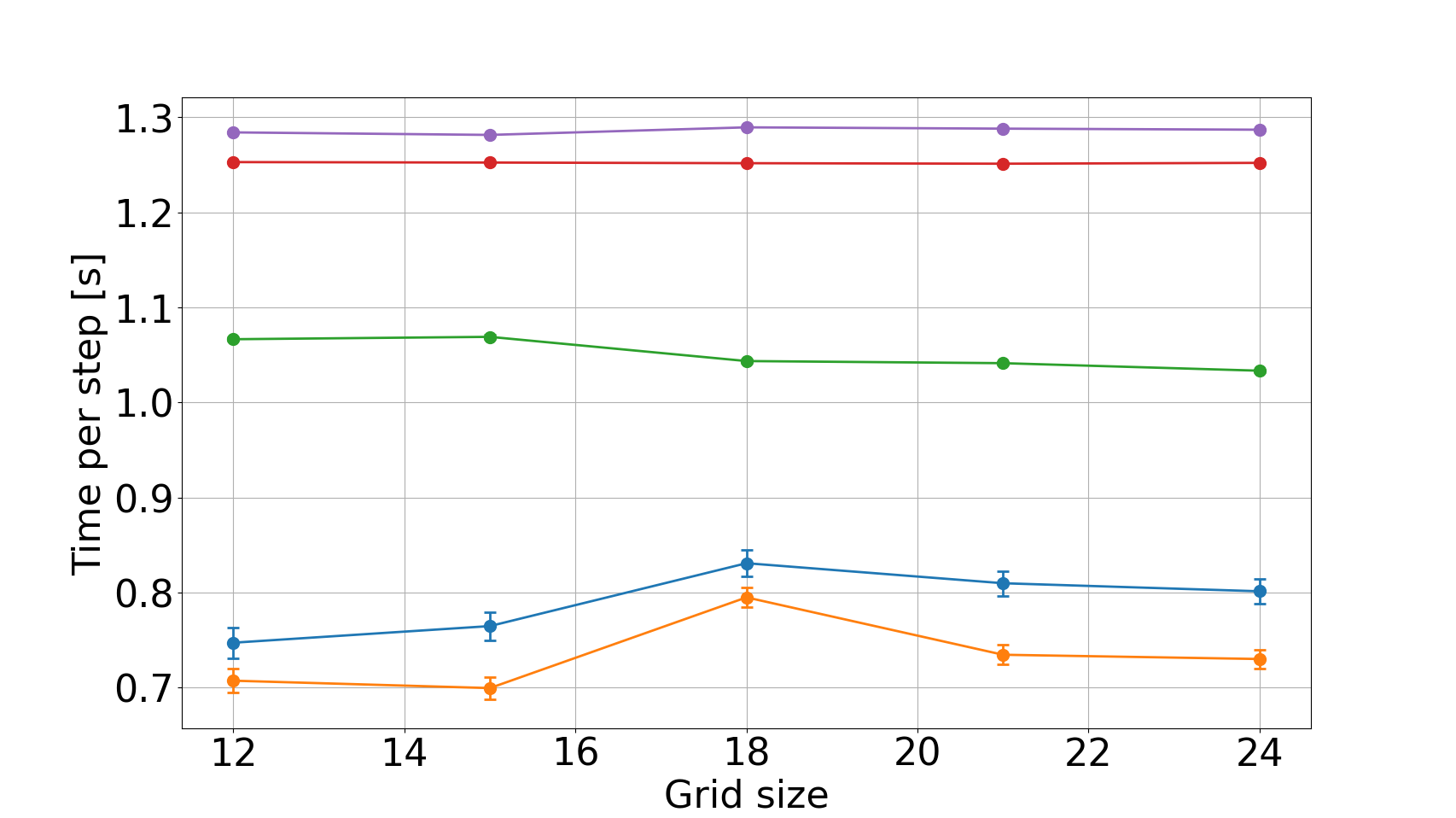}
    \caption{$M=4$ rocks.\label{fig:time_4rocks}}
    \end{subfigure}
    \begin{subfigure}{0.45\textwidth}
    \includegraphics[width=\linewidth]{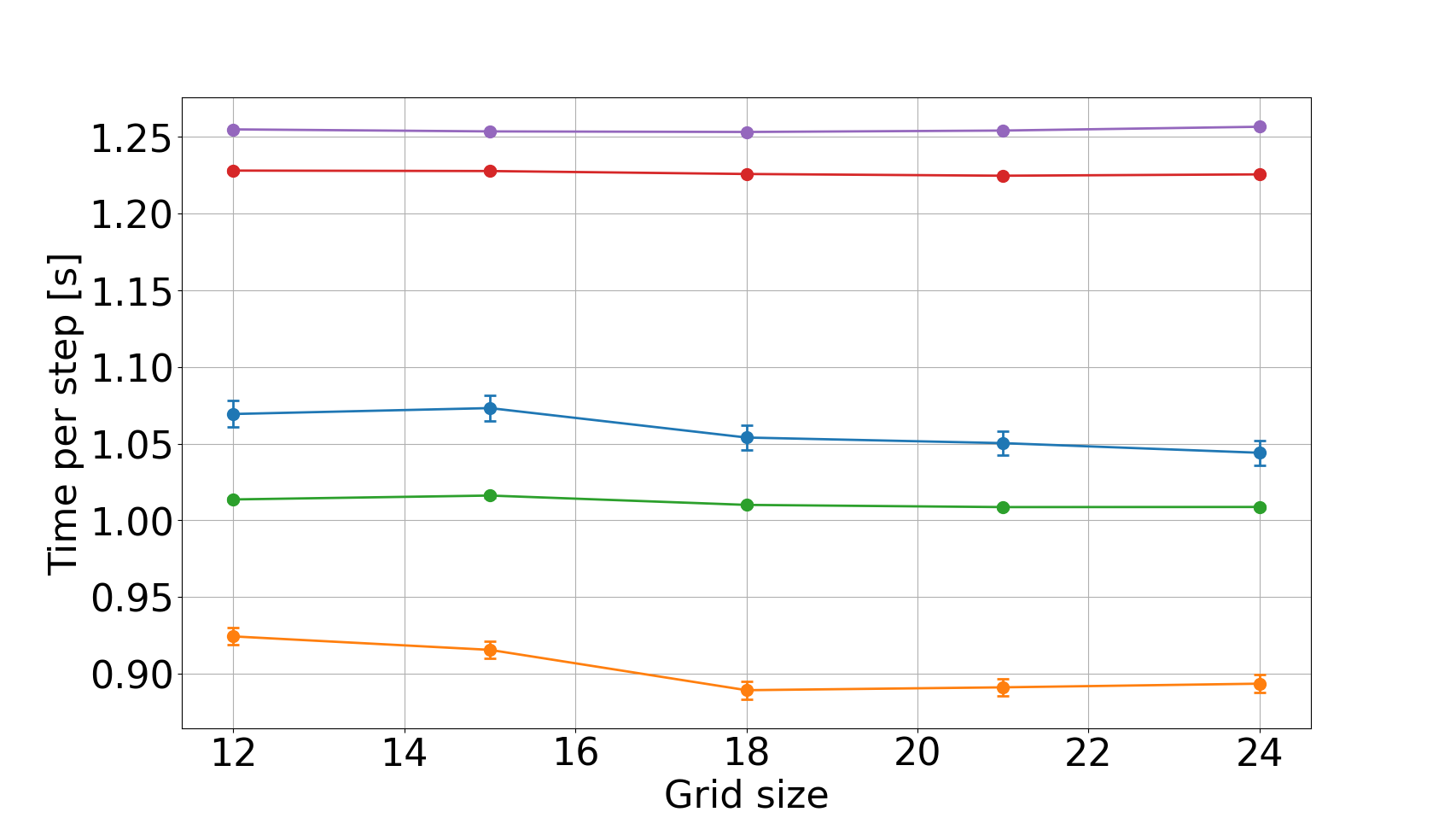}
    \caption{$M=8$ rocks.\label{fig:time_8rocks}}
    \end{subfigure}
    \caption{Discounted return (\textbf{top}) and computational time per step (\textbf{bottom}) with DESPOT \textbf{C++} (mean $\pm$ standard deviation) for rocksample in larger grids and with more rocks (\textbf{EXP-2}).}
    \label{fig:exp2_rs_despot}
\end{figure}
Following \shortciteA{ye2017despot}, in DESPOT we consider two different upper bounds $u(b_0)$:
\begin{itemize}
    \item \emph{HIND}, corresponding to the hindsight approximation proposed by (\shortciteR{yoon2008probabilistic}, see Section \ref{sec:met_despot});
    \item \emph{TRIVIAL}, corresponding to the uninformed bound $u(b_0) = \frac{R_{max}}{1-\gamma}$, with $R_{max}$ as the maximum task reward (+10 in rocksample). This is an important case study, because hindsight approximation may be computationally prohibitive to compute in large scenarios (see Section \ref{sec:met_despot}).
\end{itemize}
\noindent
We then evaluate 3 different lower bounds $l(b_0)$:
\begin{itemize}
    \item \emph{ASP}, where the default policy is selected according to the learned policy specifications;
    \item \emph{TRIVIAL}, where a default action is always chosen (moving east for rocksample);
    \item \emph{preferred}, where the default policy is selected according to the handcrafted heuristics used for POMCP.
\end{itemize}
\noindent
We then consider different combinations of lower and upper bounds in our experiments (from now on, we will use the syntax $l(b_0) + u(b_0)$ to refer to different planning combinations).
We keep the number of scenarios fixed to $K=500$ (the standard value used by \shortciteR{ye2017despot}), since in our experiments changing this parameter did not significantly impact the performance.

In Figures \ref{fig:ret_4rocks},\ref{fig:time_4rocks}, we show the performance with 4 rocks in larger grids, while results with 8 rocks are reported in Figures \ref{fig:ret_8rocks},\ref{fig:time_8rocks}. In general, DESPOT with \emph{HIND} upper bound achieves the highest performance, both with \emph{preferred} and \emph{ASP} lower bounds. Thus, when the upper bound is accurate, the solver is able to compute a good-quality plan. However, when the upper bound is uninformative (\emph{TRIVIAL}), the policy specifications to determine the lower bound become crucial. \emph{ASP} lower bound is able to achieve the same performance as the handcrafted \emph{preferred} one. On the contrary, DESPOT with \emph{TRIVIAL} lower bound is significantly worse, especially as the grid size increases to $N=24$. It is interesting to note that \emph{ASP} lower bound minimizes the computational time per step. In particular, from Figure \ref{fig:time_4rocks} (similar results can be observed from Figure \ref{fig:time_8rocks}) the combination \emph{ASP + TRIVIAL} (green curve) requires $\approx$\SI{1.1}{s} per step for all grid sizes, while \emph{preferred + TRIVIAL} requires $\approx$\SI{1.30}{s} (similarly to \emph{TRIVIAL + TRIVIAL}, i.e., the worst combination). Even when the optimal upper bound \emph{HIND} is used, \emph{ASP + HIND} requires less time than \emph{preferred + HIND}, especially with 8 rocks in Figure \ref{fig:time_8rocks} ($\approx$\SI{0.9}{s} vs. $\approx$\SI{1.05}{s}). 

We finally evaluate DESPOT in the challenging rocksample scenario with $N=M=20$.
Here, we could not compute the hindsight upper bound, since it caused memory overload on our machine, given the very large domain size. Hence, only the \emph{TRIVIAL} upper bound is considered.
Similarly to POMCP, \emph{ASP} lower bound achieves slightly worse discounted return than \emph{preferred} ($\approx 10.18 \pm 0.89$ vs. $\approx12.81 \pm 0.57$). However, learned heuristics require significantly less computational time per step ($\approx 1.43 \pm$\SI{0.07}{s} vs. $\approx2.22 \pm$\SI{0.25}{s}).

\subsubsection{AdaOPS and POMCPOW}
With AdaOPS and POMCPOW, we report similar experiments to the previous section, varying the number of rocks $M$ and the grid size $N$.
In particular, we implement ASP heuristics into the original code by \shortciteA{wu2021adaptive}, considering only \emph{HIND} and \emph{TRIVIAL} upper bounds, \emph{ASP} and \emph{TRIVIAL} lower bounds (since handcrafted lower bounds are not implemented in Julia). We then report POMCPOW\footnote{We do not perform action widening in POMCPOW, since the action space is discrete and this resulted in better performance in our experiments.} performance (in terms of discounted return) only as a baseline, since it performs significantly worse than AdaOPS.
\begin{figure}
    \centering
    \begin{subfigure}{0.45\textwidth}
    \includegraphics[width=\linewidth]{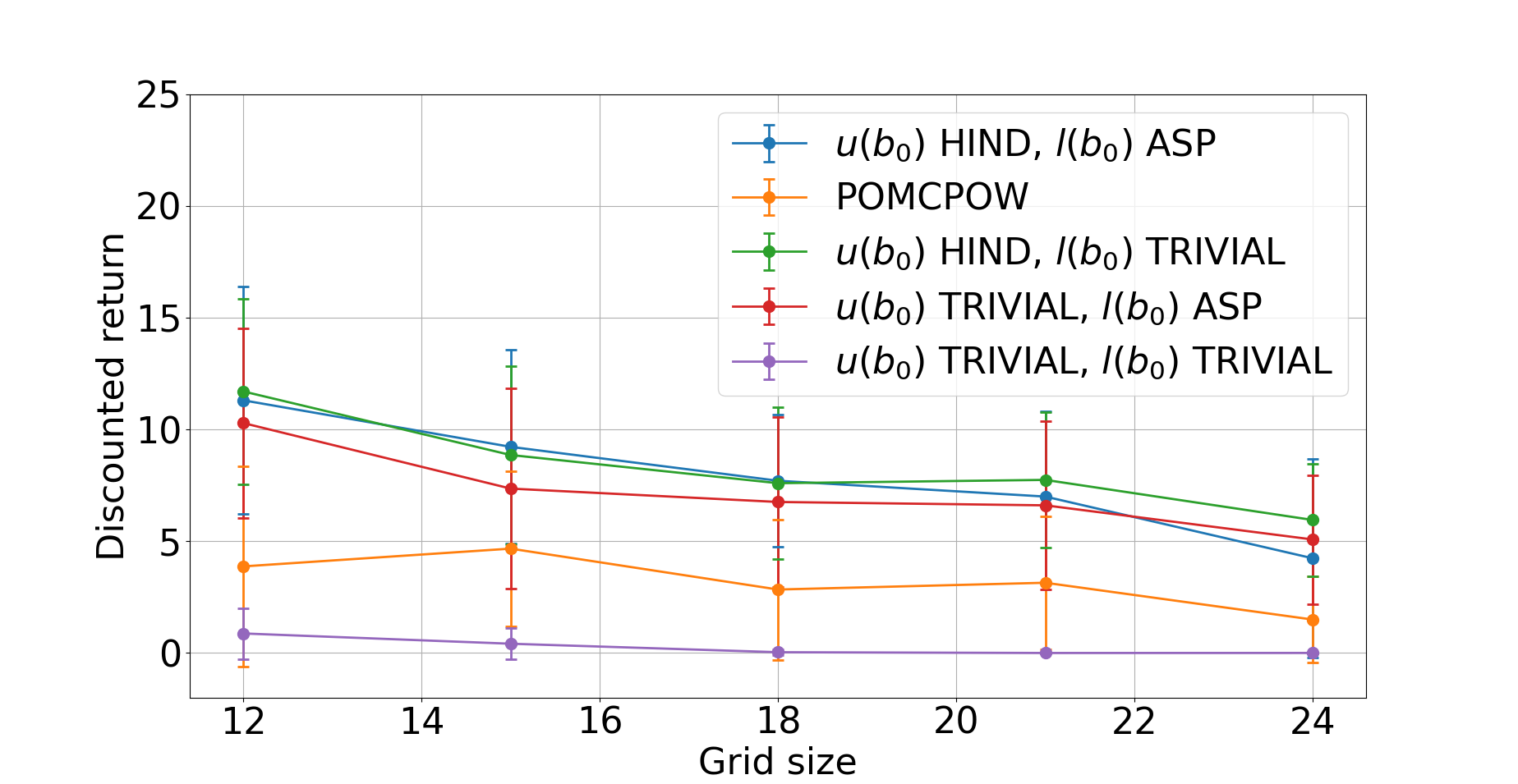}
    \caption{$M=4$ rocks.\label{fig:ada_ret_4rocks}}
    \end{subfigure}
    \begin{subfigure}{0.45\textwidth}
    \includegraphics[width=\linewidth]{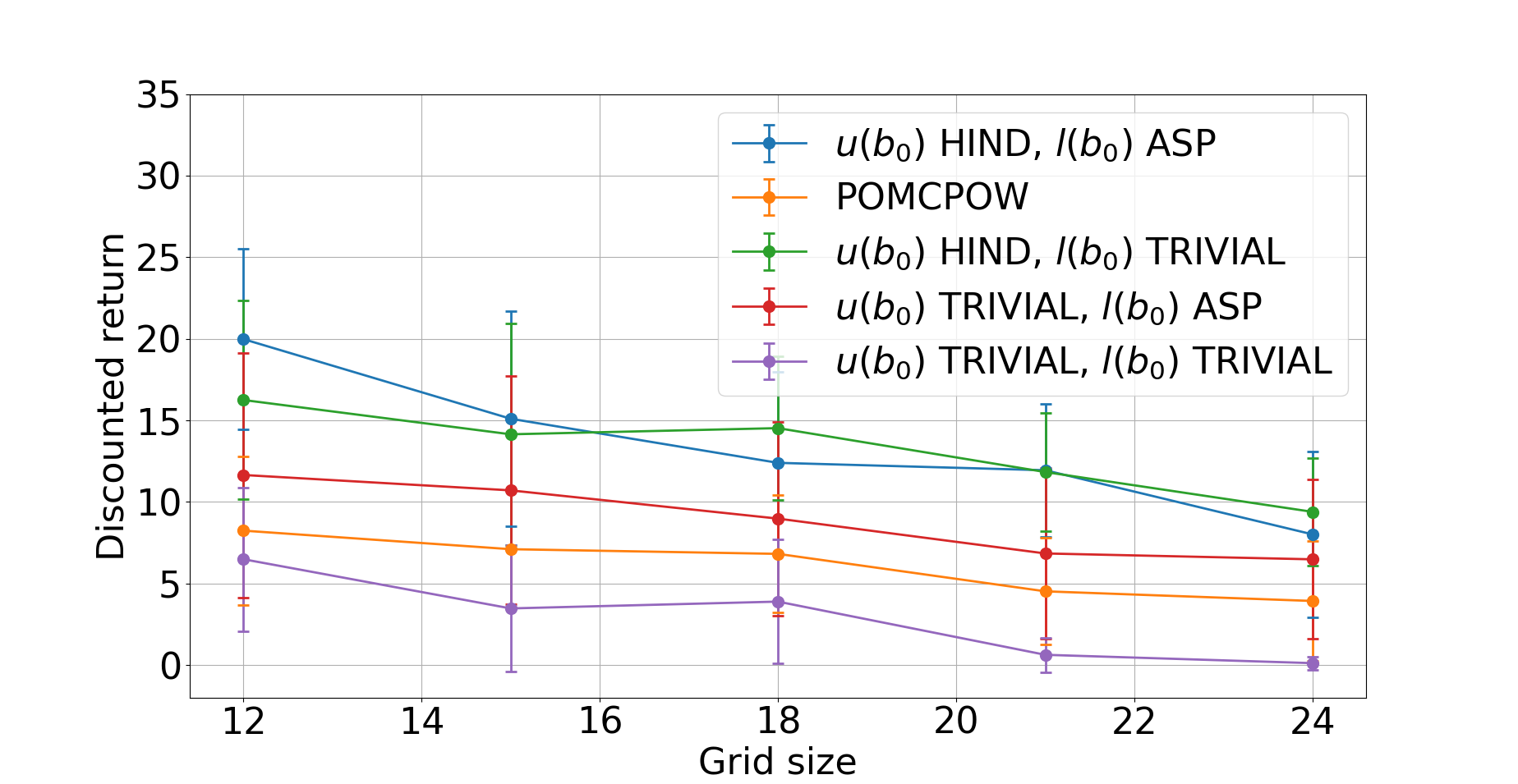}
    \caption{$M=8$ rocks.\label{fig:ada_ret_8rocks}}
    \end{subfigure}\\
    \begin{subfigure}{0.45\textwidth}
    \includegraphics[width=\linewidth]{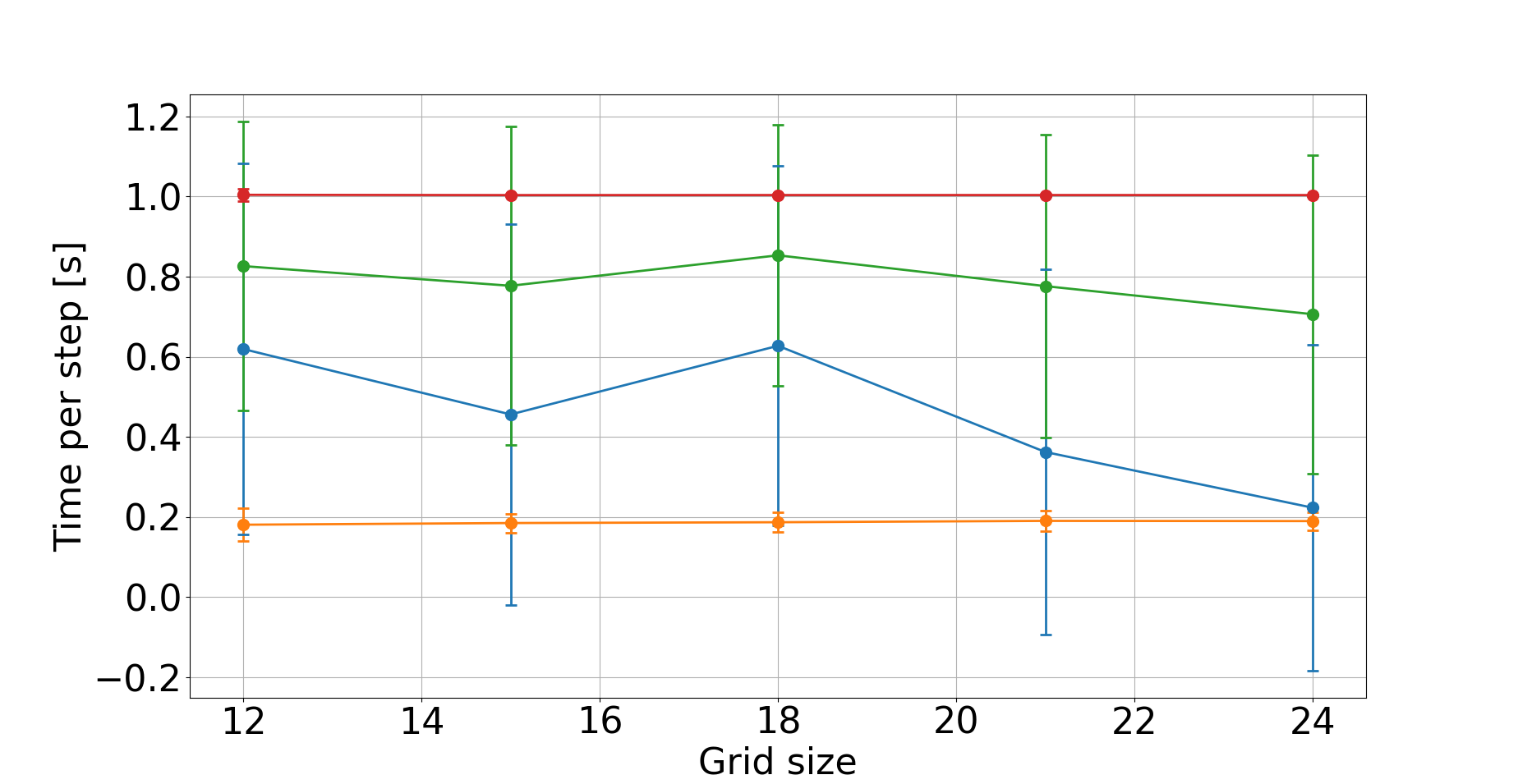}
    \caption{$M=4$ rocks.\label{fig:ada_time_4rocks}}
    \end{subfigure}
    \begin{subfigure}{0.45\textwidth}
    \includegraphics[width=\linewidth]{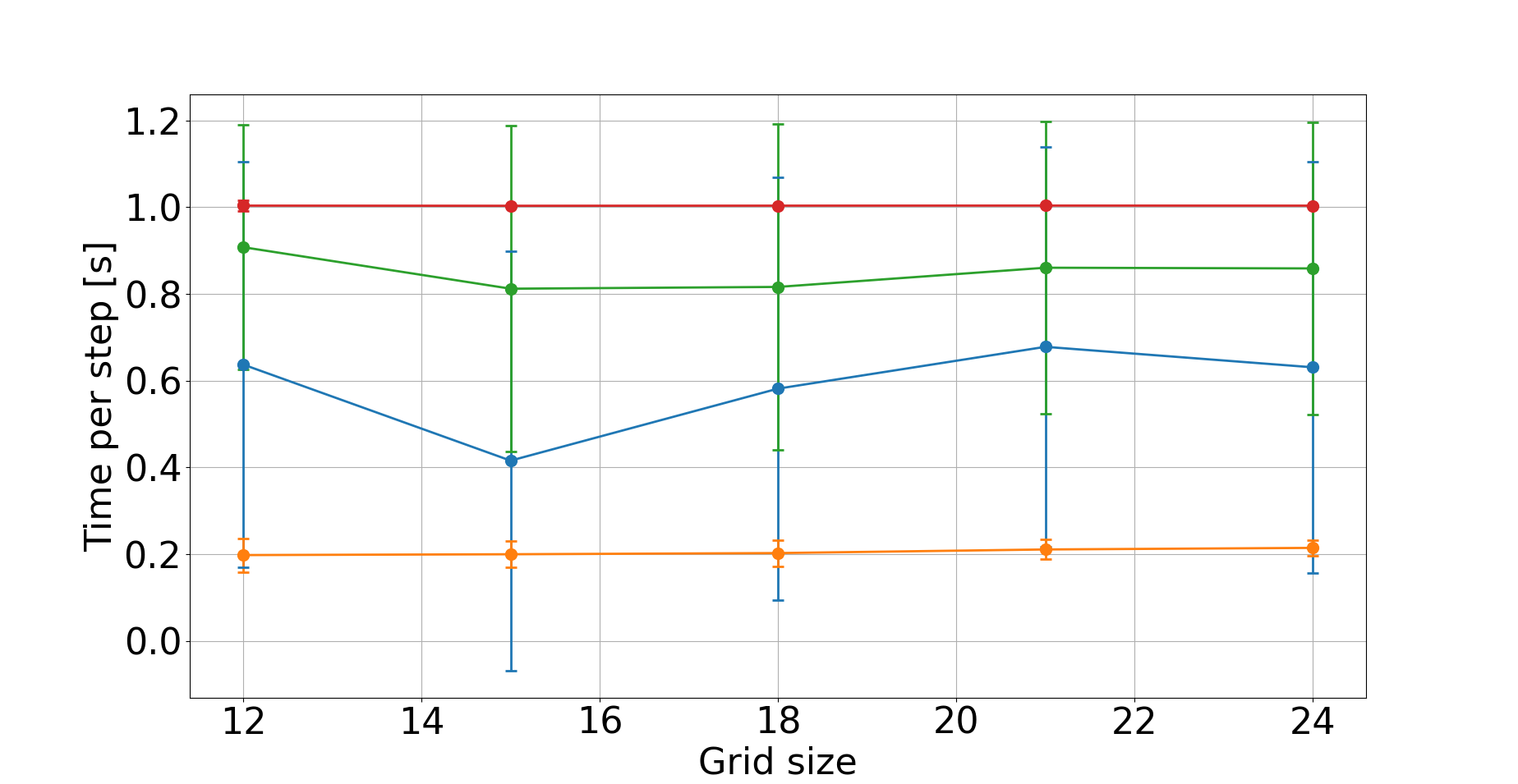}
    \caption{$M=8$ rocks.\label{fig:ada_time_8rocks}}
    \end{subfigure}
    \caption{Discounted return (\textbf{top}) and computational time per step (\textbf{bottom}) with AdaOPS and POMCPOW (mean $\pm$ standard deviation) for rocksample in larger grids and with more rocks (\textbf{EXP-2}).}
    \label{fig:exp2_rs_adaops}
\end{figure}

Figure \ref{fig:exp2_rs_adaops} shows the performance of AdaOPS.
The results confirm the ones evidenced in C++ with DESPOT.
The use of learned policy specifications for the default action and the lower bound (\emph{ASP}) does not significantly impact the discounted return achieved with the optimal \emph{HIND} upper bound, both with 4 and 8 rocks (Figures \ref{fig:ada_ret_4rocks}-\ref{fig:ada_ret_8rocks}, respectively). However, \emph{ASP} lower bound improves the performance when the uninformed upper bound (\emph{TRIVIAL)} is used. In terms of computational time per step (Figure \ref{fig:ada_time_4rocks} for 4 rocks, Figure \ref{fig:ada_time_8rocks} for 8 rocks), the \emph{ASP} lower bound improves the efficiency of the planning process with the good upper bound \emph{HIND}. With the uninformed upper bound \emph{TRIVIAL}, the solver requires the same computational effort with any lower bound, hence the curve for \emph{TRIVIAL + TRIVIAL} is superimposed to \emph{ASP + TRIVIAL} in Figures \ref{fig:ada_time_4rocks}-\ref{fig:ada_time_8rocks}.
In the $20\times 20$ grid with 20 rocks, when \emph{HIND} upper bound\footnote{In Julia, the hindsight approximation can be computed on the available hardware, differently from C++ code, probably due to the different software implementation.} is used \emph{ASP} does not bring any benefit to AdaOPS. However, when the uninformed upper bound \emph{TRIVIAL} is adopted, \emph{ASP} lower bound achieves significantly higher discounted return than \emph{TRIVIAL} ($10.81 \pm 8.30$ vs. $0.03 \pm 0.54$).

We also perform further experiments with Julia version of DESPOT, confirming the C++ results. Hence, they are reported only in the public repository for brevity.

\subsection{Generalization to Unseen Scenarios (\textbf{EXP-2}) - Pocman}
In the pocman domain we assess generalization by increasing the grid size to $17\times 19$, as proposed by \shortciteA{silver2010monte,ye2017despot}. This further increases the planning horizon with respect to the $10\times 10$ grid size used to generate training traces in Sections \ref{sec:learning_res}-\ref{sec:exp1_pocman}, rising from 73 to 85 steps per execution, on average. As a comparison, in the rocksample domain on a $20\times 20$ grid with 20 rocks, the agent takes on average 67 steps.

We also evaluate the performance when some parameters of the task are changed, i.e., the aggressivity and number of ghosts and the percentage of food pellets. This may significantly change the behavior of the agent with respect to the training setting, thus the quality and generality of policy specifications is fundamental.

We only report results in C++ with POMCP and DESPOT, since this domain is not implemented in Julia.

\subsubsection{POMCP}\label{sec:pomcp_pocman}
\begin{figure}
    \centering
    \begin{subfigure}{0.45\textwidth}
    \includegraphics[width=\linewidth]{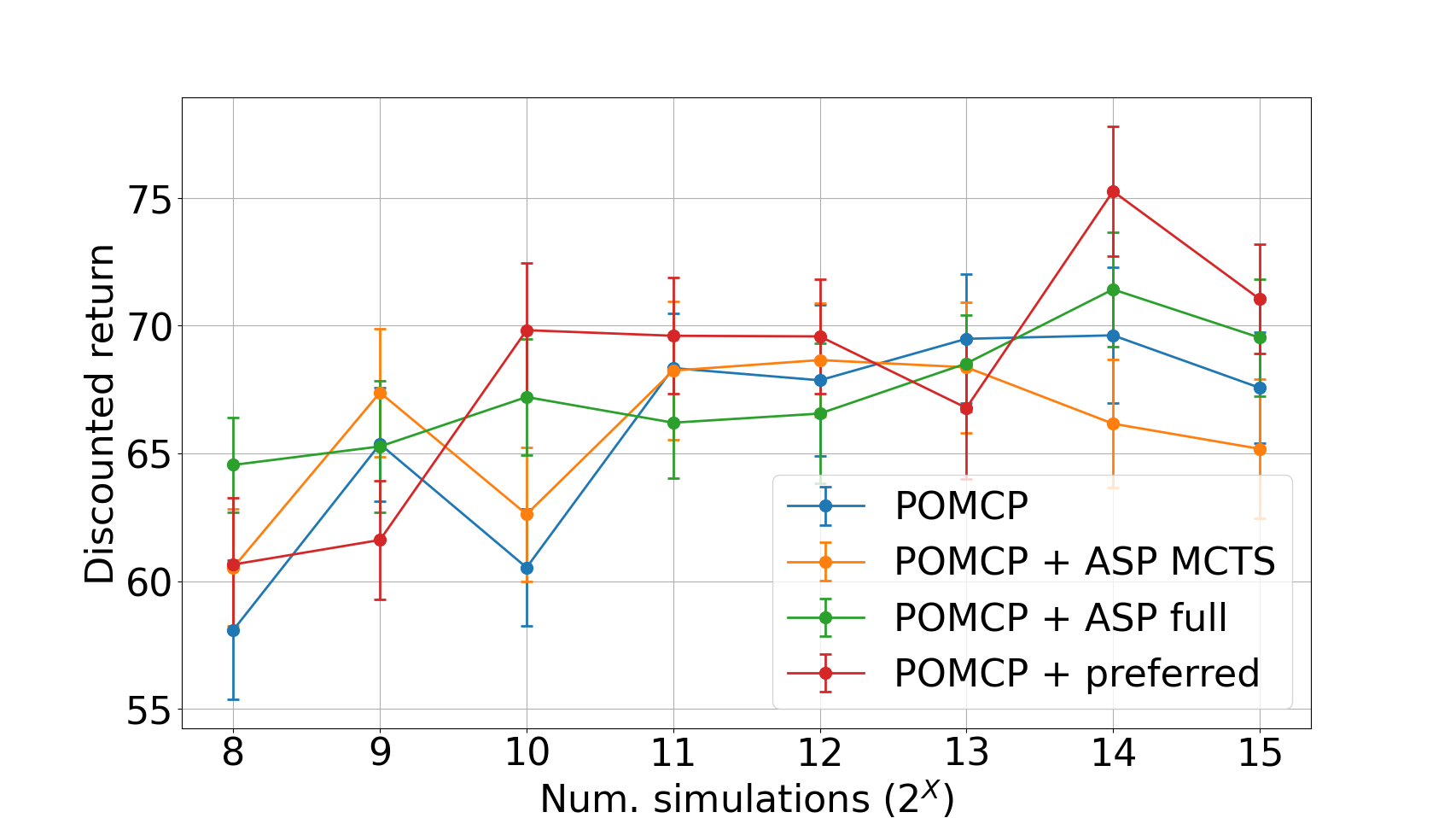}
    \caption{\label{fig:pocman_std}}
    \end{subfigure}
    \begin{subfigure}{0.45\textwidth}
    \includegraphics[width=\linewidth]{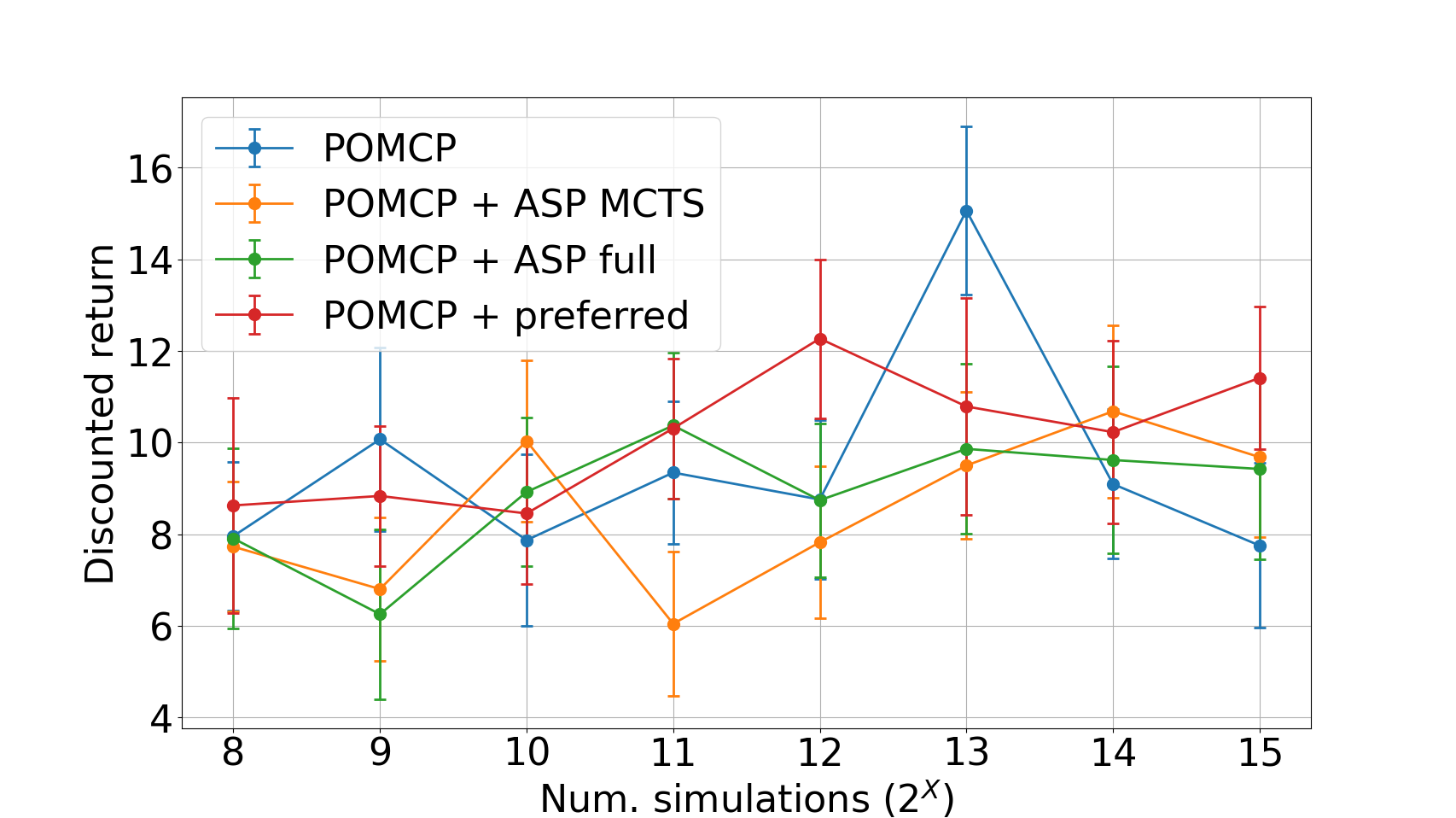}
    \caption{\label{fig:pocman_hard}}
    \end{subfigure}\\
    \begin{subfigure}{0.45\textwidth}
    \includegraphics[width=\linewidth]{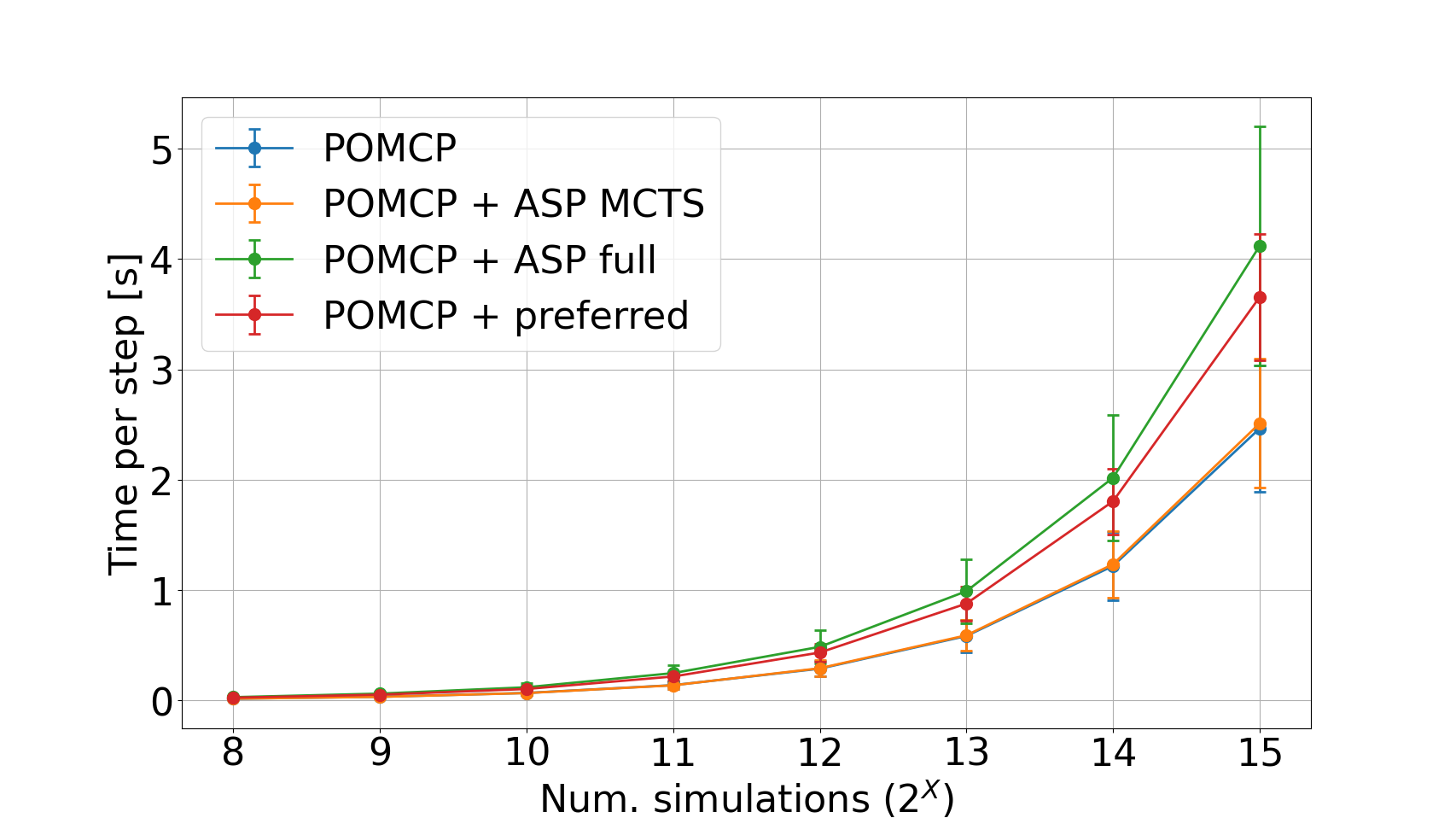}
    \caption{$17\times 19$ grid, $G=4, \rho_g = 75\%, \rho_f = 50\%$.\label{fig:time_pocman_std}}
    \end{subfigure}
    \begin{subfigure}{0.45\textwidth}
    \includegraphics[width=\linewidth]{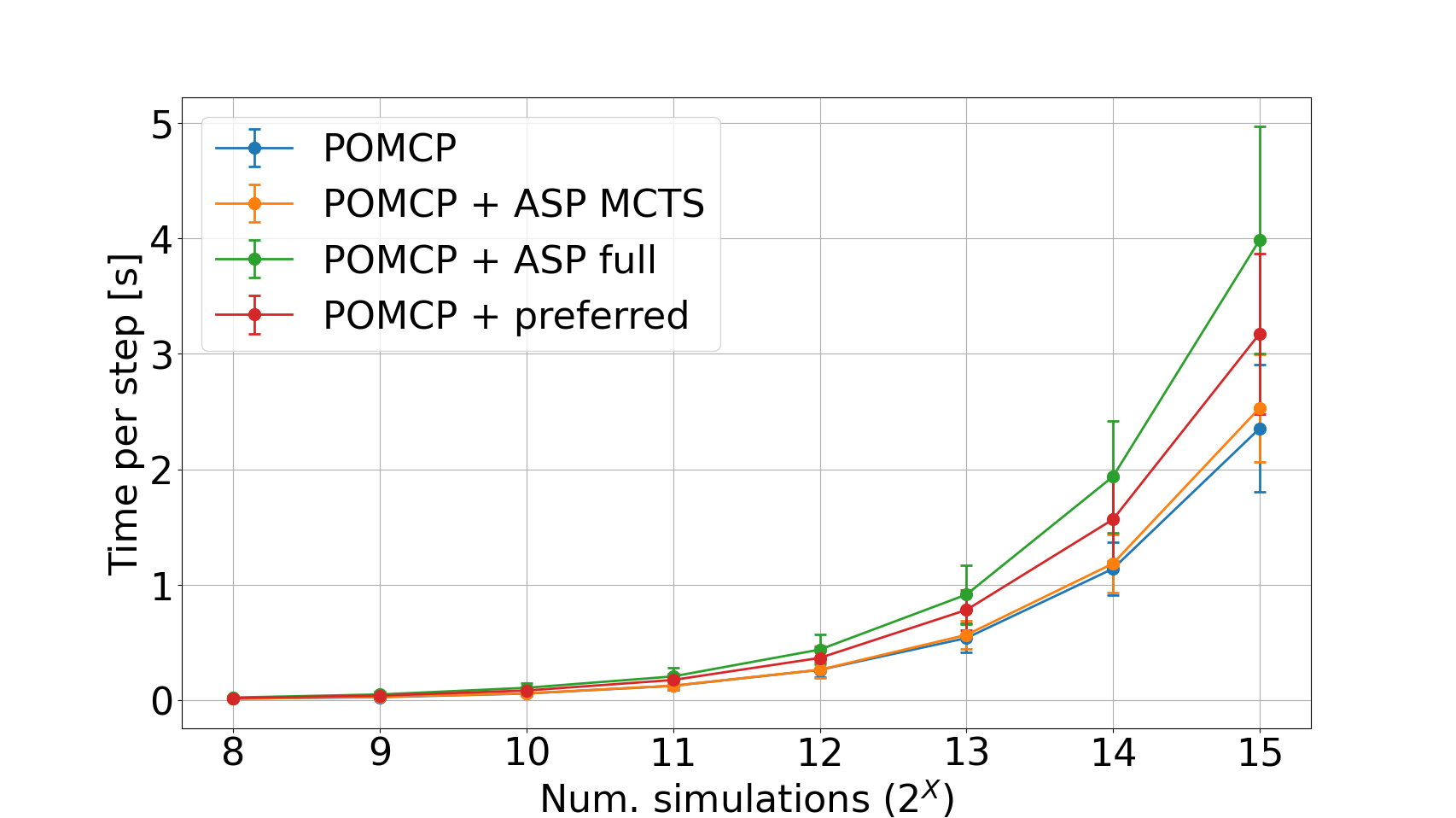}
    \caption{$17\times 19$ grid, $G=4, \rho_g = 100\%, \rho_f = 20\%$.\label{fig:time_pocman_hard}}
    \end{subfigure}
    \caption{Discounted return (\textbf{top}) and computational time per step (\textbf{bottom}) with POMCP (mean $\pm$ standard deviation) for pocman in more challenging domains (\textbf{EXP-2}).}
    \label{fig:exp2_pocman_pomcp}
\end{figure}
In Figures \ref{fig:pocman_std}, \ref{fig:time_pocman_std}, we show POMCP performance in the larger map ($17\times 19$) with $G=4$ ghosts, varying the number of online simulations and particles. In Figures \ref{fig:pocman_hard}-\ref{fig:time_pocman_hard}, we further complicate the setting for the agent, increasing the probability that ghosts chase the agent up to $\rho_g=100\%$ and decreasing the probability to find food pellets to $\rho_f=20\%$. In both cases, there is no significant difference between the performance of different POMCP versions, i.e., with or without learned and handcrafted heuristics. This confirms the results observed in Section \ref{sec:exp1_pocman} in the training setting, proving that even good policy heuristics have small impact on POMCP performance in a scenario with very long planning horizon. It is still interesting to note, however, that learned heuristics do not deteriorate the performance of the planner.

\subsubsection{DESPOT}
\begin{table}
    \centering
    \caption{\textbf{EXP-2}: Pocman results with DESPOT (mean $\pm$ standard deviation) in nominal conditions ($\rho_f=50\%$, $\rho_g=75\%$).}    
    \begin{tabular}{c|c|c|c|c|c}
        \textbf{$l(b_0)$} & \textbf{$u(b_0)$} & \multicolumn{2}{c|}{\textbf{$10\times 10, G=2$}} & \multicolumn{2}{c}{\textbf{$17\times 19, G=4$}}\\ 
        \midrule
        & & \textbf{Disc. ret.} & \textbf{Time/step [s]}& \textbf{Disc. ret.} & \textbf{Time/step [s]}\\  
        \midrule
        ASP & HIND & $55.64 \pm 2.41$ & $1.09 \pm 0.001$ & $70.00 \pm 2.82$ & $1.12 \pm 0.002$\\
        Preferred & HIND & $52.37 \pm 3.52$ & $1.10 \pm 0.002$ & $72.56 \pm 2.04$ & $1.12 \pm 0.002$\\
        ASP & TRIVIAL & $53.42 \pm 3.32$ & $1.09 \pm 0.002$ & $64.91 \pm 2.20$ & $1.11 \pm 0.002$\\
        Preferred & TRIVIAL & $51.52 \pm 3.30$ & $1.11 \pm 0.002$ & $71.31 \pm 2.03$ & $1.12 \pm 0.002$\\
        TRIVIAL & TRIVIAL & $2.67 \pm 3.16$ & $1.42 \pm 0.003$ & $-8.63 \pm 5.04$ & $1.35 \pm 0.004$\\
        \midrule
    \end{tabular}
    \label{tab:despot_pocman_std}
\end{table}
Table \ref{tab:despot_pocman_std} shows the performance of DESPOT in different maps and with standard values for ghost aggressivity ($\rho_g = 75\%$) and food probability ($\rho_f = 50\%$), considering the different upper and lower bound combinations explained for rocksample.
In general, using \emph{ASP} as a lower bound leads to similar performance as the best configuration (\emph{preferred + HIND}) even when the \emph{TRIVIAL} upper bound is used, both in terms of computational time per step and discounted return. It is interesting to note the significant improvement with respect to \emph{TRIVIAL + TRIVIAL} in both maps, thus showcasing the high quality and the fundamental role of learned heuristics for planning performance.

\begin{table}
    \centering
    \caption{\textbf{EXP-2}: Pocman results with DESPOT (mean $\pm$ standard deviation) in challenging conditions ($\rho_f=20\%$, $\rho_g=100\%$).}    
    \begin{tabular}{c|c|c|c|c|c}
        \textbf{$l(b_0)$} & \textbf{$u(b_0)$} & \multicolumn{2}{c|}{$10\times 10, G=2$} & \multicolumn{2}{c}{$17\times 19, G=4$}\\ 
        \midrule
        & & \textbf{Disc. ret.} & \textbf{Time/step [s]}& \textbf{Disc. ret.} & \textbf{Time/step [s]}\\  
        \midrule
        ASP & HIND & $17.56 \pm 3.53$ & $1.07 \pm 0.001$ & $10.63 \pm 1.68$ & $1.09 \pm 0.002$\\
        Preferred & HIND & $12.98 \pm 3.07$ & $1.08 \pm 0.002$ & $16.30 \pm 1.58$ & $1.12 \pm 0.002$\\
        ASP & TRIVIAL & $11.15 \pm 3.17$ & $1.07 \pm 0.001$ & $13.28 \pm 1.63$ & $1.08 \pm 0.002$\\
        Preferred & TRIVIAL & $14.88 \pm 3.90$ & $1.08 \pm 0.002$ & $14.67 \pm 1.66$ & $1.11 \pm 0.002$\\
        TRIVIAL & TRIVIAL & $9.29 \pm 4.33$ & $1.33 \pm 0.002$ & $-4.84 \pm 2.82$ & $1.35 \pm 0.004$\\
        \midrule
    \end{tabular}
    \label{tab:despot_pocman_challenging}
\end{table}
Results in Table \ref{tab:despot_pocman_challenging} partly confirm these observations when the parameters of the task become more challenging ($\rho_f=20\%$, $\rho_g=100\%$).
In fact, in the smaller map\footnote{We remark that even though the map size is the same as the scenario for generating learning traces, the task is more challenging, because of the fewer food pellets and the highest probability to get caught by ghosts (as evidenced by the significant difference of discounted returns between Tables \ref{tab:despot_pocman_std}-\ref{tab:despot_pocman_challenging}).} the \emph{ASP} lower bound performs better than \emph{preferred} when \emph{HIND} upper bound is used ($17.56 \pm 3.53$ vs. $12.98 \pm 3.07$), but is worse with the \emph{TRIVIAL} upper bound ($11.15 \pm 3.17$ vs. $14.88 \pm 3.90$).
Conversely, in the larger map \emph{ASP} lower bound achieves similar performance as \emph{preferred} with the \emph{TRIVIAL} upper bound ($13.28 \pm 1.63$ vs. $14.67 \pm 1.66$), but worse performance in combination with \emph{HIND} upper bound ($10.63 \pm 1.68$ vs. $16.30 \pm 1.58$). Overall, learned \emph{ASP} heuristics always achieve significantly superior discounted return than \emph{TRIVIAL + TRIVIAL}, thus showing the quality of learned policy specifications.

\subsection{Influence of Quality and Quantity of Example Traces on Planning Performance (\textbf{EXP-3}-\textbf{EXP-4})}
In Section \ref{sec:learning_res}, we have assumed that 1000 traces of execution are generated from a \emph{high-quality} planner, i.e., POMCP with $2^{15}$ particles and online simulations.
In very complex scenarios, this may be infeasible due to the required time to generate traces (in POMCP, the number of particles affects the computational time per step; moreover, larger scenarios require longer planning horizons, hence, more steps of computation from all solvers). In addition, the agent may not be able to acquire useful and sufficient experience in a limited number of scenarios, hence some actions and task configuration may be underrepresented in the training traces (e.g., with very large action or state spaces).

In this section, we then investigate the impact of the quality and quantity of training example traces on the quality of learned policy heuristics. Specifically, we study the effect on the planning performance when: i) low-quality example traces are generated (\textbf{EXP-3}), and ii) fewer example traces are available (\textbf{EXP-4}).

\subsubsection{Effect of Low-Quality Examples (\textbf{EXP-3})}
\begin{figure}
    \centering
    \begin{subfigure}{0.45\textwidth}
    \centering
    \includegraphics[width=\linewidth]{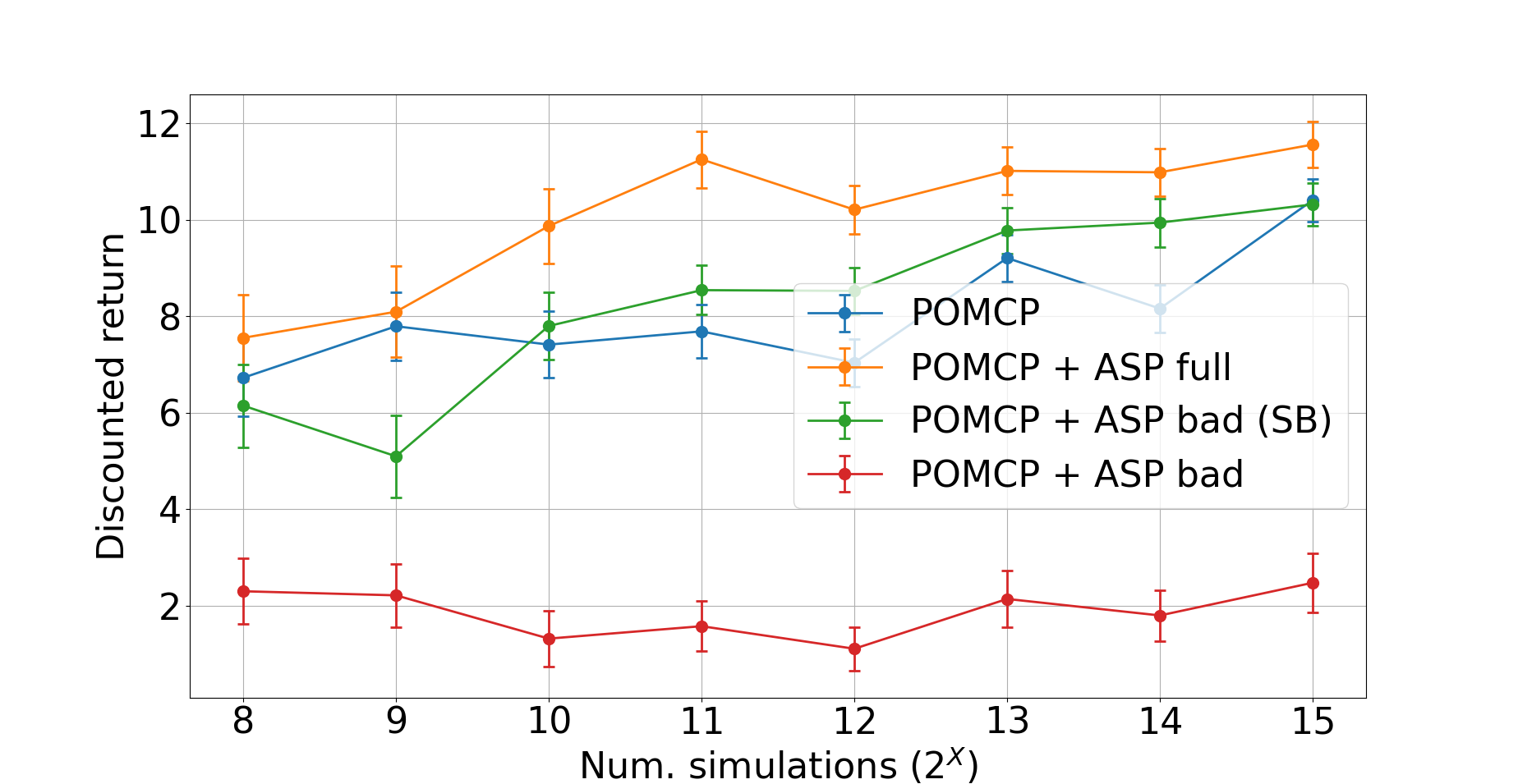}
    \caption{\label{fig:part_4rocks_bad}}
    \end{subfigure}
    \begin{subfigure}{0.45\textwidth}
    \centering
    \includegraphics[width=\linewidth]{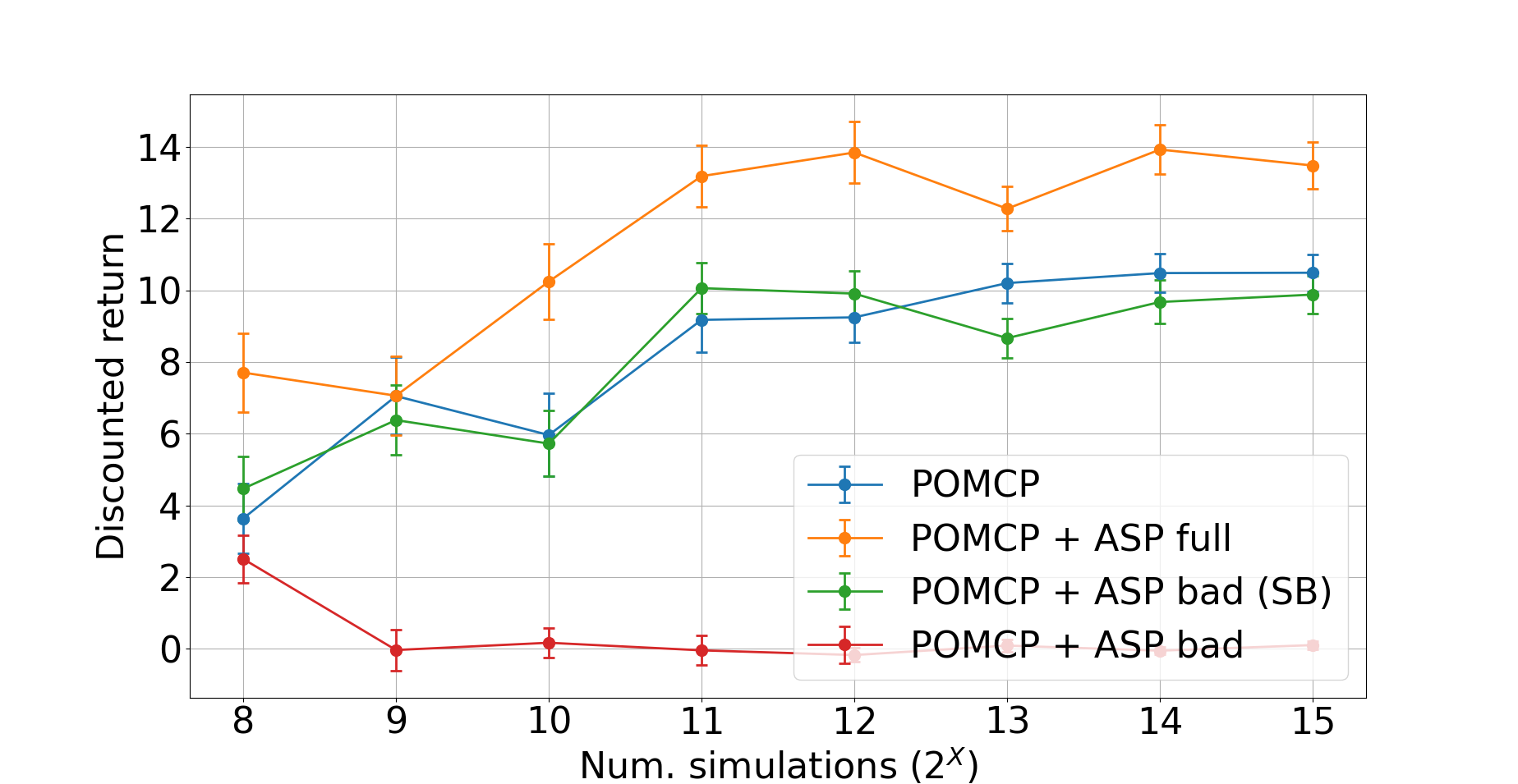}
    \caption{\label{fig:part_8rocks_bad}}
    \end{subfigure}\\
    \begin{subfigure}{0.45\textwidth}
    \centering
    \includegraphics[width=\linewidth]{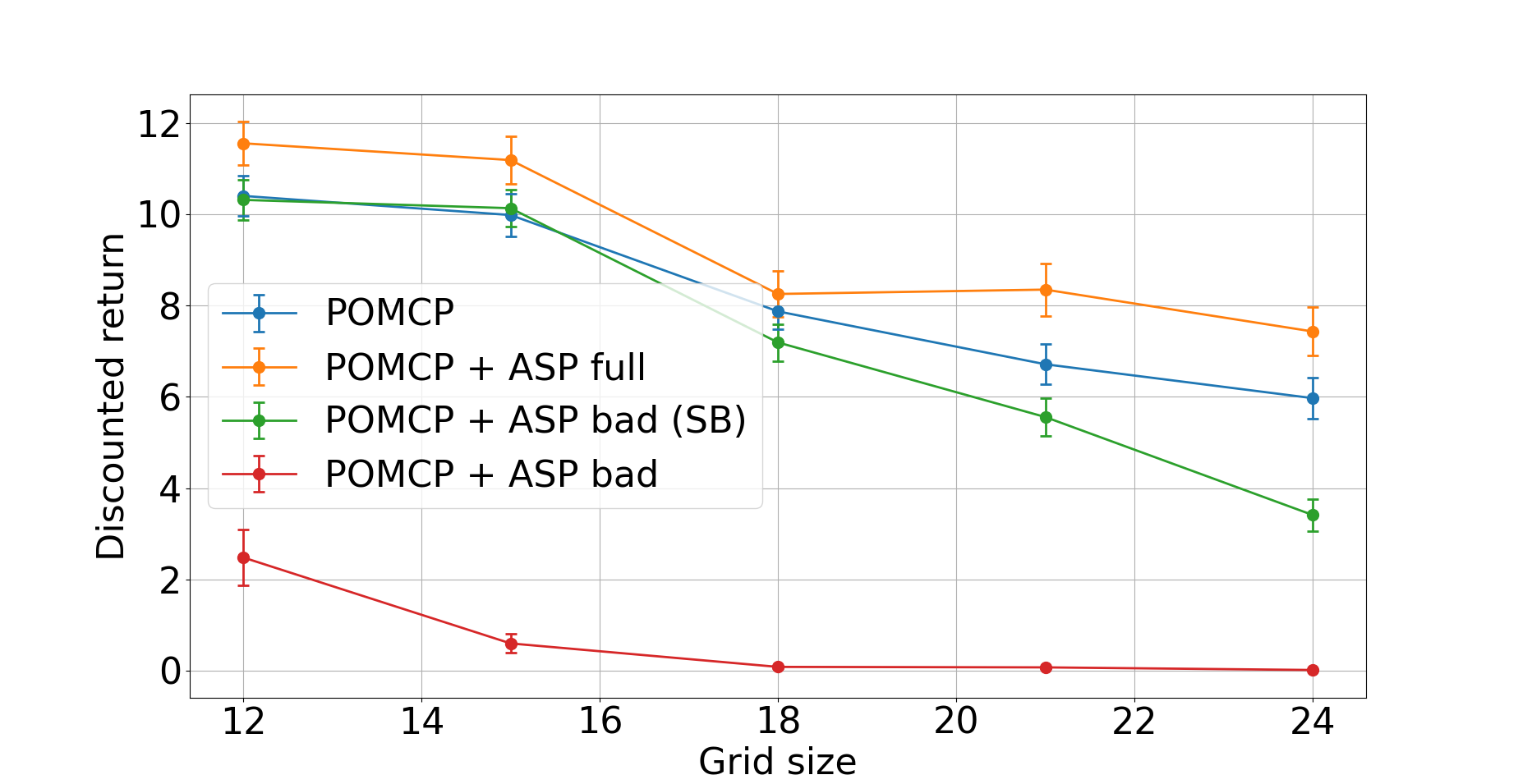}
    \caption{$M=4$ rocks.\label{fig:size_15part_4rocks_bad}}
    \end{subfigure}
    \begin{subfigure}{0.45\textwidth}
    \centering
    \includegraphics[width=\linewidth]{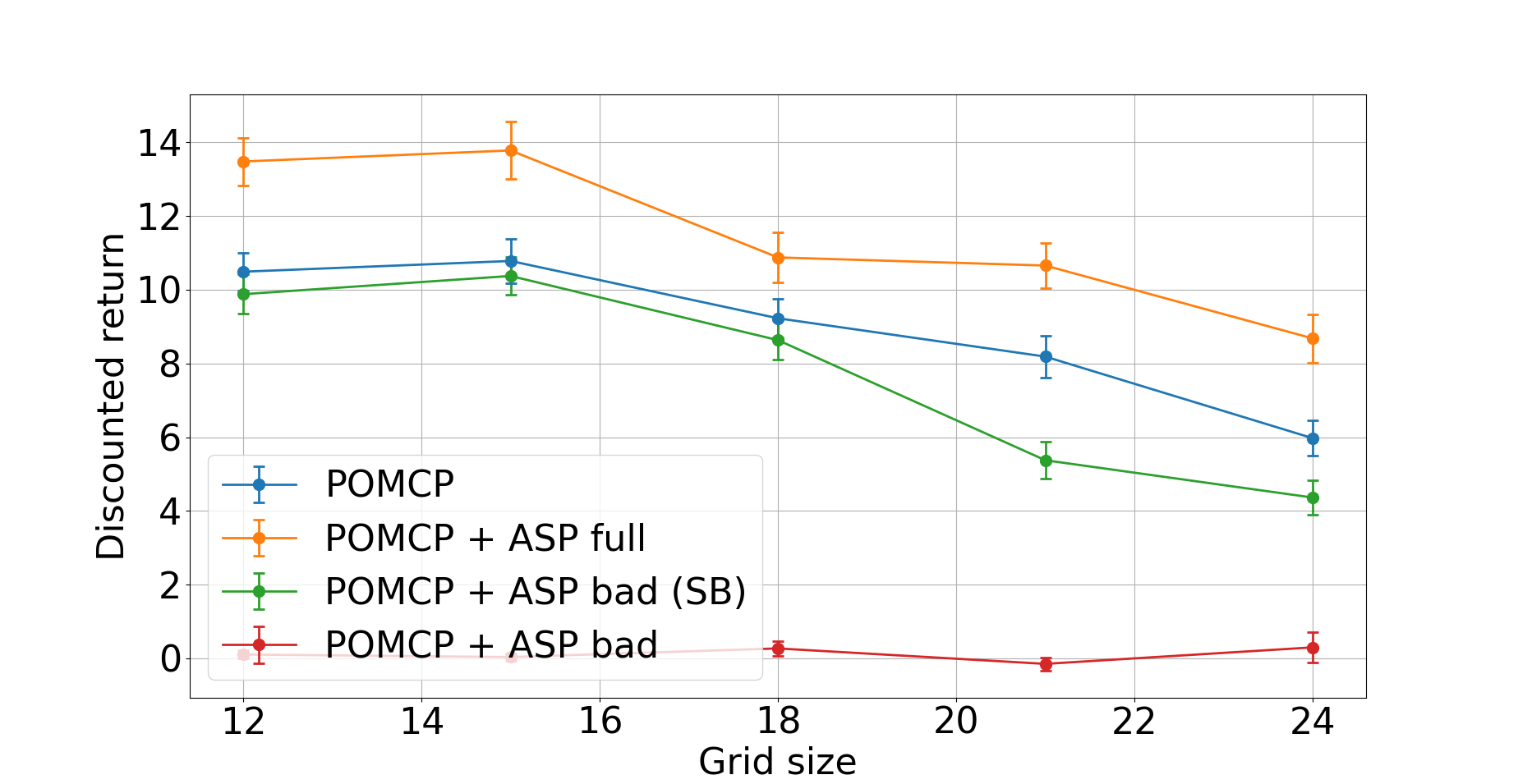}
    \caption{$M=8$ rocks.\label{fig:size_15part_8rocks_bad}}
    \end{subfigure}
    \caption{Results (mean $\pm$ standard deviation of the discounted return) for rocksample in POMCP with policy heuristics learned from bad examples (\textbf{EXP-3}). \textbf{Top}: $12 \times 12$ grid with different number of simulations. \textbf{Bottom}: $2^{15}$ simulations with increasing grid size.}
    \label{fig:rs_bad_pomcp}
\end{figure}
We generate 1000 training traces with $2^{10}$ particles in POMCP. From Figures \ref{fig:part_4rocks}, \ref{fig:part_minipocman}, this is where the discounted return achieved by pure POMCP (blue curves) drops significantly in both rocksample and pocman.
In this way, we learn the following ASP heuristics for rocksample:
\begin{align}
    \label{eq:rs_h_bad}
    &\stt{east :- target(R), delta\_x(R,D), D} \geq \stt{1.}\\
    \nonumber&\stt{west :- target(R), delta\_x(R,D), D} \leq \stt{-1.}\\
    \nonumber&\stt{north :- target(R), delta\_y(R,D), D} \geq \stt{1.}\\
    \nonumber&\stt{south :- target(R), delta\_y(R,D), D} \leq \stt{-1.}\\
    \nonumber&\stt{target(R) :- dist(R,D), not sampled(R), D}\leq \stt{2.}\\
    \nonumber&\stt{check(R) :- guess(R,V), not sampled(R), dist(R,D), D} \leq \stt{0, V} \leq \stt{40.}\\
    \nonumber&\stt{sample(R) :- target(R), dist(R,D), D} \leq \stt{0, not sampled(R), guess(R,V), V} \geq \stt{70.}\\
    \nonumber&\stt{exit :- guess(R,V), V} \leq \stt{40, num\_sampled(N), N} \geq \stt{25.}
\end{align}
\noindent
With respect to good heuristics in Equation \eqref{eq:rs_h}, axioms for \stt{exit} and \stt{target(R)} depend on a smaller subset of the semantic features. In fact, \stt{exit} only depends on the number of sampled rocks, hence the agent is pushed towards terminating the task whenever 1/4 rocks have been sampled. A rock is chosen as a target only depending on its distance, ignoring its expected value. This affects the validity of axioms for actions of motion which, in turn, depend on \stt{target(R)}. 
As a consequence, the tests here reported not only analyze the influence of the quality of examples on the planning performance, but also the impact of \emph{bad definition of environmental features $\pazocal{F}$}. In fact, though they can be easily retrieved from the POMDP definition (transition and reward maps), in very complex domains some high-level domain concepts may be still missed.

In the pocman domain, the learned heuristics are instead the same, hence we report results with the bad heuristics only for rocksample.
In POMCP, we want to verify that our soft policy guidance methodology (Section \ref{sec:met_pomcp}) preserves the performance of POMCP when bad heuristics are learned, thanks to the non-null weights for selecting actions in Equation \eqref{eq:prob_rollout}.
For this reason, we consider 4 different solver configurations:
\begin{itemize}
    \item \emph{POMCP} and \emph{POMCP + ASP full}, with good policy heuristics from Equation \eqref{eq:rs_h}, as for \textbf{EXP-1}-\textbf{EXP-2};
    \item \emph{POMCP + ASP bad (SB)}, where bad policy specifications in Equation \eqref{eq:rs_h_bad} are implemented both in rollout and UCT, following the soft policy guidance approach proposed in Section \ref{sec:met_pomcp};
    \emph{POMCP + ASP bad}, where the methodology in Section \ref{sec:met_pomcp} is modified such that the probability of selecting rollout actions not entailed by bad policy specifications is set to 0. In other words, the \emph{soft policy guidance approach} is omitted from our methodology.
\end{itemize} 
Results in Figure \ref{fig:rs_bad_pomcp} highlight the importance of the soft policy guidance approach.
In fact, \emph{POMCP + ASP bad (SB)} performs nearly as \emph{POMCP} with any number of rocks, simulations and grid sizes (a slightly more pronounced decrease is observed in Figures \ref{fig:size_15part_4rocks_bad}, \ref{fig:size_15part_8rocks_bad}, where the planning horizon increases with the grid size).
On the contrary, \emph{POMCP + ASP bad} performs dramatically worse than other solvers, since rollout actions are pruned away according to bad policy specifications.
Clearly, \emph{POMCP + ASP full} (with good heuristics) performs better.

We perform a similar analysis in DESPOT (C++): we consider both the \emph{TRIVIAL} and \emph{HIND} upper bounds, and evaluate the difference in performance when good and bad ASP heuristics are adopted in the lower bound (\emph{ASP} and \emph{ASP bad}, respectively).
\begin{figure}
    \centering
    \begin{subfigure}{0.45\textwidth}
    \centering
    \includegraphics[width=\linewidth]{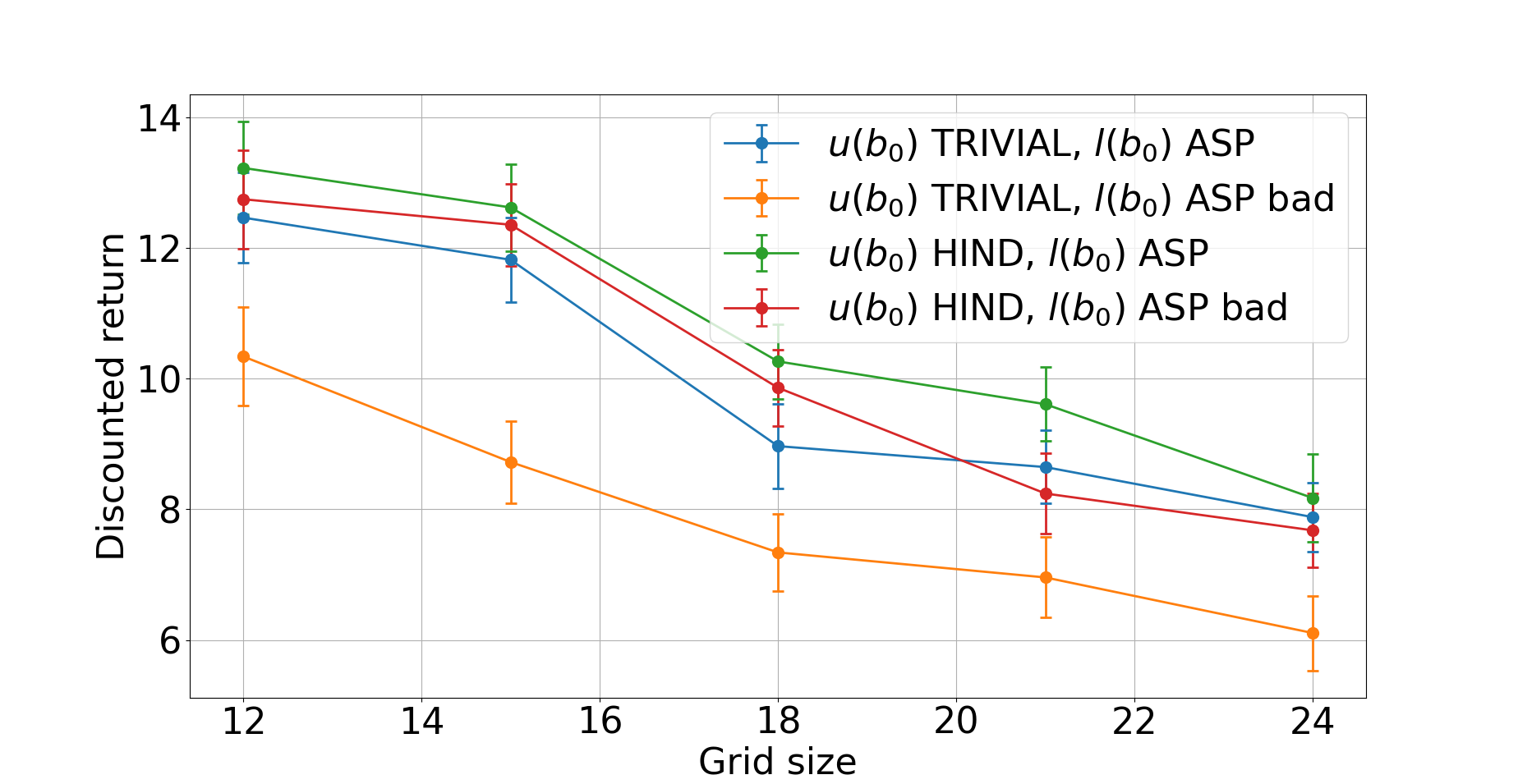}
    \caption{$M=4$ rocks\label{fig:despot_4rocks_bad}}
    \end{subfigure}
    \begin{subfigure}{0.45\textwidth}
    \centering
    \includegraphics[width=\linewidth]{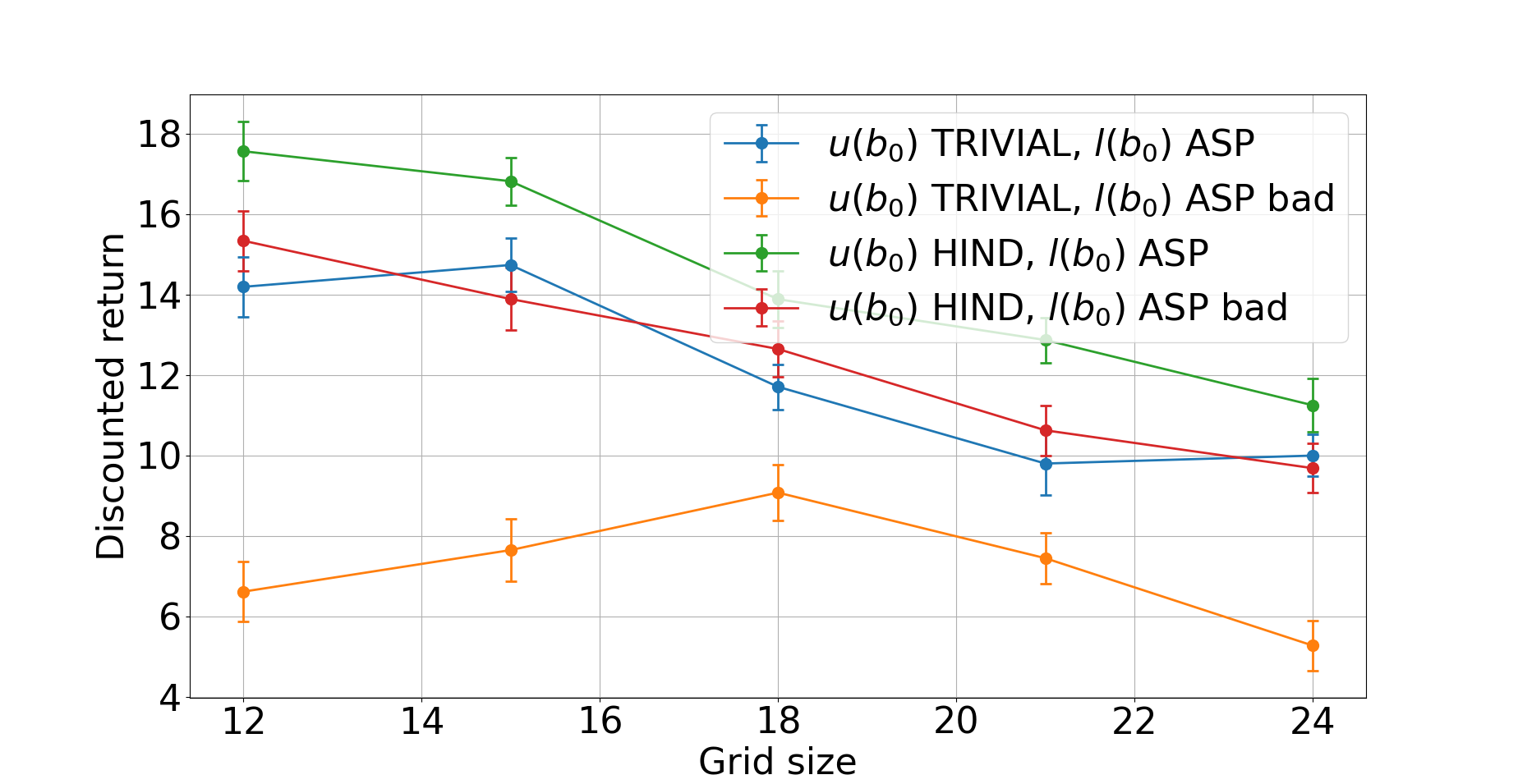}
    \caption{$M=8$ rocks\label{fig:despot_8rocks_bad}}
    \end{subfigure}
    \caption{Discounted return (mean $\pm$ standard deviation) with DESPOT (C++) for rocksample on grids with different sizes, with policy heuristics learned from bad examples (\textbf{EXP-3}).}
    \label{fig:rs_despot_bad}
\end{figure}
Results reported in Figure \ref{fig:rs_despot_bad} show that when the informed \emph{HIND} upper bound is used, the effect of bad policy heuristics is neglectable. However, the combination \emph{ASP bad + TRIVIAL} performs significantly worse than others. In fact, DESPOT strongly depends on the quality of heuristics in the bounds, hence it is not inherently robust to bad or incomplete specifications.

\subsubsection{Effect of Few Examples (\textbf{EXP-4})}
\begin{figure}
    \centering
    \begin{subfigure}{0.45\textwidth}
    \centering
    \includegraphics[width=\linewidth]{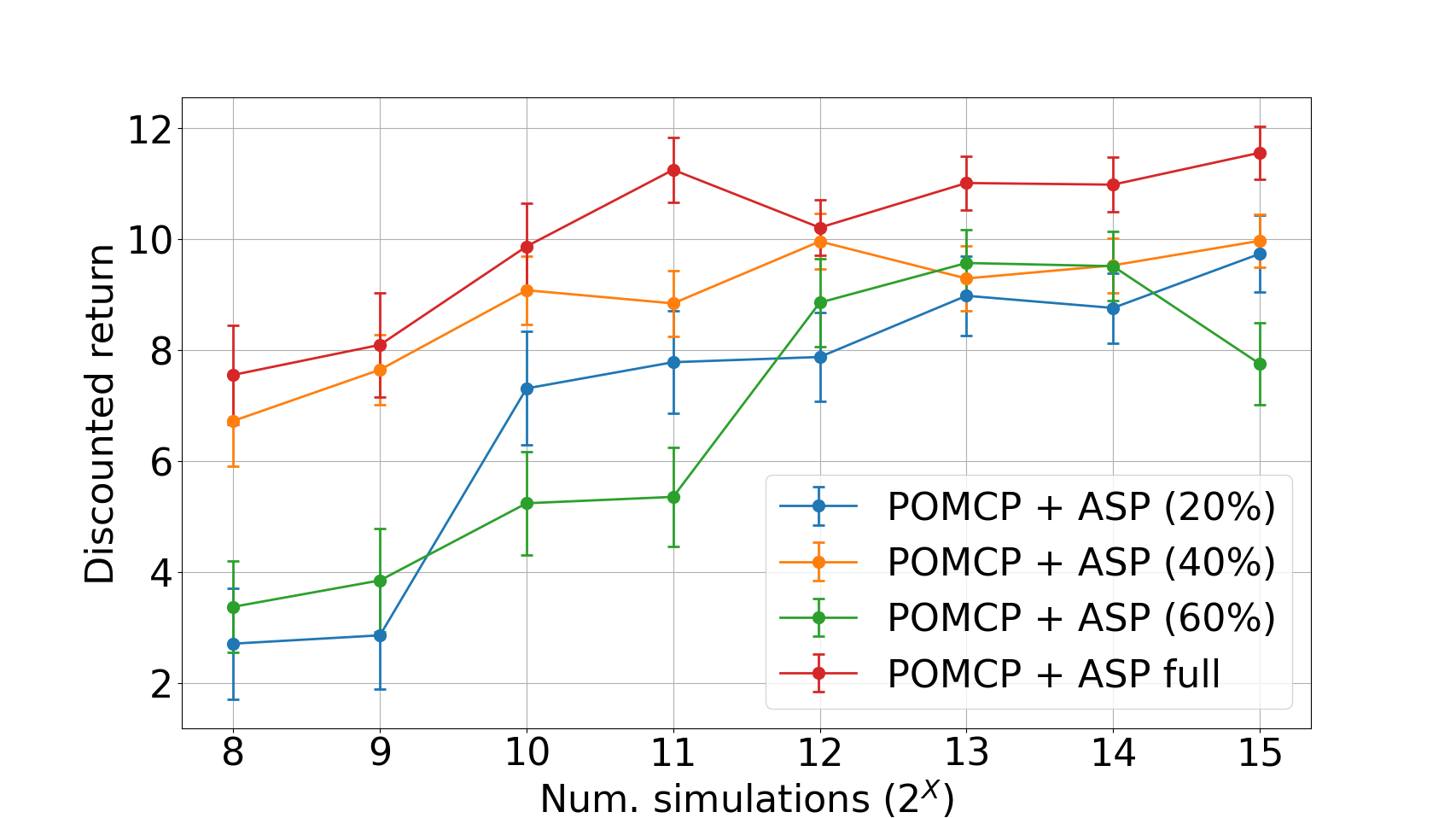}
    \caption{\label{fig:part_4rocks_perc}}
    \end{subfigure}
    \begin{subfigure}{0.45\textwidth}
    \centering
    \includegraphics[width=\linewidth]{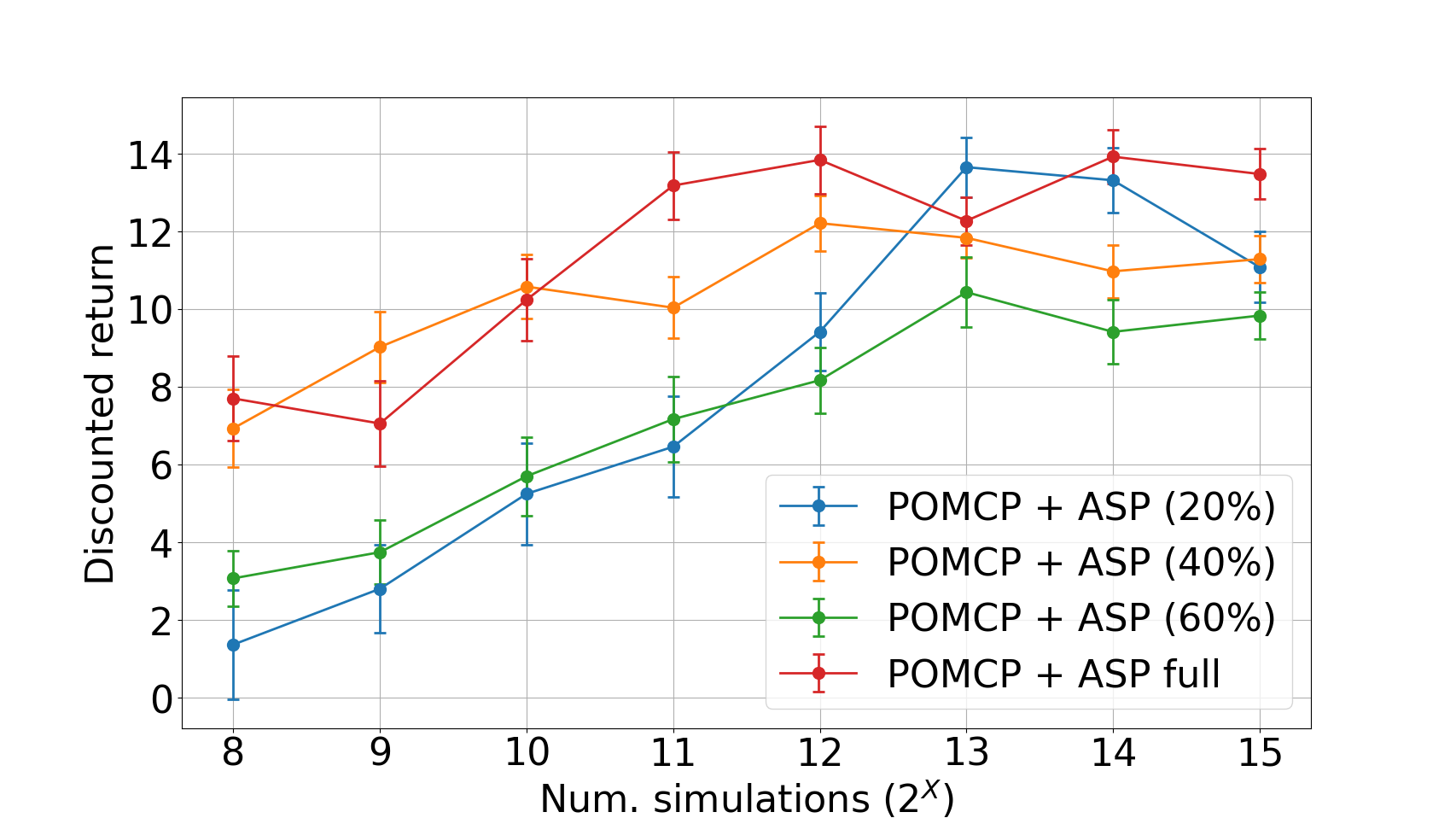}
    \caption{\label{fig:part_8rocks_perc}}
    \end{subfigure}\\
    \begin{subfigure}{0.45\textwidth}
    \centering
    \includegraphics[width=\linewidth]{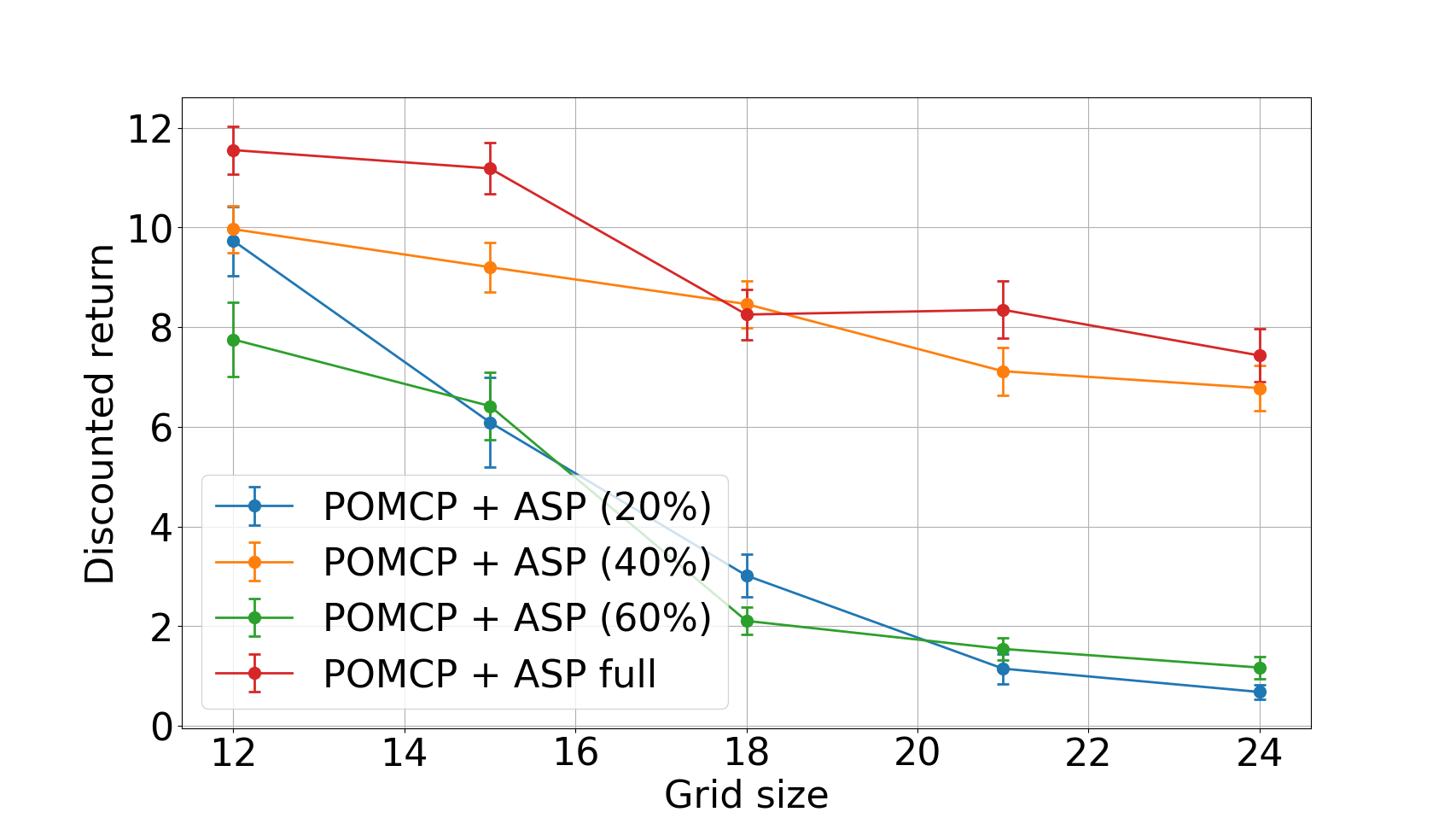}
    \caption{$M=4$ rocks.\label{fig:size_15part_4rocks_perc}}
    \end{subfigure}
    \begin{subfigure}{0.45\textwidth}
    \centering
    \includegraphics[width=\linewidth]{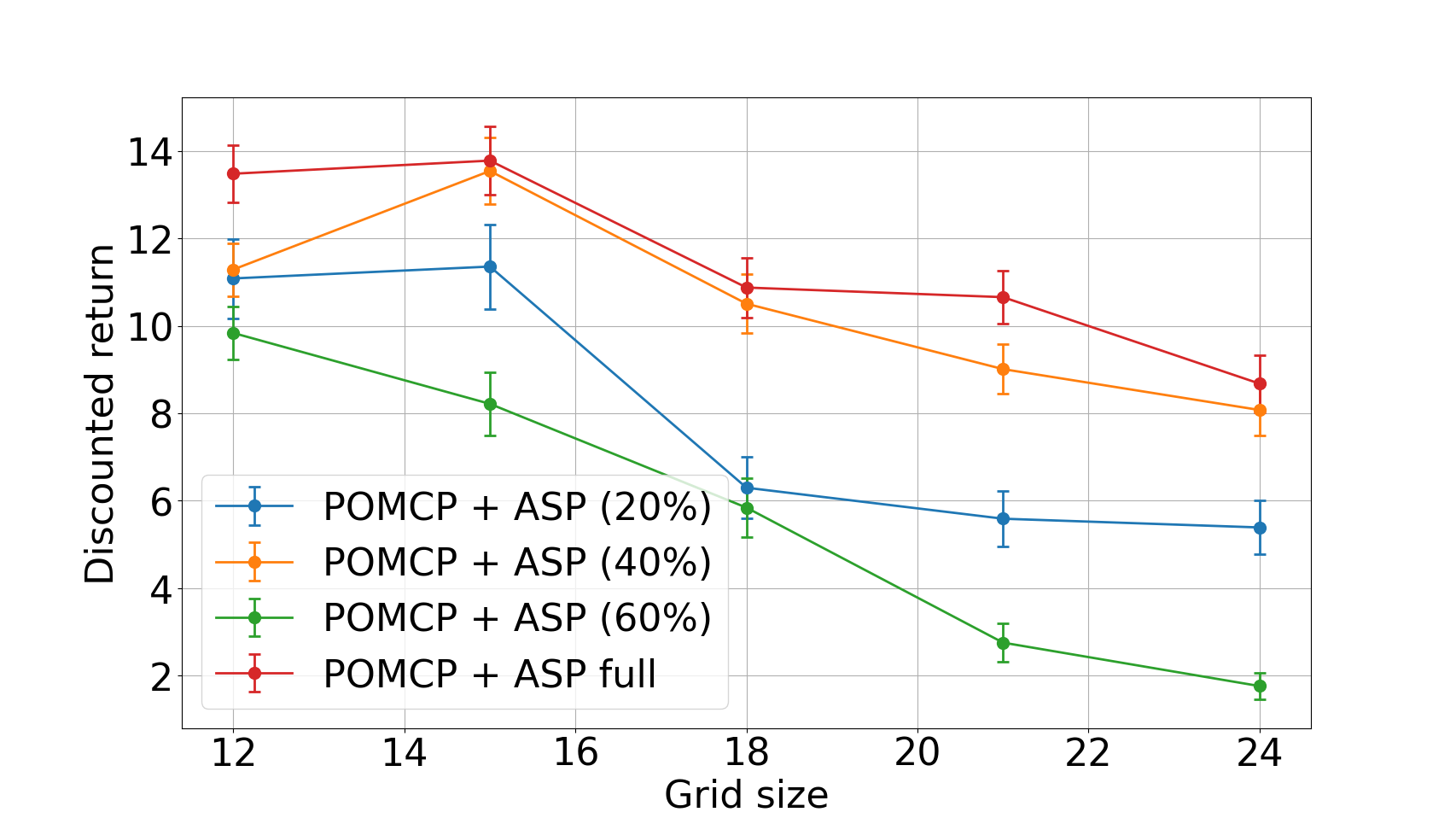}
    \caption{$M=8$ rocks.\label{fig:size_15part_8rocks_perc}}
    \end{subfigure}
    \caption{Results (mean $\pm$ standard deviation of the discounted return) for rocksample in POMCP with policy heuristics learned from few examples (\textbf{EXP-4}). \textbf{Top}: $12 \times 12$ grid with different number of simulations. \textbf{Bottom}: $2^{15}$ simulations with increasing grid size.}
    \label{fig:rs_few_pomcp}
\end{figure}
\begin{table}
    \centering
    \caption{\textbf{EXP-4}: Discrepancy between policy specifications for rocksample domain, learned from different numbers of traces / examples (with respect to final specifications in Equation \eqref{eq:rs_h}). Results for \stt{target(R)} do not consider the weak constraints for simplicity.}    
    \begin{tabular}{c|c|c|c|c}
        \textbf{Action} & \multicolumn{4}{c}{\textbf{Axiom distance (\%)}}\\ 
        \midrule
        Fraction of 1000 traces& 20\% & 40\% & 60\% & 80\%\\
        \midrule
        \stt{north} & 25 & 50 & 0 & 0 \\
        \stt{south} & 83 & 80 & 0 & 0 \\
        \stt{east} & 25 & 0 & 0 & 0 \\
        \stt{west} & 0 & 0 & 0 & 0 \\
        \stt{target(R)} & 38 & 75 & 57 & 0 \\
        \stt{check(R)} & 78 & 0 & 0 & 0 \\
        \stt{exit} & 82 & 67 & 82 & 0 \\
        \stt{sample(R)} & 0 & 0 & 0 & 0 \\
        \midrule
        \textbf{Average} & 41 & 34 & 17 & 0 \\
        \bottomrule
    \end{tabular}
    \label{tab:rules_diff}
\end{table}
We now vary the number of training example traces, selecting subsets of decreasing size from the original 1000 traces generated with $2^{15}$ particles in POMCP.
Specifically, we consider random subsets containing $\{20, 40, 60, 80\}\%$ of the total 1000 traces.
This corresponds to $\approx\{1700, 3400, 5100, 6800\}$ examples in rocksample, and $\approx\{9000, 18000, 27000, 36000\}$ for pocman, where the original number of examples (CDPIs) was $\approx 8500$ and $\approx 45000$, respectively.
As for \textbf{EXP-3}, the learned policy specifications do not change in pocman, hence we only report results for rocksample.
For brevity, we report learned heuristics for all training sizes only in Appendix \ref{app:exp_4_rules}.

First, we investigate the syntactic discrepancy between policy specifications learned from different example sets.
To this aim, we evaluate the \emph{distance between axioms} generated from a percentage $X\in\{20,40,60,80\}\%$ of the original 1000 training traces, with respect to the axioms generated from the total traces. Specifically, given 2 specifications $\pazocal{R}_1,\pazocal{R}_2$, corresponding to a same action atom \stt{a}$\in \pazocal{A}$ and involving independent sets of environmental feature atoms, $\pazocal{F}_1, \pazocal{F}_2$, we define the distance between the specifications as:
\begin{equation*}
    d(\pazocal{R}_1, \pazocal{R}_2) = \frac{|\pazocal{F}_1 \cup \pazocal{F}_2| - |\pazocal{F}_1 \cap \pazocal{F}_2|}{|\pazocal{F}_1 \cup \pazocal{F}_2|}
\end{equation*}
\noindent
denoting with $|\cdot|$ the cardinality of a set.
For instance, assume that $\pazocal{R}_1$ is the specification for action \stt{south} in Appendix \ref{app:rules_20}, and $\pazocal{R}_2$ is the specification for \stt{south} reported in Equation \eqref{eq:rs_h}. Then, $D(\pazocal{R}_1, \pazocal{R}_2) = 0.5$, since $\pazocal{F}_1 = \{\stt{delta\_y(R, D), target(R), \stt{D==-2}}\}$, $\pazocal{F}_2 = \{\stt{target(R), delta\_y(R, D), D}\leq\stt{-1}\}$. Hence, $|\pazocal{F}_1 \cup \pazocal{F}_2| = 4, |\pazocal{F}_1 \cap \pazocal{F}_2| = 2$.
Table \ref{tab:rules_diff} shows that, on average, the distance with respect to policy heuristics learned from the full dataset of traces (Equation \eqref{eq:rs_h}) decreases as the amount of examples increases, and it becomes null when 80\% of the dataset is considered.
This empirically demonstrates the convergence of ILASP to a set of stable policy specifications, as more examples are available.

Figure \ref{fig:rs_few_pomcp} shows results in POMCP, for different number of rocks ($M$) and grid sizes ($N$).
For any number of rocks, the performance of policy heuristics learned from different percentages of traces converges to the best heuristics learned from 100\% of the dataset \emph{POMCP + ASP full}, as the number of online simulations increases (Figures \ref{fig:part_4rocks_perc}-\ref{fig:part_8rocks_perc}). This again confirms that the asymptotic optimality of POMCP is guaranteed by the soft policy guidance approach, as the number of online simulations and particles for belief approximation increases. Policy specifications learned from 20\% and 60\% of the initial traces perform significantly worse than the ones learned from 40\% of the traces. This is even more evident as the grid size increases (Figures \ref{fig:size_15part_4rocks_perc}, \ref{fig:size_15part_8rocks_perc}). Looking at learned heuristics reported in Appendix \ref{app:exp_4_rules}, we can exploit \emph{interpretability of logic specifications} to easily identify a possible reason.
In fact, both for specifications learned from 20\% and 60\% of traces, the agent is incentivized to choose a rock as a target (\stt{target(R)}) according to the following axiom:
\begin{equation*}
    \stt{target(R) :- guess(R,V), not sampled(R), V}\leq \stt{80.}
\end{equation*}
\noindent
i.e., whenever the rock \stt{R} is valuable with probability \emph{lower than 80\%}.
This leads the agent to pick possibly invaluable rocks, resulting in a negative reward (-10).
On the contrary, the rule learned from 40\% of the traces only requires the rock to be at null distance from the agent:
\begin{equation}
\label{eq:bad_target_perc}
    \stt{target(R) :- dist(R,D), not sampled(R), D}\leq \stt{0.}
\end{equation}
\noindent
We remark that, as the grid size increases, the number of online simulations for POMCP ($2^{15}$) may not be sufficiently high to preserve performance with respect to pure POMCP (Figures \ref{fig:size_4rocks}, \ref{fig:size_8rocks}), especially when the learned specifications have significantly low quality. However, the deterioration in performance with specifications learned from 20\% and 60\% of the training traces is still more gradual than \emph{POMCP + ASP bad} in Figures \ref{fig:size_15part_4rocks_bad}-\ref{fig:size_15part_8rocks_bad}, where the return is very low even for small grid sizes.

\begin{figure}
    \centering
    \begin{subfigure}{0.45\textwidth}
    \centering
    \includegraphics[width=\linewidth]{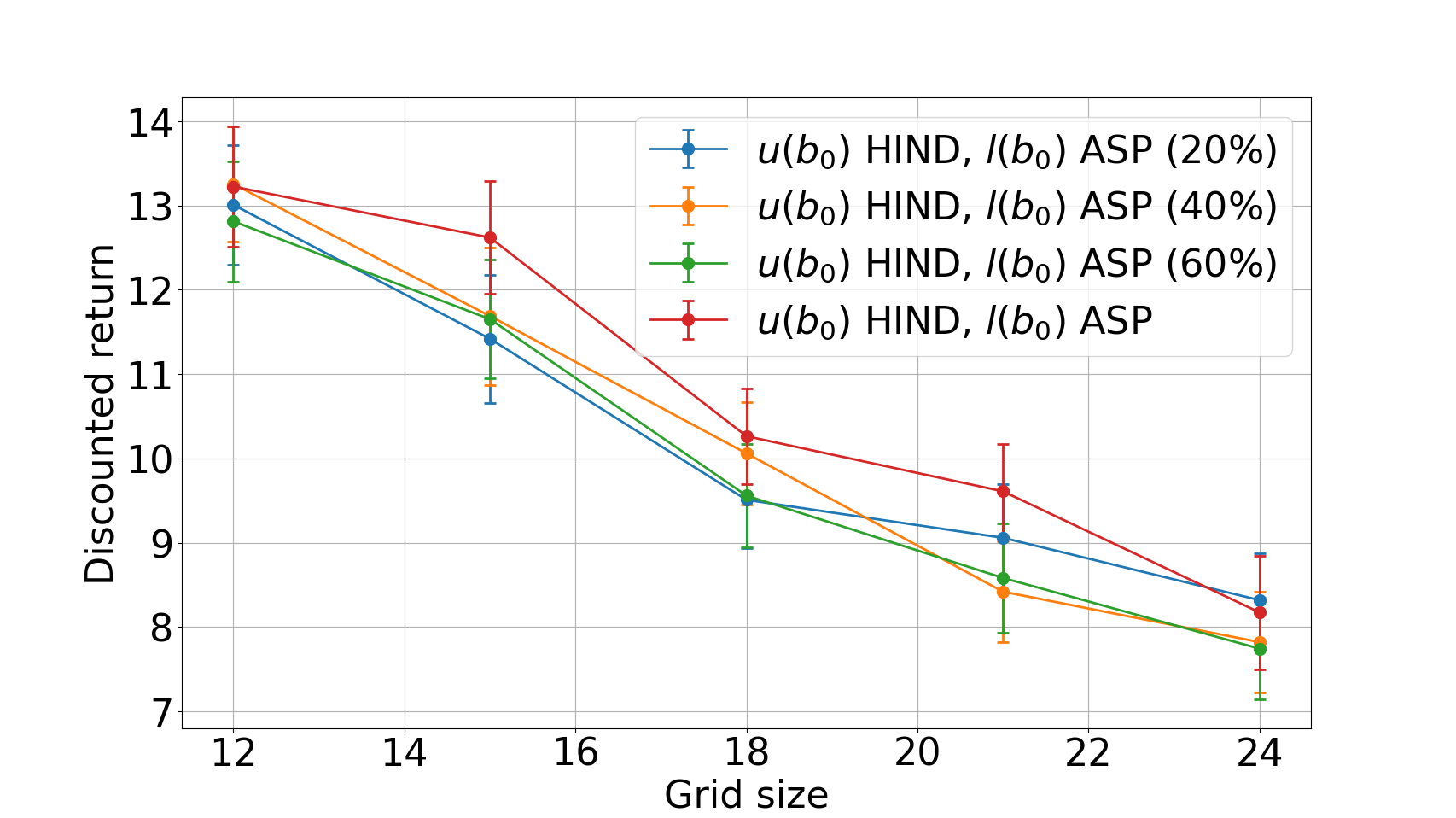}
    \caption{\label{fig:rs_perc_hind_4rocks}}
    \end{subfigure}
    \begin{subfigure}{0.45\textwidth}
    \centering
    \includegraphics[width=\linewidth]{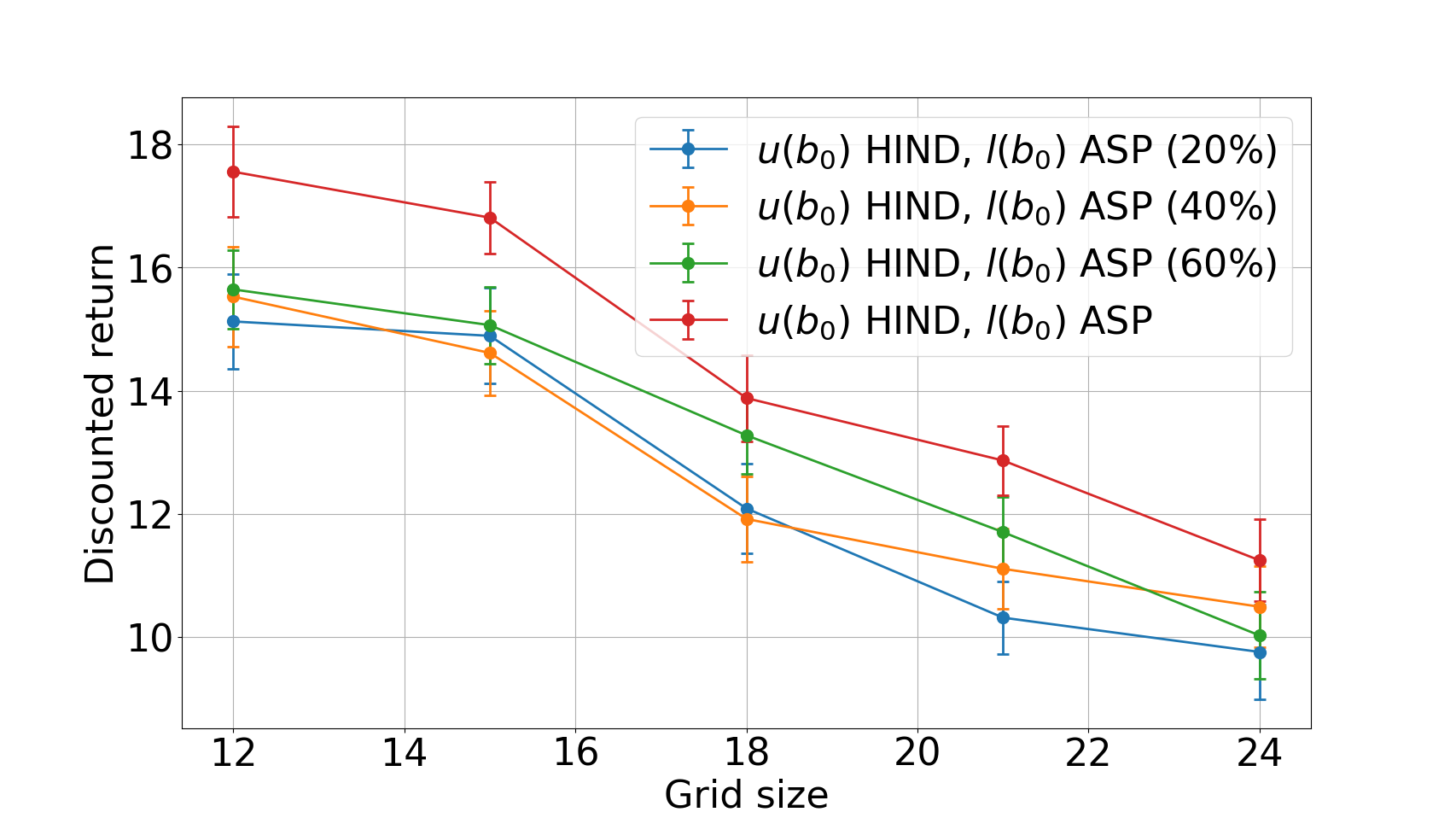}
    \caption{\label{fig:rs_perc_hind_8rocks}}
    \end{subfigure}\\
    \begin{subfigure}{0.45\textwidth}
    \centering
    \includegraphics[width=\linewidth]{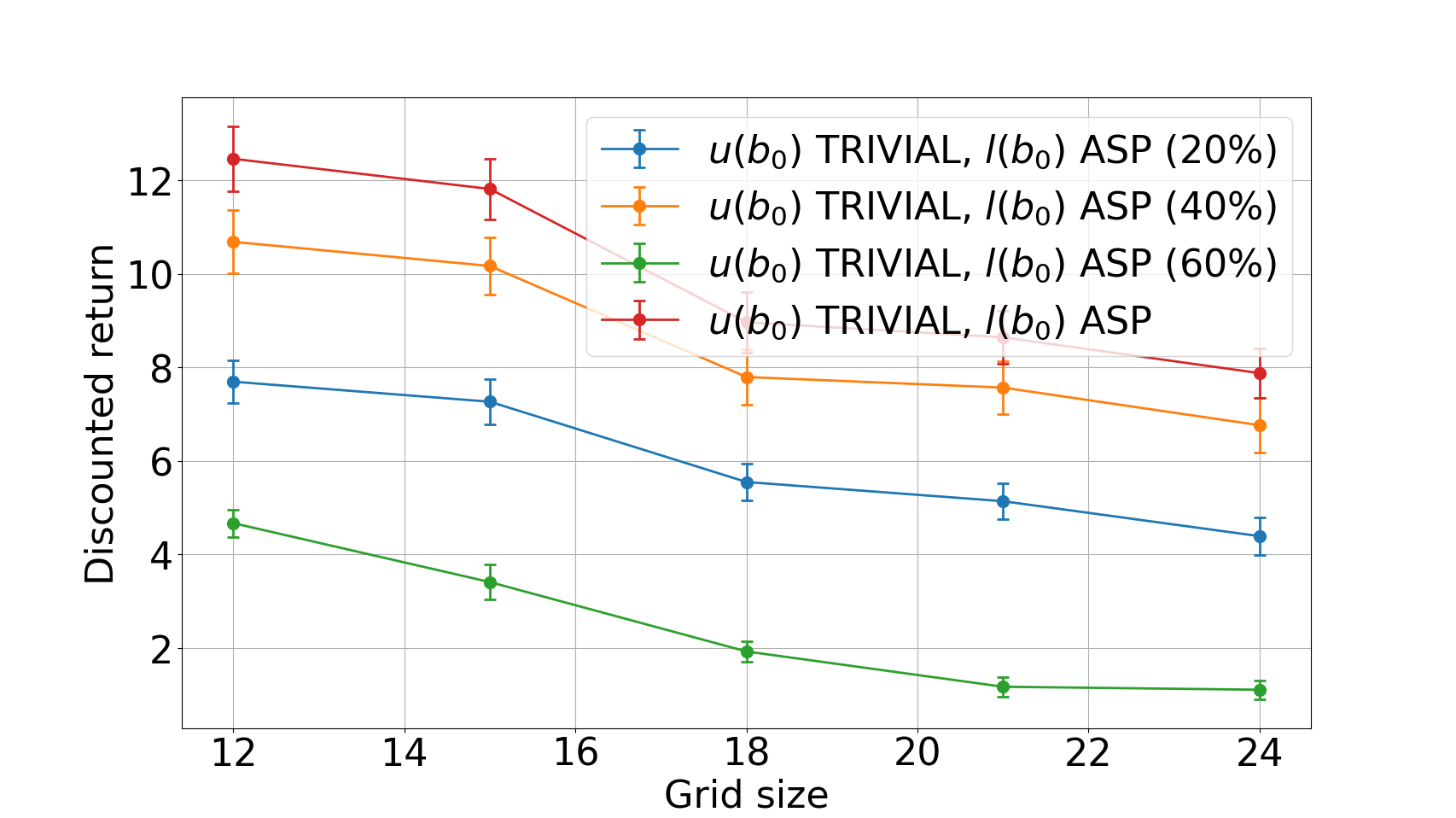}
    \caption{$M=4$ rocks\label{fig:rs_perc_triv_4rocks}}
    \end{subfigure}
    \begin{subfigure}{0.45\textwidth}
    \centering
    \includegraphics[width=\linewidth]{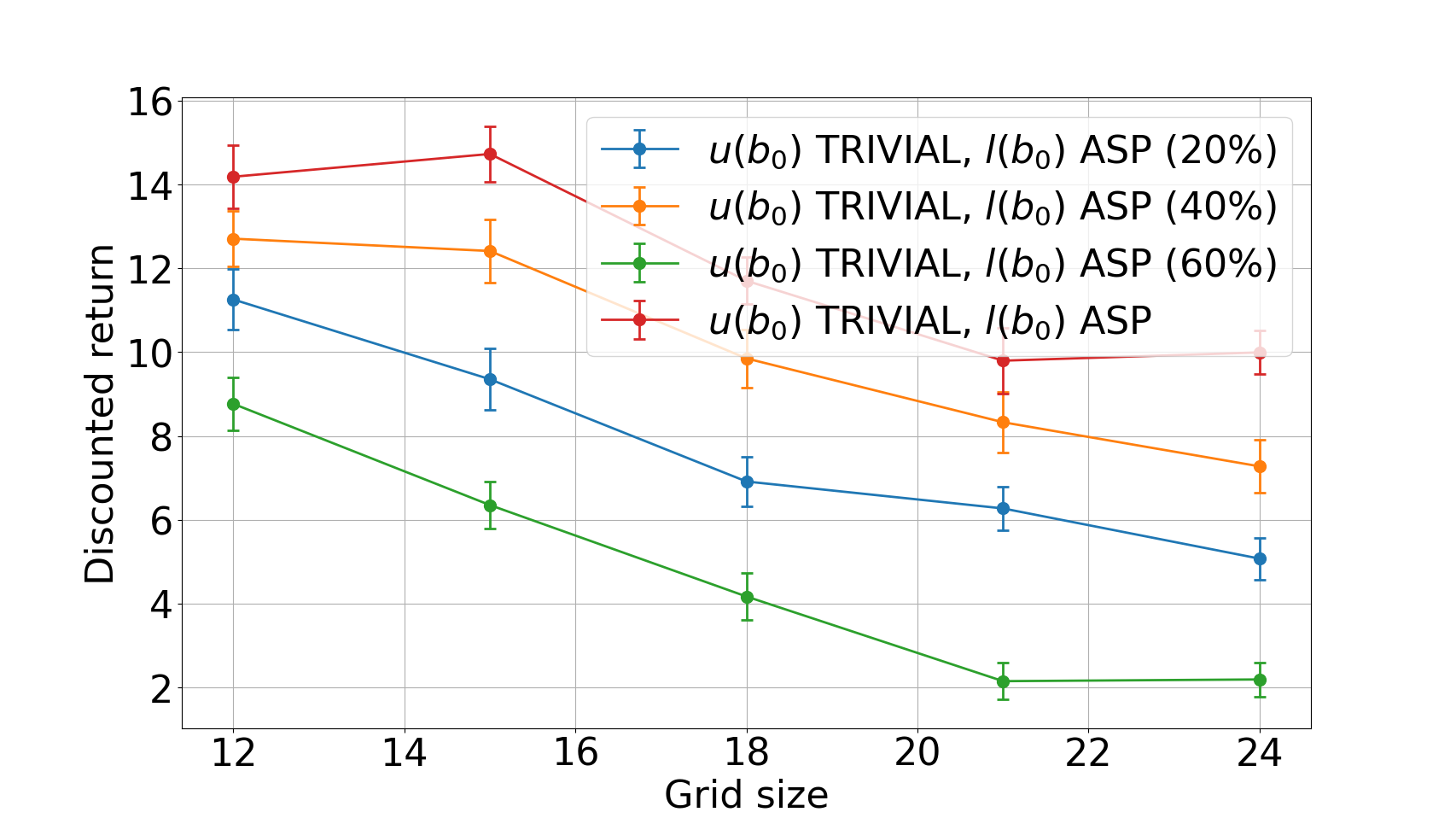}
    \caption{$M=8$ rocks\label{fig:rs_perc_triv_8rocks}}
    \end{subfigure}
    \caption{Results (mean $\pm$ standard deviation of the discounted return) for rocksample in DESPOT (C++) with policy heuristics learned from few examples (\textbf{EXP-4}). \textbf{Top}: Performance with hindsight (optimal) upper bound. \textbf{Bottom}: Performance with trivial upper bound.}
    \label{fig:rs_few_despot}
\end{figure}

In Figure \ref{fig:rs_few_despot} we show the results achieved by different policy heuristics in DESPOT (C++), with different number of rocks and on larger grid sizes.
When \emph{HIND} upper bound is used (Figures \ref{fig:rs_perc_hind_4rocks}, \ref{fig:rs_perc_hind_8rocks}), the discounted returns are comparable, confirming the important role of a good-quality upper bound for DESPOT performance.
When the \emph{TRIVIAL} upper bound is used (Figures \ref{fig:rs_perc_triv_4rocks}-\ref{fig:rs_perc_triv_8rocks}), instead, the difference in policy heuristics is crucial to determine final performance of the algorithm, and matches the trend observed in Figure \ref{fig:rs_few_pomcp}.
In particular, results evidence that specifications learned from 60\% of the original traces perform even worse than the ones learned from 20\% of traces (this is also partly evident from Figure \ref{fig:size_15part_8rocks_perc}, as the grid size increases). A possible reason lies in the higher number of \stt{check(R)} actions executed with \emph{ASP (60\%)} lower bound. Exploiting the interpretability of logical specifications, from Appendix \ref{app:exp_4_rules} we notice that, in this configuration, the following axiom holds:
\begin{equation*}
    \stt{check(R) :- target(R), guess(R,V), V}\leq\stt{50.}
\end{equation*}
This specification, combined with Equation \eqref{eq:bad_target_perc}, leads to more frequent grounding of \stt{check(R)}.
On the other hand, with \emph{ASP (20\%)} lower bound, the following axiom is learned: 
\begin{equation*}
    \stt{check(R) :- target(R), guess(R,V), V}\leq\stt{30.}
\end{equation*}
\noindent
which is more strict.

We can conclude that, though the axioms converge to the best ones as the number of examples increases, the quality of learned policy heuristics for online POMDP planning may not follow the same trend. Hence, an interpretable formalism to understand the meaning of policy heuristics is desired, to identify potential fallacies and prevent unwanted behavior of the agent.

\subsection{Comparison with Deep RL (\textbf{EXP-5})}
As mentioned in Section \ref{sec:sota} and highlighted in Section \ref{sec:learning_res}, inductive logic programming for learning policy specifications increases interpretability and requires relatively few examples (1000 executions, though even 800 are enough for rocksample, from Table \ref{tab:rules_diff}).
In this section, we perform further quantitative experiments to highlight the benefits of our method with respect to deep RL approaches.
We consider the work and code by \shortciteA{subramanian2022approximate}, where a deep neural architecture is adopted to efficiently train a Reinforcement Learning (RL) agent to solve the rocksample problem. 
Specifically, the authors outperform a state-of-the-art methodology based on proximal policy optimization and recurrent units, exploiting the approximate informativeness of the partially observable state (i.e., the so-called \emph{Approximate Information State, AIS}\footnote{We use the maximum mean discrepancy to evaluate the informativeness of partially observable states, since it achieves the best results in rocksample in the original paper by \shortciteA{subramanian2022approximate}.}).
\begin{figure}
    \centering
    \begin{subfigure}{0.45\textwidth}
    \centering
    \includegraphics[width=\linewidth]{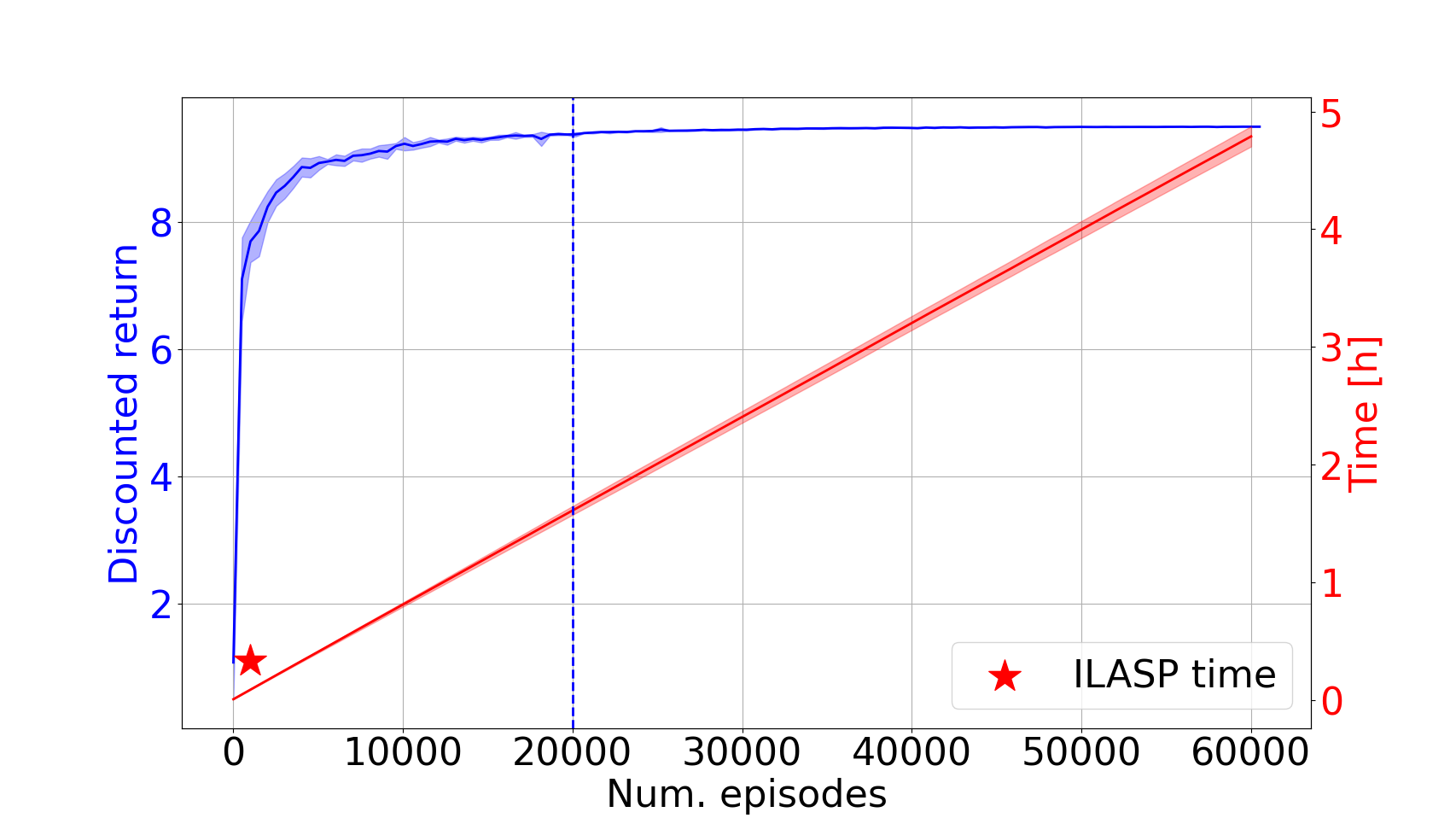}
    \caption{\label{fig:ais_training_4rocks}}
    \end{subfigure}
    \begin{subfigure}{0.45\textwidth}
    \centering
    \includegraphics[width=\linewidth]{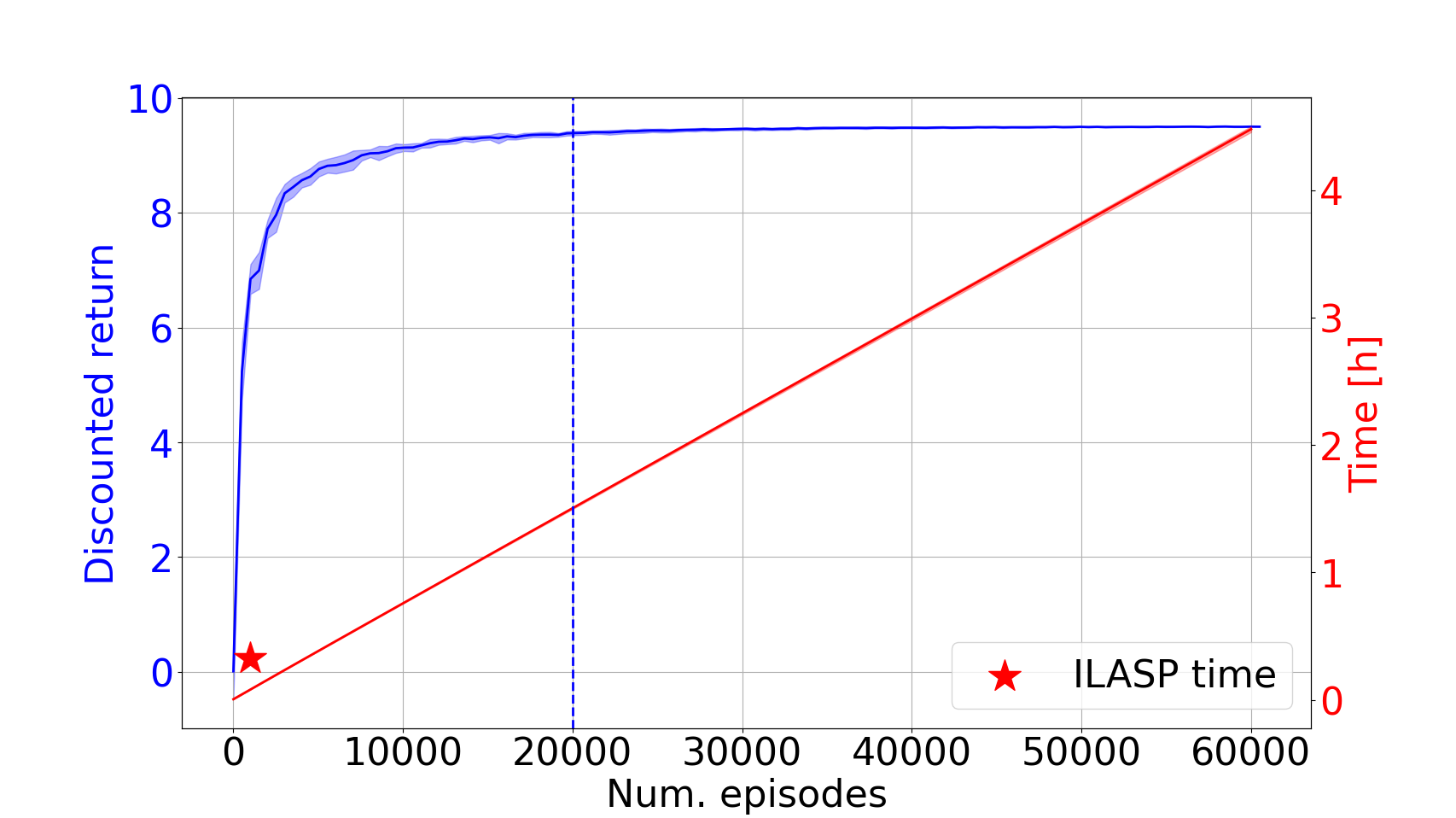}
    \caption{\label{fig:ais_training_8rocks}}
    \end{subfigure}\\
    \begin{subfigure}{0.45\textwidth}
    \centering
    \includegraphics[width=\linewidth]{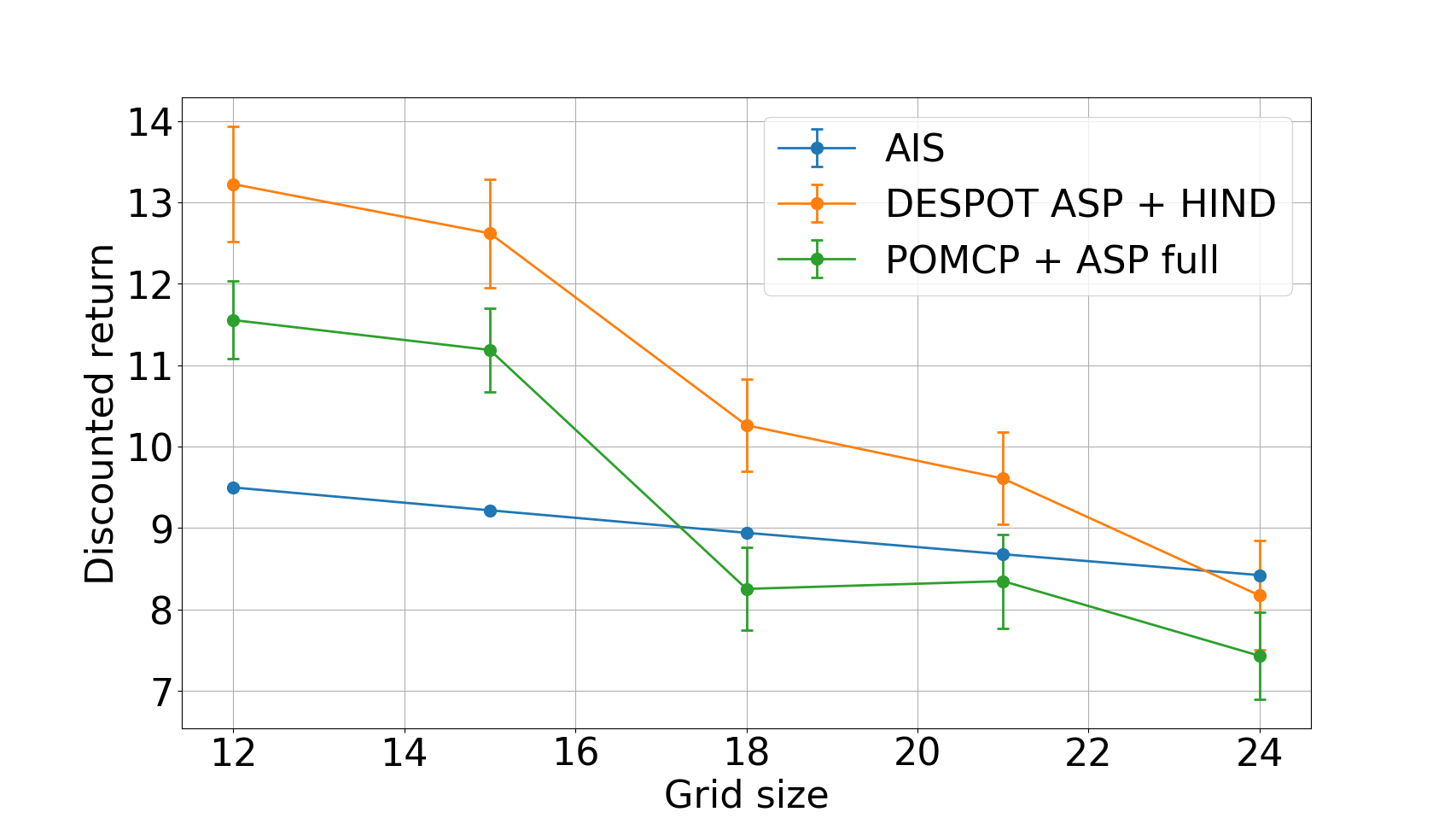}
    \caption{$M=4$ rocks\label{fig:ais_inference_4rocks}}
    \end{subfigure}
    \begin{subfigure}{0.45\textwidth}
    \centering
    \includegraphics[width=\linewidth]{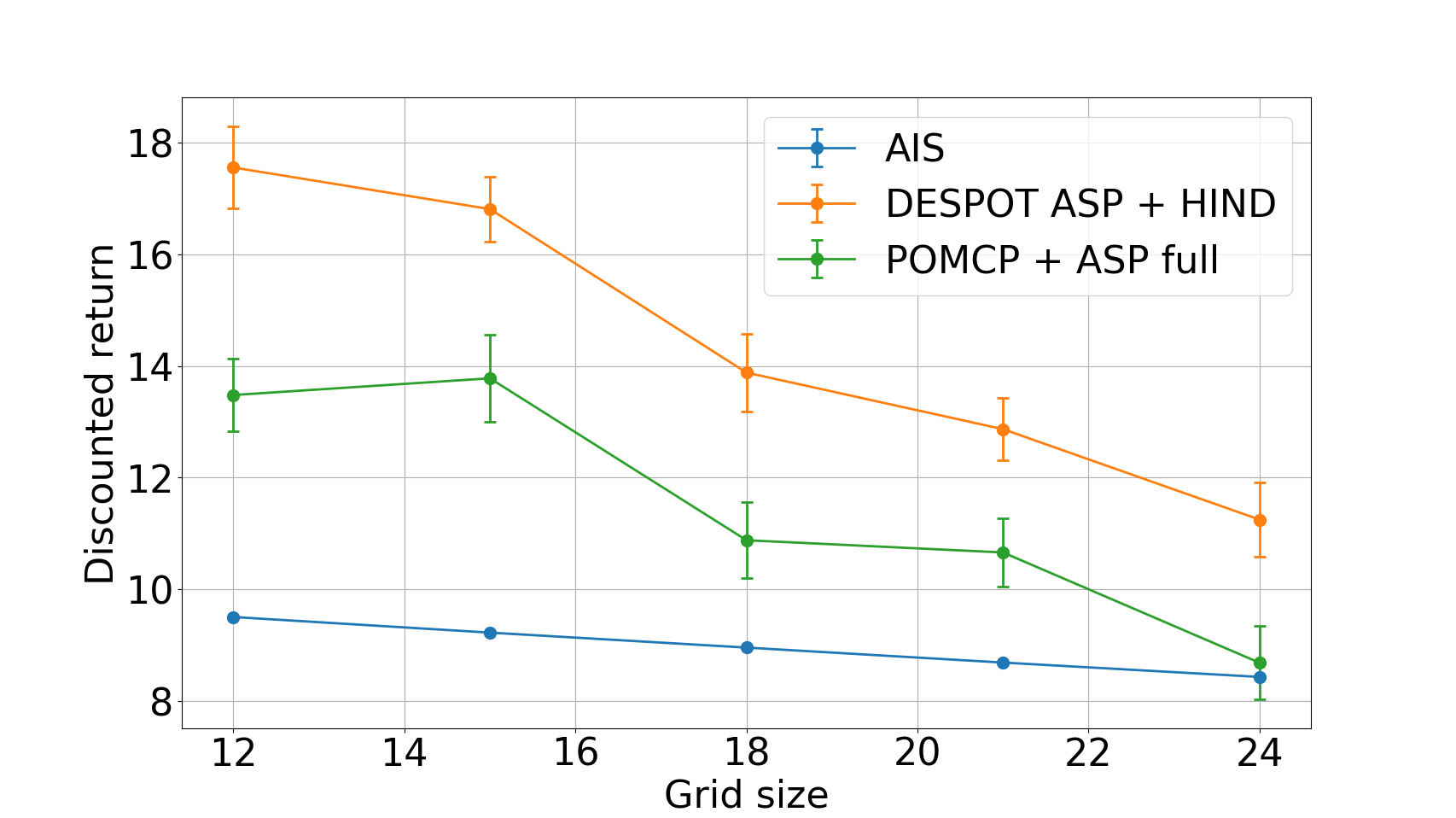}
    \caption{$M=8$ rocks\label{fig:ais_inference_8rocks}}
    \end{subfigure}
    \caption{\textbf{EXP-5}: Comparison with black-box RL for rocksample. \textbf{Top}: Training return and time (mean $\pm$ standard deviation) on a $12\times 12$ grid over 5 random seeds. \textbf{Bottom}: Inference return (mean $\pm$ standard deviation), training AIS for 20000 episodes.}
    \label{fig:rs_ais}
\end{figure}

Figures \ref{fig:ais_training_4rocks}-\ref{fig:ais_training_8rocks} show the training performance of RL (\emph{AIS}), in terms of training time and discounted return. Plots are obtained randomizing over 5 seeds. 
Since the RL method can generalize a learned policy to different grid sizes, but not to different number of rocks, we train it both with 4 (Figure \ref{fig:ais_training_4rocks}) and 8 rocks (Figure \ref{fig:ais_training_8rocks}), on a $12\times 12$ grid. 
Approximately 20000 episodes of rocksample are required to achieve a stable discounted return, corresponding to $\approx$\SI{1.5}{h} training time (blue dashed lines in Figures \ref{fig:ais_training_4rocks}-\ref{fig:ais_training_8rocks}). On the contrary, our learning methodology (\emph{ILASP time}) requires only $\approx$\SI{0.3}{h}, including the time to generate 1000 traces with 4 rocks and the time to learn policy specifications from ILASP (see Section \ref{sec:learning_res}). Moreover, we can learn rules from a scenario with 4 rocks and generalize them to a scenario with more rocks.
Our methodology is also clearly superior in terms of discounted return and generalization. In fact, from Figures \ref{fig:ais_inference_4rocks}-\ref{fig:ais_inference_8rocks}, DESPOT (C++) and POMCP solvers equipped with learned logical specifications significantly outperform the performance of \emph{AIS} on larger grid sizes (\emph{POMCP + ASP full} achieves lower return only with 4 rocks with grid size $N\geq 18$, from Figure \ref{fig:ais_inference_4rocks}).

\section{Discussion}\label{sec:discussion}
In this section we summarize the results of our experiments (with settings depicted in Table \ref{tab:exps}), highlighting the advantages of our pipeline for inductive learning of POMDP policy specifications and evidencing current limitations and potential directions to overcome them.
With reference to the objectives \textbf{EXP-1}-\textbf{EXP-5} stated in Section \ref{sec:exp}, we draw the following conclusions:
\begin{itemize}
    \item \emph{quality of learned policy heuristics} (\textbf{EXP-1}): we learned logical policy specifications for rocksample and pocman, starting from 1000 traces of execution (belief-action pairs) in small settings, namely, a $12\times 12$ grid with 4 rocks in rocksample and a $10\times 10$ map with 50\% food probability and 2 mildly aggressive ghosts (75\% chasing probability) in pocman. The traces correspond to POMCP executions achieving high discounted return, generated with $2^{15}$ online simulations and particles for belief approximation. The learned policy specifications were integrated in the UCT and rollout phases of POMCP, and used to compute the default policy and corresponding lower bound for DESPOT tree exploration. In POMCP, in the rocksample domain with the largest action space ($6+M$ actions, being $M$ the number of rocks), \textbf{policy heuristics significantly improve the performance of standard POMCP when the number of online simulations and particles is reduced} (up to $2^{10}$). Our ablation study highlights that \emph{performance increase when policy specifications are implemented in the rollout phase}. This introduces a significant computational cost; however, since a lower number of online simulations can be performed, \textbf{the overall cost of computing the best action at each time step is lower than POMCP without heuristics}. In pocman domain, POMCP performance do not significantly vary when heuristics are used. However, the lower bound based on logical specifications leads to performance (computational time per step and discounted return) comparable to optimal handcrafted heuristics by \shortciteA{silver2010monte} in DESPOT. 
    \item \emph{Generality and scalability of learned heuristics} (\textbf{EXP-2}): we tested learned policy specifications in more challenging conditions for the two tasks, which were not present in the original 1000 traces. In particular, we increased the number of rocks ($M$) and the grid size ($N$) in rocksample, and modified the map size and ghost and food parameters in pocman. This increases the size of the action space and the planning horizon, i.e., the number of actions required to accomplish the task. In general, \emph{learned heuristics achieve good performance also in these settings, proving their quality and generality even with very large actions space and long planning horizon} (see Table \ref{tab:exps}). Specifically, very good performance is achieved in rocksample with $20\times 20$ grid and 20 rocks (corresponding to 26 available actions), where learned heuristics are computationally more efficient than handcrafted ones both in POMCP and DESPOT (requiring $\approx 40\%-50\%$ time less to compute the best action at each step), while yielding an improvement with respect to pure POMCP and \emph{TRIVIAL} lower bound. Finally, such results are confirmed when the heuristics are applied to state-of-the-art extensions of POMDP solvers, e.g., AdaOPS by \shortciteA{wu2021adaptive} which extends DESPOT, proving the cross-solver generality of learned policy specifications. \textbf{In pocman, learned heuristics perform much better than the \emph{TRIVIAL} strategy and similarly to handcrafted, even when the task is significantly altered with respect to the training setting, requiring a different rationale to solve the task efficiently}. In fact, the large $17\times 19$ map has 4 more aggressive ghosts (chasing probability $\rho_g = 100\%$, while in the training traces in Section \ref{sec:learning_res} $\rho_g=75\%$) and very little food (food probability $\rho_f = 20\%$, while in the training traces in Section \ref{sec:learning_res} $\rho_g=50\%$). This scenario is significantly different from the one used to generate the 1000 training traces, since the policy of the agent must account for the scarcity of food pellets (giving positive reward) and a more strict constraint to avoid ghosts. Moreover, the planning horizon is very long (on average 85 actions are executed by the agent). Unfortunately, it was not possible to perform pocman experiments in AdaOPS, since this domain is not implemented in the publicly available Julia package. Nevertheless, we believe that rocksample results are sufficiently convincing, since it is the task with the largest action space (spanning from 10 actions with 4 rocks to 26 actions with 20 rocks) and the planning horizon reaches 67 actions when $N=M=20$, and 47 actions when $N=24, M=8$.
    \item \emph{Effect of low quality policy specifications} (\textbf{EXP-3})\footnote{\textbf{EXP-3}-\textbf{EXP-4} are performed only for rocksample, since the learned specifications for pocman are not affected when either the quality or the number of traces change.}: in the rocksample domain, we generated traces for learning logical policy specifications, using POMCP with $2^{10}$ online simulations and particles. This corresponds to the value where the discounted return achieved by \emph{POMCP + ASP full} significantly deteriorates (see Figure \ref{fig:exp2_rs_pomcp}). In these conditions, new policy specifications in Equation \eqref{eq:rs_h_bad} are learned, which contain low quality information about the task. Specifically, the criterion for selecting target rocks (\stt{target(R)}) and for exiting the grid (\stt{exit}) is different. In DESPOT, this deteriorates the performance of the online solver. However, \textbf{in POMCP, thanks to the \emph{soft policy guidance} approach adopted in the rollout phase instead of action pruning (see Section \ref{sec:met_pomcp}), the achieved discounted return is not severely affected, because the optimality of the algorithm with many particles and online simulations is preserved}. This experiment also evidences the effect of considering a non exhaustive set of environmental features to learn logical specifications. In fact, learned specifications exclude some environmental atoms, with respect to the best axioms learned with $2^{15}$ simulations. For instance, the specification for \stt{target(R)} includes only information about the distance between rocks and the agent, ignoring the probability of rock value. Conversely, the specification for \stt{exit} does not depend on the relative distance to unsampled rocks.
    \item \emph{Effect of the number of training traces} (\textbf{EXP-4}): in the rocksample domain, we varied the number of traces (hence, generated examples for ILASP), considering different percentages of them ranging from 20\% to 80 \%. This experiment evidences that \textbf{ILASP is able to converge to the best rules (from 1000 traces) as the number of examples increases}, progressively reducing the average distance from them (see Table \ref{tab:rules_diff}). Furthermore, this experiment strenghtens the results achieved with \textbf{EXP-3}, since learned rules often have very low quality (e.g., \stt{target(R)} selects rocks with low value probability, see Appendix \ref{app:exp_4_rules}) and depend only on a subset of the environmental features. Nevertheless, the performance of \emph{POMCP + ASP full} is still acceptable, at least for POMCP when the number of online simulations and particles is sufficiently high and the grid size does not increase much ($N \leq 18$). With fewer online simulations or in larger grids asymptotic optimality of POMCP cannot be guaranteed anymore, since bad policy heuristics may require significantly many online simulations. Hence, the performance significantly deteriorates when policy specifications are not good, as for DESPOT. However, \textbf{the interpretability of logical policy specifications allows to identify potential fallacies in the learned axioms, with the potential to prevent unwanted behavior of the agent}.
    \item \emph{Comparison with black-box architectures (deep RL)} (\textbf{EXP-5}): we compare our methodology with the work by \shortciteA{subramanian2022approximate}, a state-of-the-art methodology based on approximate information state exploration and the use of autoencoders and recurrent neural units to learn the best strategy to solve the rocksample domain. Our methodology has three fundamental advantages, compared to training an agent with deep RL. First, it has \textbf{higher data and computational efficiency}, requiring only 1000 example executions (for the rocksample task) vs. 20000 executions, and $\approx$\SI{0.3}{h} vs. $\approx$\SI{1.5}{h}. Second, \textbf{the discounted return is much higher with learned heuristics}, since POMCP and DESPOT equipped with logical specifications are mostly superior to the neural method. We believe that one reason for this lies in the use of the POMDP model (e.g., the transition model) in online solving with POMCP, DESPOT and AdaOPS, which is not available to a neural network. Third, \textbf{logical specifications are more interpretable}. This allows to identify potential fallacies in learned specifications, e.g., it is useful in \textbf{EXP-4} to explain the lower performance achieved by specifications learned from 20\% and 60\% of the training traces. Hence, logical policy heuristics increase the overall transparency of the decision making process.
\end{itemize}
\noindent
Our empirical analysis also evidenced potential directions of improvement of our methodology. Specifically, the following open problems remain to be addressed:
\begin{itemize}
    \item \emph{Definition of complete environmental features}: in our experiments, we assume that environmental features representing the POMDP belief are available from the transition and reward maps. This holds for many POMDP problems, thus, in general, the problem of defining these features is easier than manually defining policy specifications, since the policy is clearly unknown in POMDPs. Moreover, \textbf{EXP-3}-\textbf{EXP-4} show that our approach to exploit logical specifications in POMCP is robust to the unavailability of some features, yielding a moderate degradation of performance. However, in very complex real-world domains, e.g., robotic systems \shortcite{zuccotto2022,castellini2021partially,pajarinen2022pomdp} and environmental exploration \shortcite{thompson2019review,Steccanella2020,bravo2019use}, it may not be easy to extract all features from the definition of the POMDP problem, which may be partly incomplete. Moreover, in the rocksample domain we introduced the \stt{target(R)} atom, which is a useful high-level concept for the task, but cannot be directly retrieved from the transition and reward maps. A possible solution to this problem is to exploit other algorithms for ILP, e.g., bottom-up meta-interpretative learning by \shortciteA{hocquette2021complete} and Popper by \shortciteA{cropper2021learning}, which support predicate invention. This helps discover relevant concepts about the planning domain directly from examples. However, usually these algorithms do not support the full expressiveness of an action language for planning, e.g., weak constraints for optimal decision making in ASP. Hence, a tradeoff must be reached.
    \item \emph{Expressiveness of logical specifications}: we showed how to learn logical specifications for actions from examples of execution. However, our rules only consider the current situation of the task, e.g., the current observable state and belief, and provide suggestions to the POMDP solver only about the next action to be executed. In complex tasks involving, e.g., long planning horizons or non-trivial interaction with the environment, it may be useful to learn \emph{temporal policy heuristics}. However, the problem of learning temporal logic axioms is still an open challenge. In fact, typically only specific fragments of temporal logics are learned, e.g., in the event calculus paradigm \shortcite{meli2020towards,meli2021inductive}, or propositional temporal rules can be extracted from examples exploiting graph structure learning \shortcite{de2020imitation}. Hence, extension to temporal first order logic and the full ASP expressiveness still have to be investigated from a theoretical perspective. We believe that combining temporal expressiveness with predicate invention could improve the quality of learned policy heuristics and their impact on online POMDP planning, allowing to emulate the performance of handcrafted heuristics by \shortciteA{silver2010monte}, e.g., considering the number of times a rock has been checked in rocksample (see Section \ref{sec:learning_res}).
    \item \emph{Online learning of policy specifications}: our methodology allows to learn policy specifications from traces of execution offline. In the future, it will be interesting to investigate \emph{online learning} of such specifications, as the agent acquires experience from its interaction with the environment. This has the potential to refine and improve agent's performance in more challenging task scenarios. It could be achieved by, e.g., extending state-of-the-art approaches for online learning of ASP rewards (\shortciteR{furelos2021induction}, currently focusing on fully observable scenarios, i.e., MDPs) with fast online ILP techniques, such as incremental ILASP from streams of data \shortcite{law2022search}.
\end{itemize}

\section{Conclusion}\label{sec:conc}
State-of-the-art online POMDP planners require good task-specific policy heuristics to achieve good performance in complex environments involving many actions or long planning horizons.
In this paper, we proposed a methodology based on ILP to learn ASP policy specifications from POMDP traces of execution generated by \emph{any solver without heuristics}.
We integrate learned heuristics in different state-of-the-art online POMDP planners, specifically for biasing rollout and action values in POMCP, and to define the default policy and lower bound for state value in DESPOT and AdaOPS. In this way, we achieve similar performance as optimal handcrafted heuristics in two challenging benchmark domains: rocksample, involving many actions and potentially long planning horizon; and pocman, with very long planning horizon. Moreover, learned heuristics help reduce the time for optimal action computation when the action space is large, and can be learned from small domains and efficiently generalize to larger more challenging problem instances (e.g., more rocks and larger grid sizes in rocksample, larger maps and more numerous and aggressive ghosts in pocman with fewer food pellets). Also, the use of logic programming allows to learn \emph{interpretable heuristics} which provide some explanation of the decision-making process of the POMDP agent, significantly outperforming a state-of-the-art methodology based on autoencoders and recurrent neural units in terms of \emph{training data and computational efficiency} and \emph{achieved discounted return}. We achieve this by integrating commonsense environmental semantic features in our pipeline, which are related to domain knowledge concerning the basic POMDP problem definition (the transition map), hence can be defined by a user with basic domain knowledge. Crucially, adopting a \emph{soft policy guidance} approach in POMCP, the planner is robust to bad heuristics learned, e.g., from low-quality and few training traces or when some semantic features are missing.

This paper is a first attempt to learn heuristics for POMDP solvers from execution traces. As such, it paves the way towards several interesting research directions. In the future, we plan to investigate extensions to temporal logic and predicate invention, to increase the quality of policy specifications and online planning performance. Moreover, we will investigate more challenging real-world domains, e.g., robotic domains, and implement an ILP methodology for online learning and refinement of policy heuristics from experience.

\vskip 0.2in

\acks{
This project has received funding from the Italian Ministry for University and Research, under the PON “Ricerca e Innovazione” 2014-2020 (grant agreement No. 40-G-14702-1).
}

\appendix
\section{Learned Policy Specifications for Rocksample in \textbf{EXP-4}}\label{app:exp_4_rules}
Here, we report learned policy specifications for rocksample, starting from different subsets of the original 1000 training traces used in Section \ref{sec:learning_res}. Specifically, we consider subsets containing $\{20, 40, 60, 80\}\%$ of the original traces, hence corresponding CDPIs. We do not report learned specifications from $80\%$ of the total examples, since they correspond to the best policy heuristics in Equation \eqref{eq:rs_h}.

\subsection{Policy Heuristics from 20\% of the Original Traces}\label{app:rules_20}
The following specifications are learned from this subset of traces / examples.
\begin{align*}
    \nonumber&\stt{east :- target(R), delta\_x(R,D), D} \geq \stt{1, D} \leq \stt{2.}\\
    \nonumber&\stt{west :- target(R), delta\_x(R,D), D} \leq \stt{-1.}\\
    \nonumber&\stt{north :- target(R), delta\_y(R,D), D} \geq \stt{1.}\\
    \nonumber&\stt{south :- target(R), delta\_y(R,D), D == -2.}\\
    \nonumber&\stt{target(R) :- dist(R,D), not sampled(R), D}\leq \stt{0.}\\
    \nonumber&\stt{target(R) :- guess(R,V), not sampled(R), V}\leq \stt{80.}\\
    \nonumber&\stt{check(R) :- guess(R,V), target(R), V} \leq \stt{30.}\\
    \nonumber&\stt{sample(R) :- target(R), dist(R,D), D} \leq \stt{0, not sampled(R), guess(R,V), V} \geq \stt{90.}\\
    \nonumber&\stt{exit :- num\_sampled(N), N} \geq \stt{25.}
\end{align*}

\subsection{Policy Heuristics from 40\% of the Original Traces}
The following specifications are learned from this subset of traces / examples.
\begin{align*}
    \nonumber&\stt{east :- target(R), delta\_x(R,D), D} \geq \stt{1.}\\
    \nonumber&\stt{west :- target(R), delta\_x(R,D), D} \leq \stt{-1.}\\
    \nonumber&\stt{north :- target(R), delta\_y(R,D), D} \geq \stt{2.}\\
    \nonumber&\stt{south :- target(R), delta\_x(R,D), D} \leq \stt{0.}\\
    \nonumber&\stt{target(R) :- dist(R,D), not sampled(R), D}\leq \stt{0.}\\
    \nonumber&\stt{check(R) :- guess(R,V), target(R), V} \leq \stt{50.}\\
    \nonumber&\stt{check(R) :- guess(R,V), dist(R,D), not sampled(R), V} \leq \stt{80, D} \leq \stt{0.}\\
    \nonumber&\stt{sample(R) :- target(R), dist(R,D), D} \leq \stt{0, not sampled(R), guess(R,V), V} \geq \stt{90.}\\
    \nonumber&\stt{exit :- num\_sampled(N), N} \geq \stt{25, guess(R,V), V} \geq \stt{80, not sampled(R).}
\end{align*}

\subsection{Policy Heuristics from 60\% of the Original Traces}
The following specifications are learned from this subset of traces / examples.
\begin{align*}
    \nonumber&\stt{east :- target(R), delta\_x(R,D), D} \geq \stt{1.}\\
    \nonumber&\stt{west :- target(R), delta\_x(R,D), D} \leq \stt{-1.}\\
    \nonumber&\stt{north :- target(R), delta\_y(R,D), D} \geq \stt{1.}\\
    \nonumber&\stt{south :- target(R), delta\_y(R,D), D == -2.}\\
    \nonumber&\stt{target(R) :- guess(R,V), not sampled(R), V}\leq \stt{80.}\\
    \nonumber&\stt{check(R) :- guess(R,V), target(R), V} \leq \stt{50.}\\
    \nonumber&\stt{check(R) :- guess(R,V), dist(R,D), not sampled(R), V} \leq \stt{80, D} \leq \stt{0.}\\
    \nonumber&\stt{sample(R) :- target(R), dist(R,D), D} \leq \stt{0, not sampled(R), guess(R,V), V} \geq \stt{90.}\\
    \nonumber&\stt{exit :- num\_sampled(N), N} \geq \stt{25.}
\end{align*}

\bibliography{biblio}
\bibliographystyle{theapa}

\end{document}